\documentclass{article} % For LaTeX2e
\usepackage{iclr2027_conference,times}

\usepackage{amsmath,amsfonts,bm}

\def\eqref#1{equation~\ref{#1}}
\def\1{\bm{1}}

\DeclareMathAlphabet{\mathsfit}{\encodingdefault}{\sfdefault}{m}{sl}
\SetMathAlphabet{\mathsfit}{bold}{\encodingdefault}{\sfdefault}{bx}{n}

\newcommand{\bla}[1]{}

\usepackage{xcolor}
\usepackage{graphicx}
\usepackage{subcaption}
\usepackage{booktabs}
\usepackage{enumitem}
\usepackage{float} % only needed if you use [H]
\usepackage[normalem]{ulem} % only needed if you use \sout, \uline, etc.

\usepackage{mathtools} % loads amsmath
\usepackage{amssymb}   % loads amsfonts
\usepackage{amsthm}
\usepackage{bm}
\usepackage{mathrsfs}
\usepackage{cancel}    % only needed if you use \cancel
\usepackage{dsfont}

\usepackage{algorithm}
\usepackage{algorithmic}

\usepackage{listings}
\usepackage[most]{tcolorbox}

\usepackage{hyperref}
\usepackage{url}
\theoremstyle{plain}
\newtheorem{theorem}{Theorem}[section]
\newtheorem{proposition}[theorem]{Proposition}
\newtheorem{lemma}[theorem]{Lemma}
\newtheorem{corollary}[theorem]{Corollary}

\theoremstyle{definition}

\theoremstyle{remark}

\newcommand{\theorembg}{gray!10}
\newcommand{\propositionbg}{gray!10}
\newcommand{\lemmabg}{gray!10}
\newcommand{\corollarybg}{gray!10}
\newcommand{\remarkbg}{gray!10}

\tcbset{
  theoremstylebox/.style={
    enhanced,
    breakable,
    colframe=gray!15,
    boxrule=0.5pt,
    arc=2pt,
    before skip=8pt,
    after skip=8pt,
    left=6pt,
    right=6pt,
    top=6pt,
    bottom=6pt
  }
}

\tcolorboxenvironment{proof}{
  theoremstylebox,
  colback=gray!7
}

\tcolorboxenvironment{theorem}{
  theoremstylebox,
  colback=\theorembg
}

\tcolorboxenvironment{proposition}{
  theoremstylebox,
  colback=\propositionbg
}

\tcolorboxenvironment{lemma}{
  theoremstylebox,
  colback=\lemmabg
}

\tcolorboxenvironment{corollary}{
  theoremstylebox,
  colback=\corollarybg
}

\tcolorboxenvironment{remark}{
  theoremstylebox,
  colback=\remarkbg
}

\usepackage{etoolbox}

\pretocmd{\appendix}{%
  \renewcommand{\theorembg}{red!5}%
  \renewcommand{\propositionbg}{orange!10}%
  \renewcommand{\lemmabg}{blue!5}%
  \renewcommand{\corollarybg}{green!5}%
  \renewcommand{\remarkbg}{yellow!10}%
}{}{}

\definecolor{rwblue}{HTML}{24515F}
\definecolor{rwgray}{HTML}{6B7280}
\definecolor{rwbg}{HTML}{F7F7F8}

\NewDocumentCommand{\RelatedWork}{O{} O{} m m m +m}{%
  \begin{tcolorbox}[
    enhanced,
    breakable,
    colback=rwbg,
    colframe=rwblue!35,
    boxrule=0.5pt,
    arc=2mm,
    left=1.5mm,
    right=1.5mm,
    top=1mm,
    bottom=1mm,
    before skip=0.7em,
    after skip=0.7em
  ]
  \noindent
  \href{#3}{\textbf{#4}}%

\if\relax\detokenize{#2}\relax
  \else
  \par\vspace{0.25em}
  \noindent{\footnotesize\color{rwgray}#2}
\fi

  \par\vspace{0.25em}
  \noindent{\footnotesize\color{rwgray}#1}

  \par\vspace{0.25em}
  \noindent{\footnotesize\color{rwgray}#5}

  \par\vspace{0.35em}
  \noindent\textbf{Summary.} {\itshape #6}
  \end{tcolorbox}
}

\newcounter{researchidea}

\newtcolorbox{ResearchIdeaBox}[2][]{%
  enhanced,
  breakable,
  width=\linewidth,
  colback=yellow!4,
  colframe=yellow!45!black,
  boxrule=0.6pt,
  arc=1.5mm,
  left=6pt,
  right=6pt,
  top=6pt,
  bottom=6pt,
  before skip=8pt,
  after skip=8pt,
  before upper={%
    \refstepcounter{researchidea}%
    \raggedright%
    \setlength{\parindent}{0pt}%
  },
  title={Idea \theresearchidea: #2},
  fonttitle=\bfseries,
  #1
}

\usepackage[most]{tcolorbox}

\newtcolorbox{questionbox}{
    enhanced,
    width=0.85\textwidth,
    center,
    colback=gray!4,
    colframe=gray!45,
    boxrule=0.5pt,
    arc=2mm,
    left=5mm,
    right=5mm,
    top=3mm,
    bottom=3mm,
    before skip=8pt,
    after skip=8pt,
    fontupper=\centering\itshape
}

\newtcolorbox{proofplan}{
  colback=gray!5,
  colframe=gray!60,
  title=Proof sketch,
  fonttitle=\bfseries,
  breakable
}

\usepackage{booktabs}
\usepackage[table]{xcolor}
\usepackage{makecell}

\usepackage[most]{tcolorbox}

\newtcolorbox{todobox}{
    breakable,
    enhanced,
    colback=red!4,
    colframe=red!70!black,
    boxrule=0.8pt,
    arc=2mm,
    left=4mm,
    right=4mm,
    top=2mm,
    bottom=2mm,
    title={TODO List},
    fonttitle=\bfseries,
    coltitle=red!70!black,
    before skip=1em,
    after skip=1em
}

\definecolor{polinatagcolor}{HTML}{F59E0B}
\definecolor{lukatagcolor}{HTML}{80669D}

\definecolor{pdblue}{HTML}{1F5AA6}

\usepackage{amssymb}
\usepackage{enumitem}

\definecolor{newquestionblue}{HTML}{2563EB}

\usepackage{bbm}
\usepackage{centernot}
\usepackage{threeparttable}

\definecolor{diffblue}{HTML}{4C78A8}
\newcommand{\diff}[1]{\textcolor{diffblue}{\scriptsize\ensuremath{(#1)}}}

\usepackage{titletoc}

\usepackage{booktabs}
\usepackage[table]{xcolor}
\usepackage{array}
\usepackage{graphicx}
\usepackage{makecell}
\definecolor{featblue}{RGB}{74,120,170}

\definecolor{nearGreen}{RGB}{201,230,211}
\definecolor{warmOrange}{RGB}{244,196,143}

\title{Do unlearning algorithms learn\\ the right features?}

\author{
Polina Dolgova$^{1,2}$ \quad
Amartya Sanyal$^{3}$ \quad
Sebastian U. Stich$^{1}$\\
$^1$CISPA Helmholtz Center for Information Security \quad
$^2$Saarland University\\
$^3$Department of Computer Science, University of Copenhagen\\
\texttt{\{polina.dolgova, stich\}@cispa.de} \quad
\texttt{amsa@di.ku.dk}
}

\iclrfinalcopy 
\begin{document}

\maketitle

\begin{abstract}
Machine unlearning is commonly evaluated with standard post-deletion metrics that compare an unlearned model to one retrained on the remaining data. Such snapshot evaluations can conceal consequential differences in the features the two models encode. We prove that if an unlearned and a retrained model agree on their outputs but encode different features, a single gradient step on their prediction heads can break this agreement. We further show that deleting an arbitrarily small fraction of the data can cause retraining to select qualitatively different features, so matching retraining can require learning features that the full model never had. We then introduce a measure of feature recovery that separates missing from extraneous features. Across eight unlearning methods, models that perform well on standard metrics often fail to recover the retrained features. This mismatch predicts deviations from retraining under subsequent relearning and downstream adaptation, and directly correcting it systematically reduces these deviations.
\end{abstract}

\section{Introduction}

Training data may need to be removed after a model has been trained, for example, in response to privacy requests, licensing issues, or discovered errors. Deleting these examples from the dataset, however, does not remove their influence from the trained model. Retraining from scratch on the remaining data therefore provides the natural reference for successful deletion, but it can be expensive. Empirical machine unlearning instead aims to approximate retraining at substantially lower cost.\looseness=-1

In image classification, most evaluations assess unlearning immediately after the model has been updated. These evaluations range from standard utility and forgetting metrics~\citep{fan2024salun, ebrahimpourboroojeny2025amunadversarialmachineunlearning} to more targeted comparisons of predictive behavior \citep{ebrahimpourboroojeny2025necessityoutputdistributionreweighting,dolgova2026forgettingneighborslocalizedcollateral}, but largely evaluate the model at a single point in time. An unlearned model may later be trained on new data, adapted to downstream tasks, updated in response to further deletion requests, or exposed to relearning attacks~\citep{hu2025unlearning, xu2026unlearning, siddiqui2025from}. Differences that are invisible immediately after unlearning may become consequential during such further training. This leads to our central question: \looseness=-1
\begin{quote}

When an unlearned model matches retraining in its predictions, what else must it recover for that agreement to persist under further training; and do common unlearning algorithms achieve this?
\end{quote}

We argue that a key part of the answer is the \emph{feature solution} selected by retraining:\ which predictive features a model learns and how strongly it represents them. Two models can make similar predictions using different features and therefore respond differently when training continues. 

\textbf{We first ask when differences in learned features can remain harmless.} Different representations can produce the same predictions under a fixed head, but this agreement may disappear once the head is updated. 
We show that preserving agreement across all one-step head updates requires matching model-defined feature directions and strengths. Feature recovery therefore affects whether agreement with retraining persists during adaptation.

\textbf{Next, we show that unlearning can require learning new features.} One might hope that retraining after deletion largely preserves the features learned on the full data, so that successful unlearning only needs to remove or weaken existing information. We show otherwise:\ in a simple theoretical construction, the full model relies on an easy shortcut, while deleting an arbitrarily small fraction of the data causes retraining to suppress it and learn a predictive feature that the full model never had. Matching retraining can therefore require learning a new feature, rather than merely erasing old ones. We also show that several approximate-unlearning dynamics either fail to produce this transition or can require optimization time approaching that of retraining.

\textbf{Empirically, we distinguish feature change from feature recovery.} Using the model-defined feature framework of \citet{jiang2024on}, we introduce metrics for comparing unlearned and retrained feature solutions. These separately capture missing features that appear during retraining and features that should disappear after unlearning. Together with comparison to the full-data solution, this distinguishes feature persistence from off-target change. On ViT-Tiny/CIFAR-100, we evaluate eight approximate unlearning methods for image classification and find that methods with strong static unlearning performance can nevertheless produce very different future solutions:\ some remain close to the full-data model, while others move far from it without approaching retraining. We observe the same qualitative behavior with ResNet-56, SVHN, and smaller deletion fractions.

\textbf{Finally, we connect feature mismatch to behavior under adaptation.} Feature mismatch on the forget distribution tracks how far the model deviates from retraining after partial relearning. On downstream tasks, missing retraining features on target inputs similarly track losses in transfer performance (Table~\ref{tab:feature-gap-teaser}). We then test these relationships more directly by intervening on the feature components identified in our comparison: we restore missing retraining features and remove unmatched unlearned features. Reducing the measured mismatch through these interventions generally brings the model closer to retraining. Together, these results motivate evaluating unlearning not only by immediate behavior, but also by whether it recovers the feature solution needed to remain aligned with retraining under subsequent adaptation.

\begin{table*}[t]
\caption{\textbf{Missing retraining features predict downstream accuracy loss.} ViT-Tiny/CIFAR-100 with 20\% of the training data removed; a fresh linear head is trained on Flowers-102 on top of each frozen representation. Feature discrepancy measures how much of the retraining feature solution is missing on Flowers-102 inputs (Section~\ref{sec:feature-recovery}). Retraining transfers as well as the full-data model, while unlearning methods lose accuracy as their discrepancy grows.}
\label{tab:feature-gap-teaser}
\centering
\scriptsize
\setlength{\tabcolsep}{5pt}
\renewcommand{\arraystretch}{1.2}
\setlength{\aboverulesep}{1pt}
\setlength{\belowrulesep}{1pt}

\begin{tabular}{lcccccccccc}
\toprule
& \textbf{Retrain} & \textbf{Initial model} & \textbf{IU} & \textbf{ReprUnl} & \textbf{LTD}
& \textbf{SalUn} & \textbf{FT} & \textbf{NegGrad+} & \textbf{AMUN} & \textbf{RL} \\
\midrule

Feature Discrepancy
& \cellcolor{nearGreen}0.000
& \cellcolor{nearGreen!70!warmOrange}0.049
& \cellcolor{nearGreen!69!warmOrange}0.050
& \cellcolor{nearGreen!50!warmOrange}0.079
& \cellcolor{nearGreen!38!warmOrange}0.099
& \cellcolor{nearGreen!16!warmOrange}0.134
& \cellcolor{nearGreen!13!warmOrange}0.139
& \cellcolor{nearGreen!10!warmOrange}0.144
& \cellcolor{nearGreen!6!warmOrange}0.151
& \cellcolor{warmOrange}0.160
\\

Mean Accuracy (\%)
& \cellcolor{nearGreen}\textbf{90.29}
& \cellcolor{nearGreen!96!warmOrange}\textbf{89.71}
& \cellcolor{nearGreen!96!warmOrange}\textbf{89.69}
& \cellcolor{nearGreen!68!warmOrange}\textbf{85.61}
& \cellcolor{nearGreen!68!warmOrange}\textbf{85.49}
& \cellcolor{nearGreen!33!warmOrange}\textbf{80.40}
& \cellcolor{nearGreen!28!warmOrange}\textbf{79.62}
& \cellcolor{nearGreen!61!warmOrange}\textbf{84.44}
& \cellcolor{nearGreen!17!warmOrange}\textbf{78.00}
& \cellcolor{warmOrange}\textbf{75.46}
\\
\bottomrule
\end{tabular}
\end{table*}

\subsection{Related Work}
\textbf{Representation-level unlearning.} 
Recent work goes beyond output-level evaluation.
\citet{almudevar2026representation} and \citet{sepahvand2025selective} aim to suppress forget-set information in learned representations. Others show that output-level forgetting can leave representations close to the full model \citep{kim2026we}, retain forgotten-class information \citep{gao2026illusion}, or differ from retraining despite output agreement \citep{yong2026erasedgoneoutputforgetting}. We distinguish missing retraining features from unmatched features in the unlearning model. In particular, we ask which features retraining selects, whether unlearning algorithms recover them, and whether missing them affects downstream adaptation. Our analysis builds on the model-defined feature framework of \citet{jiang2024on}; we discuss connections to representation-similarity measures in Appendix~\ref{app:cka}.

\textbf{Post-unlearning adaptation and relearning.}
Prior work studies whether forgotten behavior can re-emerge under subsequent optimization \citep{hu2025unlearning,xu2026unlearning,siddiqui2025from}%
  or whether apparently forgotten class information remains recoverable from learned representations~\citep{dehghani2026source, jang2026suppression}.
\citet{siddiqui2025from}, in particular, relates relearning after retain-set fine-tuning to weight-space displacement from the full-data model. We instead ask whether unlearning recovers the features selected by retraining and whether failure to do so leads to divergence from retraining under subsequent adaptation, theoretically under head adaptation and empirically under relearning and downstream transfer.

Additional work on unlearning guarantees, methods, and evaluation is discussed in Appendix~\ref{app:related-work}.

\section{Preliminaries}\label{sec:prelim_features}

\textbf{Machine Unlearning Setup.} Let $D=D_r\sqcup D_f$ be the training set, partitioned into a retain set $D_r$ and a forget set $D_f$. For a possibly randomized training algorithm $\mathcal A$, let
\[
\theta_{\mathrm{full}}\sim\mathcal A(D),
\qquad
\theta_{\mathrm{retr}}\sim\mathcal A(D_r),
\qquad
\theta_{\mathrm{unl}}
\sim
\mathcal U(\theta_{\mathrm{full}},D_f,D_r),
\]
where $\mathcal U$ is an unlearning method. The ideal unlearning target is to match the distribution of $\theta_{\mathrm{retr}}$. In practice, approximate unlearning is often evaluated by the weaker
criterion of output agreement with retraining. We study whether this agreement extends to the feature solution selected by retraining, and whether feature-level differences become observable under subsequent model use.

\textbf{Model Representations and Features.}
We follow the model-defined feature framework of \citet{jiang2024on}. For a model $f_\theta(x)=W_\theta^\top\phi_\theta(x)$ and a comparison set $\mathcal X=\{x_1,\ldots,x_N\}$, let $\Phi_\theta=[\phi_\theta(x_1),\ldots,\phi_\theta(x_N)]^\top$ be the representation matrix, and let
$\widetilde{\Phi}_\theta=(I_N-\frac{1}{N}\mathbf 1\mathbf 1^\top)\Phi_\theta$
be its column-centered version. Write its Singular Value Decomposition as
\[
\widetilde{\Phi}_\theta=U_\theta\Sigma_\theta V_\theta^\top,
\qquad
U_\theta=[u_{\theta,1},u_{\theta,2},\ldots],
\qquad
\lambda_{\theta,i}=(\Sigma_\theta)_{ii}^2,
\]
The left singular vector $u_{\theta,i}$ gives the normalized response pattern of the $i$th PCA direction across $\mathcal X$, with strength $\lambda_{\theta,i}$.

Let $\mathcal R$ be a set of independently trained models obtained under the same training setup. From each model $\theta\in\mathcal R$, we take the top-$K$ PCA response patterns and cluster them jointly by similarity, yielding clusters $C_1,\ldots,C_{J_{\mathcal R}}$. Each cluster $C_t$ defines a \emph{feature}, represented by its sign-aligned response centroid $c_t$ and average strength \(w_t=|C_t|^{-1}\sum_{(\theta,i)\in C_t}\lambda_{\theta,i}\). We collect these strengths in $w_{\mathcal R}=(w_1,\ldots,w_{J_{\mathcal R}})^\top$. Details of the clustering and centroid construction are given in Appendix~\ref{app:feature-construction}.

\section{Why Feature Mismatch Matters Under Adaptation}
\label{sec:feature-mismatch-theory}

Empirical unlearning evaluations are typically static: they compare the unlearned and retrained models immediately after unlearning. In practice, however, an unlearned model may later be fine-tuned, trained on additional data, or adapted to a downstream task. We ask when predictive agreement between unlearned and retrained models remains stable under such updates.

We begin with a lightweight adaptation regime in which the learned representation is frozen and only the prediction head is updated. Such updates arise naturally in relearning attacks and in downstream adaptation with a frozen pretrained backbone. Proofs of all results in this section are in Appendix~\ref{app:harmlessness}.

Consider retrained and unlearned models with \(f_{\mathrm{retr}}(x), f_{\mathrm{unl}}(x)\in\mathbb{R}^C\),
\[
    f_{\mathrm{retr}}(x)
    =
    W_{\mathrm{retr}}^\top\phi_{\mathrm{retr}}(x),
    \qquad
    f_{\mathrm{unl}}(x)
    =
    W_{\mathrm{unl}}^\top\phi_{\mathrm{unl}}(x).
\]
Fix a labeled dataset \(\{(x_i,y_i)\}_{i=1}^N\) with input set \(\mathcal X=\{x_1,\ldots,x_N\}\), and suppose the models are initially predictively equivalent on \(\mathcal X\),
\begin{align}~\label{eq:output-equiv}
W_{\mathrm{retr}}^\top\phi_{\mathrm{retr}}(x)
    =
    W_{\mathrm{unl}}^\top\phi_{\mathrm{unl}}(x)
    \qquad \forall x \in \mathcal X.
\end{align}
For a differentiable loss $\ell(z,y)$ applied to the prediction $z=f(x)$, define for each $(x_i,y_i)$
\[
g_i
:=
\nabla_z \ell(z,y_i)\big|_{z=f_{\mathrm{retr}}(x_i)}
=
\nabla_z \ell(z,y_i)\big|_{z=f_{\mathrm{unl}}(x_i)} \in\mathbb{R}^C,
\]
where the equality follows from predictive equivalence. Updating only the head on $(x_i,y_i)$ with step size $\eta$ gives
\[
W_{m,i}^+
=
W_m-\eta\,\phi_m(x_i)g_i^\top,
\qquad
m\in\{\mathrm{retr},\mathrm{unl}\}.
\]
Let \(F_{m,i}^+\in\mathbb{R}^{N\times C}\) collect the predictions on \(\mathcal X\) after this update, and let \(\Phi_m\) and \(\widetilde{\Phi}_m\) denote the uncentered and centered representation matrices on \(\mathcal X\), respectively.

The result below shows that predictive equivalence can be fragile under even a single head update.

\begin{proposition}~\label{prop:harmless-quantity}
Suppose that $g_i\neq 0$ for all $i$ and that
$f_{\mathrm{retr}}(x)=f_{\mathrm{unl}}(x)$ for all $x\in\mathcal X$.
Then
\[
    \sum_{i=1}^N
    \frac{\|F_{\mathrm{retr},i}^+-F_{\mathrm{unl},i}^+\|_F^2}
    {\eta^2 \|g_i\|_2^2}
    =
\left\|
\Phi_{\mathrm{retr}}\Phi_{\mathrm{retr}}^\top
-
\Phi_{\mathrm{unl}}\Phi_{\mathrm{unl}}^\top
\right\|_F^2
    \geq
    \left\|
    \widetilde{\Phi}_{\mathrm{retr}}\widetilde{\Phi}_{\mathrm{retr}}^\top
    -
    \widetilde{\Phi}_{\mathrm{unl}}\widetilde{\Phi}_{\mathrm{unl}}^\top
    \right\|_F^2.
\]
\end{proposition}
The proposition quantifies how representation mismatch can be exposed by even a single head update. The centered Gram matrix decomposes as
\(
\widetilde{\Phi}\widetilde{\Phi}^{\top}
=
\sum_j \lambda_j u_j u_j^\top,
\)
so discrepancies in it reflect differences in both the response directions and strengths of model-defined features. As a special case, if output equivalence on \(\mathcal X\) is preserved after updating on each \((x_i,y_i)\), \(i\in[N]\), the centered Gram matrices must coincide, and the two models therefore have the same feature structure on $\mathcal X$, up to the usual eigenspace ambiguities. 

More generally, larger centered Gram mismatch implies larger normalized prediction disagreement after adaptation. The bound is one-sided: small centered mismatch does not rule out additional disagreement from components removed by centering. When the adaptation gradient is non-negligible, this lower bound translates centered Gram mismatch into a non-negligible prediction gap, which is particularly relevant for relearning on the forget set.

\textbf{A fixed-head perspective.}
The analysis above considers output equivalence between complete retrained and unlearned models. Alternatively, if unlearning is viewed as removing forget-set information from the model, the representation and prediction head may be judged separately rather than required to jointly match the same retrained model. Under this view, a retrained representation $\phi_{\mathrm{retr}}$ is indistinguishable from $\phi_{\mathrm{unl}}$ under the current unlearned head if 
\[
    W_{\mathrm{unl}}^\top
    \bigl(
        \phi_{\mathrm{retr}}(x)-\phi_{\mathrm{unl}}(x)
    \bigr)
    =0
    \qquad \forall x \in \mathcal X.
\]
Such a difference is ``harmless'' in the sense of \citet{grant2026addressing}. In Appendix~\ref{app:harmlessness-shared-head}, we show that this harmlessness is generally unstable: once the head is adapted, previously invisible representation differences can affect the predictions.

\section{Data Deletion Can Require Learning New Features}
\label{sec:feature-substitution}

The previous section shows why agreement with the retrained features matters under subsequent model use. We now ask how unlearning changes that solution relative to the full-data model. A common view of unlearning is that it is a subtractive operation: forgetting data should eliminate or weaken features that depended on those data, so recovering retraining would mainly require suppressing features already present in the full model. We show that deletion can instead induce a genuinely different feature solution. Even an arbitrarily small deletion can cause retraining to learn features that are absent in the full-data model. Thus, matching retraining may require more than simply erasing existing features: an unlearning method may need to \emph{learn new ones}.

We study this phenomenon in a minimal model with a linear shortcut and a classical XOR feature. The construction captures a setting where the shortcut is predictive across the data, while the XOR signal differs across subpopulations. Removing even a small group can therefore change which feature is favored by training. Exact trajectories and proofs are deferred to Appendix~\ref{app:xor}.

\textbf{Setting.}
Let $\varepsilon\in(0,1/2)$ be the forget-set fraction and $c>0$ control the strength of the shortcut feature. With a slight abuse of notation, we write the input as the triplet $x=(s,\tilde z)\in\mathbb R^3$, where $s\in\mathbb R$ and $\tilde z\in\mathbb R^2$, with label $y\in\{-1,1\}$. The data distribution is a mixture of two distributions \(D_r\) and \(D_f\) with weights \(1-\varepsilon\) and \(\varepsilon\), respectively:
\[
\begin{array}{c|c|ccc}
  & y & s & \tilde z \\ \hline
D_r  & \chi(z) & cy & \sqrt{\varepsilon}\,z \\
D_f  & -\chi(z) & cy & \sqrt{1-\varepsilon}\,z
\end{array}
\qquad
z\sim\mathrm{Unif}(\{-1,1\}^2),\quad
\chi(z)=z_1z_2.
\]
To capture the mechanisms of both the shortcut and XOR features while keeping the training dynamics of the unlearning algorithms analytically tractable, we use a quadratic network,
\[
    f_\theta(s,\tilde z)
    =
    as
    + \frac12 (u^\top \tilde z)^2
    - \frac12 (v^\top \tilde z)^2,
\]
trained with logistic loss under full-batch gradient flow from $a(0)=0$, $u(0)=v(0)=\delta e_1$, $\delta>0$. The linear term captures the shortcut, while the quadratic terms can represent the XOR feature, since $z_1z_2$ is a difference of two squared linear forms. 

The shortcut $s=cy$ is predictive in both groups \(D_r\) and \(D_f\). The tuple $\tilde z$ carries a classical XOR feature: $\tilde z_1\tilde z_2$ agrees with $y$ on $D_r$ and is reversed on $D_f$. In the full distribution, \(\tilde z\) is scaled so that these contributions cancel, leaving the shortcut as the predictive signal, whereas after deleting $D_f$ the XOR feature becomes predictive on $D_r$. We note that despite this rescaling, the groups have comparable input norms for large \(c\), differing by at most a factor \(\sqrt{1+2c^{-2}}\). With \(D_f\) an \(\varepsilon\)-fraction of the data, the construction covers arbitrarily small deletions.

For feature analysis, we use the corresponding three-dimensional feature representation
{
\setlength{\abovedisplayskip}{4pt}
\setlength{\belowdisplayskip}{4pt}
\setlength{\abovedisplayshortskip}{4pt}
\setlength{\belowdisplayshortskip}{4pt}
\[
    \phi_\theta(x)
    =
    \bigl(
        as,\,
        (u^\top\tilde z)^2,\,
        (v^\top\tilde z)^2
    \bigr)^\top
\]
}%
Below, we analyze full-data training, retraining, and three unlearning dynamics: Random Labels (RL), which optimizes on randomly relabeled forget examples; RL combined with fine-tuning on the retain set (RL+FT)~\citep{WarneckePirchWressnegger2023_1000166047}; and Gradient Ascent (GA)~\citep{9797378}, which maximizes the original loss on the forget data. Along these trajectories, the model-defined features have the same identities and are associated with the same PCA directions,
{
\setlength{\abovedisplayskip}{4pt}
\setlength{\belowdisplayskip}{4pt}
\setlength{\abovedisplayshortskip}{4pt}
\setlength{\belowdisplayshortskip}{4pt}
\[
q_{\mathrm{s}}=(1,0,0)^\top,\qquad
q_{+}=\frac{1}{\sqrt{2}}(0,1,1)^\top,\qquad
q_{-}=\frac{1}{\sqrt{2}}(0,1,-1)^\top,
\]}%
with varying strengths. For trajectory $m$, we denote these strengths at time $t$ by $\lambda_{\mathrm{s}}^{m}(t)$, $\lambda_{+}^{m}(t)$, and $\lambda_{-}^{m}(t)$.

We now compare the features selected by full-data training and by retraining on $D_r$. Although the forget set has a mass of only $\varepsilon$, the two procedures converge to qualitatively different solutions.
\begin{theorem}
\label{thm:xor-substitution}
As training time $t\to\infty$, 
\begin{itemize}[leftmargin=1em]
    \item full-data training satisfies  \(\frac{\lambda_{\mathrm{s}}^{\mathrm{full}}(t)}
         {\lambda_+^{\mathrm{full}}(t)}
    \to \infty\)  and \(    \lambda_-^{\mathrm{full}}(t)=0 \quad\text{for all } t \ge 0\).
    \item under retraining on \(D_r\),  \(\frac{\lambda_{\mathrm{s}}^{\mathrm{retr}}(t)}
         {\lambda_+^{\mathrm{retr}}(t)}
    \to 0,~\frac{\lambda_{\mathrm{s}}^{\mathrm{retr}}(t)}
         {\lambda_-^{\mathrm{retr}}(t)}
    \to 0,\text{and}~\frac{\lambda_+^{\mathrm{retr}}(t)}
     {\lambda_-^{\mathrm{retr}}(t)}
\to (1-2\varepsilon)^2.\)
\end{itemize}
\end{theorem}

Theorem~\ref{thm:xor-substitution} shows that deletion can require more than reweighting features learned by the full model. Under retraining, the shortcut becomes asymptotically negligible, while a core feature $q_{-}$ absent from the full-data solution emerges. Thus, matching retraining will require learning a feature that the full model never learned.

\textbf{Unlearning Can Miss The Retrained Feature Solution.}
We now ask whether unlearning, initialized from the full-data model, can reproduce the above feature substitution induced by retraining. We first consider Random Labels (RL) and its version with retain-set fine-tuning (RL+FT).

\begin{theorem}
\label{thm:xor-rl}
As the unlearning time $t\to\infty$, for
$M\in\{\mathrm{RL},\mathrm{RL+FT}\}$,
\[
    \frac{\lambda_{\mathrm{s}}^{M}(t)}
         {\lambda_-^{M}(t)}
    \longrightarrow
    \frac{1-\varepsilon}{2\varepsilon}>\frac12,
\]
while the corresponding ratio converges to zero under retraining, as shown in Theorem~\ref{thm:xor-substitution}.
\end{theorem}
For small $\varepsilon$, the ratio $\frac{1-\varepsilon}{2\varepsilon}$ is large, so RL and RL+FT remain shortcut-dominated like the full-data solution, unlike retraining where this ratio vanishes. Even combined with retain-set fine-tuning, RL does not recover the retraining feature transition. Appendix~\ref{app:xor} further shows that RL generally does not recover the retraining ratio between the two core strengths, whereas RL+FT does.

Next, we analyse Gradient Ascent (GA). We show that GA behaves differently: it eventually recovers the retraining-like feature solution, but requires nearly as much optimization as retraining from scratch. Train the full-data and retrained models for the same duration \(T\). Let \(\tau_{\mathrm{GA}}(T)\) be the first time gradient ascent, initialized from the full-data model, matches the retrained model’s feature-strength ratio \(\lambda_s/\lambda_-\).

\begin{theorem}\label{theorem:ga-time-main} For every \(\zeta\in (0,1)\), if the initialisation \(\delta>0\) is sufficiently small or the shortcut strength \(c\) is sufficiently large, keeping other parameters fixed, then
\[
\lim_{T\to\infty}\frac{\tau_{\mathrm{GA}}(T)}{T}\ge 1-\zeta\]
\end{theorem}

Thus, GA requires a non-vanishing fraction of the retraining time to recover the same feature balance, and this fraction can approach one. Together, these results expose two distinct difficulties for approximate unlearning: converging to the wrong feature solution, as RL does, or recovering the right one only with retraining-scale optimization, as can occur for GA.

The key point is that successful unlearning requires a specific feature transition, not arbitrary movement away from the full model. Some full-model features should vanish or weaken, while others may need to emerge. This motivates the directional comparisons in Section~\ref{sec:feature-recovery}: we measure both what unlearning retains from the full-model solution and what it fails to recover from retraining.

\section{Does Unlearning Recover Retrained Features?}
\label{sec:feature-recovery}

We now turn to neural networks and ask whether approximate unlearning recovers the specific feature solution selected by retraining. We first introduce a directional comparison between feature solutions and then use it to compare full-data, retrained, and unlearned models.

To compare feature solutions, we ask which features of a reference solution are reproduced by another model. Since model-defined features need not recur across random seeds \citep{jiang2024on}, we compare sets of independently trained or unlearned models rather than individual runs. Let $\mathcal R$ denote the reference model set and $\mathcal M$ the evaluated model set (e.g., retrained runs and runs of one unlearning method, respectively). We write $\{c_t\}_{t=1}^T$ and $w_{\mathcal R}$ for the feature clusters and strength vector of $\mathcal R$, as defined in Section~\ref{sec:prelim_features}. All comparisons below are computed on a shared input set, on which the resulting metrics depend; we suppress this dependence in the notation for readability.

For each reference feature cluster with centroid $c_t$, we ask how well it is matched by any of the top-$K$ PCA components in the evaluated group $\mathcal M$. We quantify its match quality by
\[
    \rho_t(\mathcal{R}, \mathcal M)
    =
    \max_{\theta\in\mathcal M,\;j\in[K]}
    \left|
        \operatorname{corr}(c_t,u_{\theta,j})
    \right|.
\]
We convert this match quality into a soft penalty,
\[
    d_t(\mathcal R,\mathcal M)
    =
    \sqrt{
        1-
        \min\{1,\rho_t(\mathcal{R}, \mathcal M)/\gamma\}^{2}
    },
\]
where $\gamma$ is a matching threshold derived from the correlation threshold used for feature clustering (App.~\ref{app:feature-construction}). Features matched at or above $\gamma$ incur no penalty, while weaker matches incur larger penalties. Collecting these penalties in $d(\mathcal R,\mathcal M)$, we define the directional feature discrepancy as 
\begin{align}
    D_{\mathrm{feat}}(\mathcal R,\mathcal M)
    =
    \frac{
        \left\|
            w_{\mathcal R}
            \odot
             d(\mathcal R,\mathcal M)
        \right\|_2
    }{
        \|w_{\mathcal R}\|_2
    }.
\end{align}
Thus, intuitively, the metric asks whether each feature of the reference solution has a recognizable counterpart in $\mathcal M$. If some response in $\mathcal M$ resembles that feature, its contribution is small; if nothing does, its contribution is large, with stronger reference features weighted more heavily. The metric therefore provides a soft, strength-weighted measure of how much of the reference feature solution is missing from $\mathcal M$. 

The particular form of the discrepancy is motivated by the centered Gram-matrix discrepancy in Section~\ref{sec:feature-mismatch-theory}. For a single model, \(\widetilde\Phi\widetilde\Phi^\top = \sum_i \lambda_i u_i u_i^\top \). Consider first the idealized case where a set $\mathcal Q$ of reference features is absent from the other solution, with their response directions orthogonal to those represented there. By orthonormality, their contribution to the squared Frobenius discrepancy is \looseness=-1
\[
\|
    \sum_{i\in\mathcal Q}\lambda_i u_i u_i^\top
\|_F^2
=
\sum_{i\in\mathcal Q}\lambda_i^2.
\]
This motivates weighting each feature-recovery error by its strength and aggregating these errors in $L_2$. For partially matched features, the discrepancy between two unit response directions satisfies
\[
\|uu^\top-vv^\top\|_F^2
=
2\bigl(1-(u^\top v)^2\bigr).
\]
This motivates the squared-correlation dependence in $d_t$. We rescale the correlation by the matching threshold $\gamma$, so that sufficiently well-matched directions incur no penalty. Thus, $w_t$ captures the strength of a reference feature, while $d_t$ captures how well its response direction is recovered.

\begin{figure}[t]
    \centering
    \begin{subfigure}{0.33\linewidth}
        \centering
        \includegraphics[width=\linewidth]{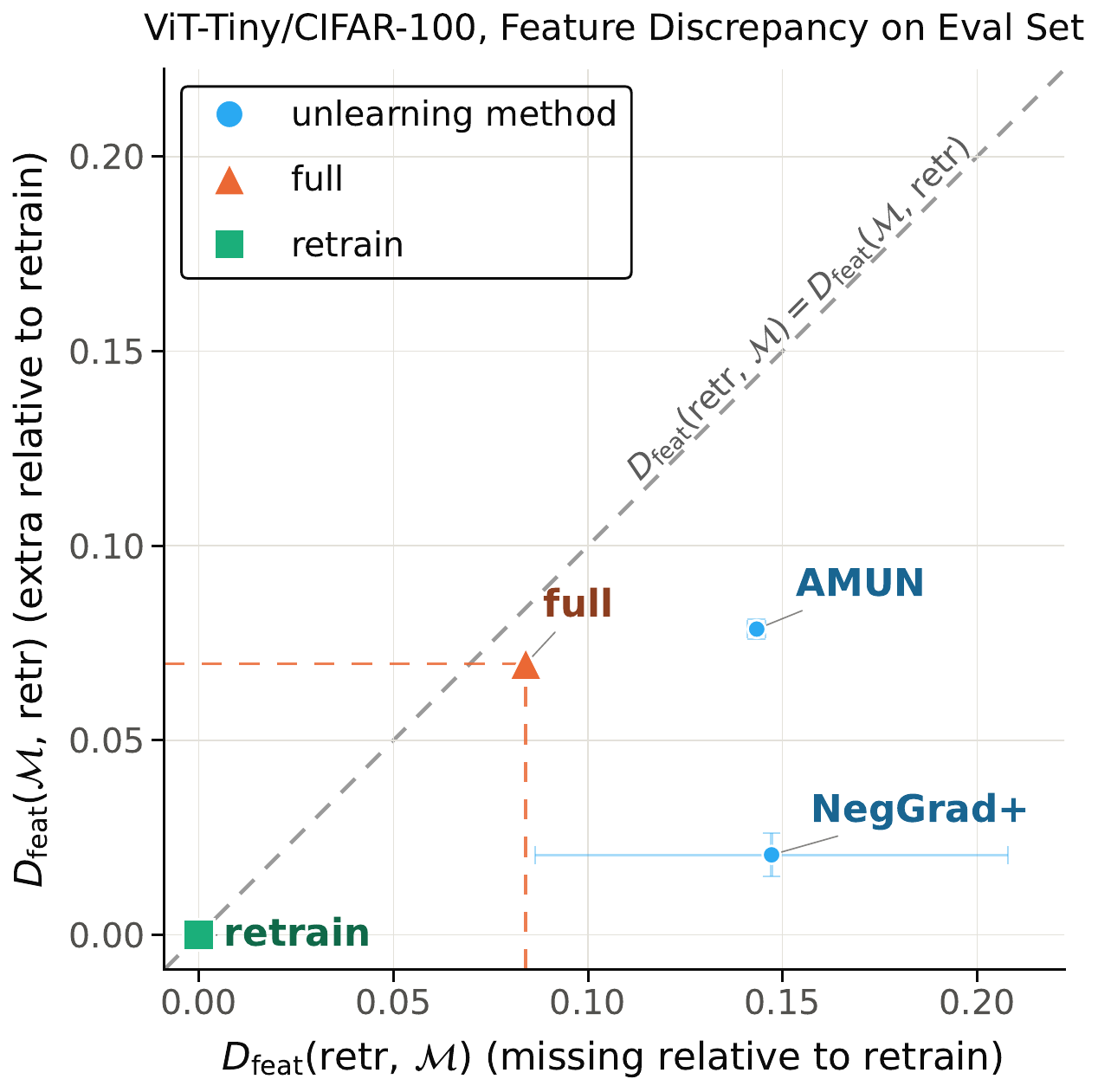}
        \label{fig:feature-metric}
    \end{subfigure}
    \hfill
    \begin{subfigure}{0.64\linewidth}
        \centering
        \includegraphics[width=\linewidth]{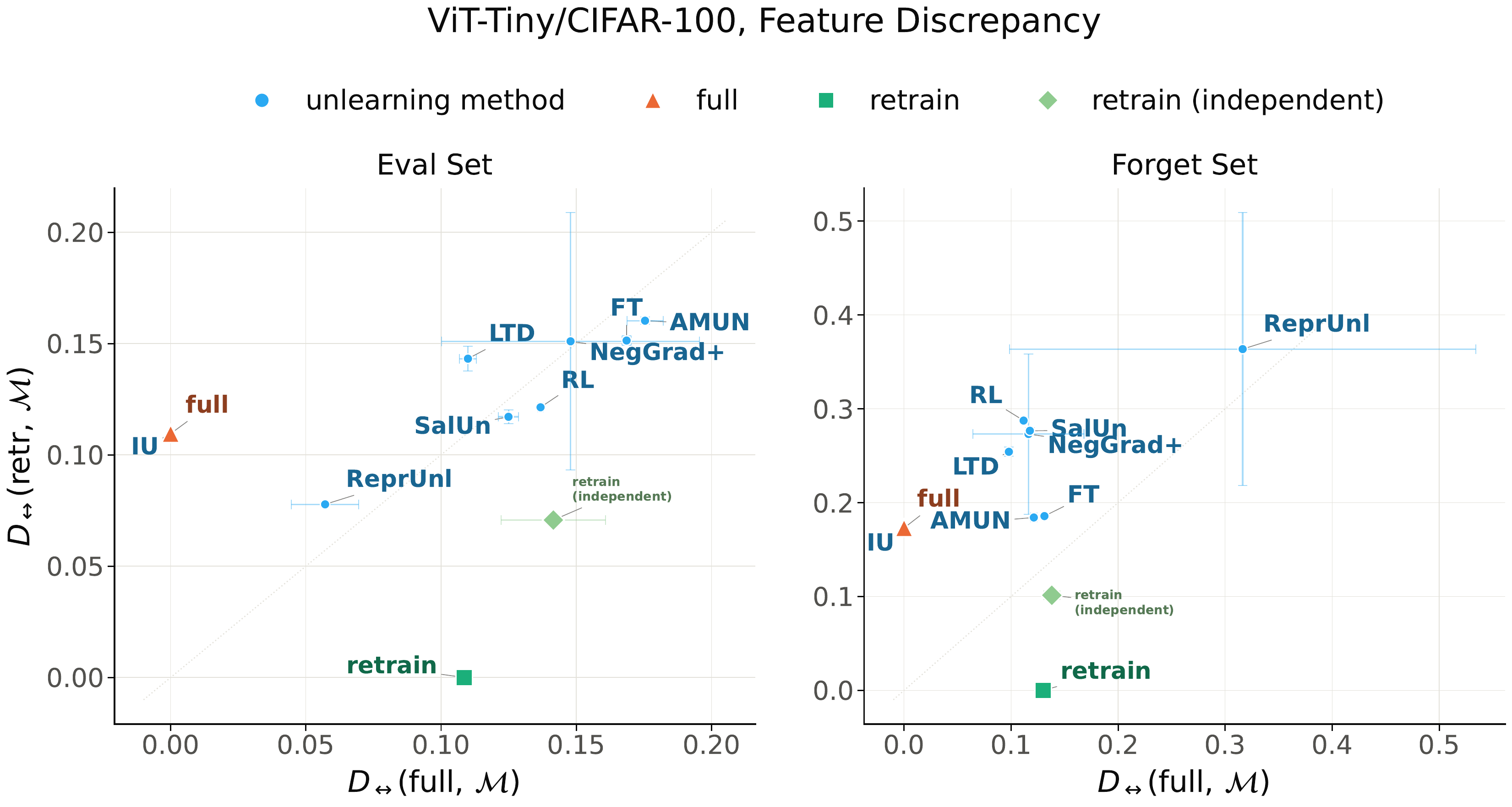}
        \label{fig:coverage-cifar100}
    \end{subfigure}\vspace{-4mm}
    \caption{
    The left panel illustrates that retraining moves away from some full-data features while favoring others that were previously less well represented; for NegGrad+, incomplete recovery of retraining features dominates the mismatch, consistent with over-suppression. The middle and right panels show the two-sided feature discrepancy from the full-data and retrained model sets on the test and forget sets, respectively.}
    \label{fig:feature-recovery}
    \vspace{-3mm}
\end{figure}

Because $D_{\mathrm{feat}}(\mathcal R,\mathcal M)$ scores only features of the reference solution, it is directional. To capture mismatch in both directions, we define
\begin{align}
    D_{\leftrightarrow}(\mathcal R,\mathcal M)
    =
    \sqrt{
        \frac{
            \|w_{\mathcal R}\|_2^2
            \left(
                D_{\mathrm{feat}}(\mathcal R,\mathcal M)
            \right)^2
            +
            \|w_{\mathcal M}\|_2^2
            \left(
                D_{\mathrm{feat}}(\mathcal M,\mathcal R)
            \right)^2
        }{
            \|w_{\mathcal R}\|_2^2
        }
    }.
\end{align}
The first term captures reference features missing from $\mathcal M$, while the second captures features of $\mathcal M$ that are absent from the reference solution. We combine the two contributions quadratically, mirroring the structure of the Gram-matrix discrepancy, and normalize by the total feature strength of the reference solution, which is fixed across methods and therefore does not affect their ordering. Thus, $D_{\leftrightarrow}$ measures two-sided feature mismatch relative to the reference solution.

Although motivated by the Gram-matrix discrepancy, our metric isolates feature recovery rather than full Gram-matrix mismatch. Once a feature is well matched, strength differences incur no additional penalty, and the metric does not account for how frequently that feature recurs across models within a group. We analyze feature-strength differences separately for matched features in Appendix~\ref{app:additional-results}.

For unlearning, we compare each model group with both the full-data and retrained feature solutions. Discrepancy from retraining is our primary measure of feature recovery, while discrepancy from the full-data solution measures how much the feature solution has changed during unlearning. Together, the two comparisons distinguish methods that preserve the full-data solution from those that change it without recovering the solution selected by retraining.

\textbf{Feature Change Does Not Imply Feature Recovery.} We study ViT-Tiny on CIFAR-100, removing one fine-grained class from each of the 20 superclasses ($20\%$ of the training data). Each model set uses 20 seeds; feature results are averaged over three repetitions. We evaluate eight unlearning baselines: fine-tuning (FT), Random Labels (RL)~\citep{WarneckePirchWressnegger2023_1000166047}, NegGrad+~\citep{kurmanji2023towards}, SalUn~\citep{fan2024salun}, AMUN~\citep{ebrahimpourboroojeny2025amunadversarialmachineunlearning}, Influence Unlearning (IU)~\citep{pmlr-v70-koh17a,jia2023model}, LTD~\citep{dolgova2026forgettingneighborslocalizedcollateral}, and Representation Unlearning (ReprUnl)~\citep{almudevar2026representation}, which explicitly modifies learned representations. Most perform well under standard post-unlearning metrics, allowing us to ask whether they also recover the retrained feature solution. Results for ResNet-56, smaller deletion fractions, and SVHN are in Appendix~\ref{app:additional-results}, with standard unlearning metrics in Appendix~\ref{app:standard-metrics}.

Importantly, retraining changes the feature solution in both directions: some full-data features disappear, while others emerge only after deletion. In our main setting, the two directional discrepancies are comparable in magnitude (Figure~\ref{fig:feature-recovery}). This echoes Section~\ref{sec:feature-substitution}: matching retraining may require both suppressing existing features and recovering features absent from the full-data solution.

Figure~\ref{fig:feature-recovery} compares each method with the full-data and retrained feature solutions on the test and forget distributions. Methods with similar performance under standard unlearning metrics can nevertheless reach very different feature solutions. Some remain close to the full-data model, indicating feature persistence; others move far from it without approaching retraining, indicating off-target change. This spread is especially clear on the forget distribution, where methods range from strongly full-like to substantially displaced, yet none closely recovers retraining. Thus, similar behavioral success can mask qualitatively different internal outcomes. The same pattern holds across additional architectures, datasets, and deletion settings; see Appendix~\ref{app:additional-results}.

\vspace{-0.5em}
\section{Consequences of Feature Mismatch}
\label{sec:consequences}
The previous sections show that approximate unlearning can perform well on standard post-unlearning evaluations without recovering the feature solution selected by retraining. We now examine how this mismatch relates to behavior under subsequent adaptation.

Throughout this section, we freeze the learned representation, discard the original prediction head, and train a new randomly initialized head for the task of interest. This removes information specific to the old head and isolates what remains encoded in the representation. We also report standard unlearning metrics after training a fresh head on the retain set in Appendix~\ref{app:head-reset}; these generally improve, including for methods that did not initially match retraining.

\textbf{Feature Mismatch Predicts Relearning.} We first consider partial relearning. For each frozen representation, we train a fresh head on the retain set together with half of the forget examples, then evaluate on the full forget set. If forgotten information remains encoded in the representation, access to only part of the forget data may reveal it, producing behavior that differs from retraining under the same adaptation. We therefore evaluate feature discrepancy on the forget distribution and compare $D_{\leftrightarrow}(\mathrm{retr},\mathcal M)$ with the absolute forget-accuracy gap from retraining after partial relearning.

\begin{figure*}[t]
    \centering
    \begin{subfigure}{0.32\textwidth}
        \centering
        \includegraphics[width=\linewidth]{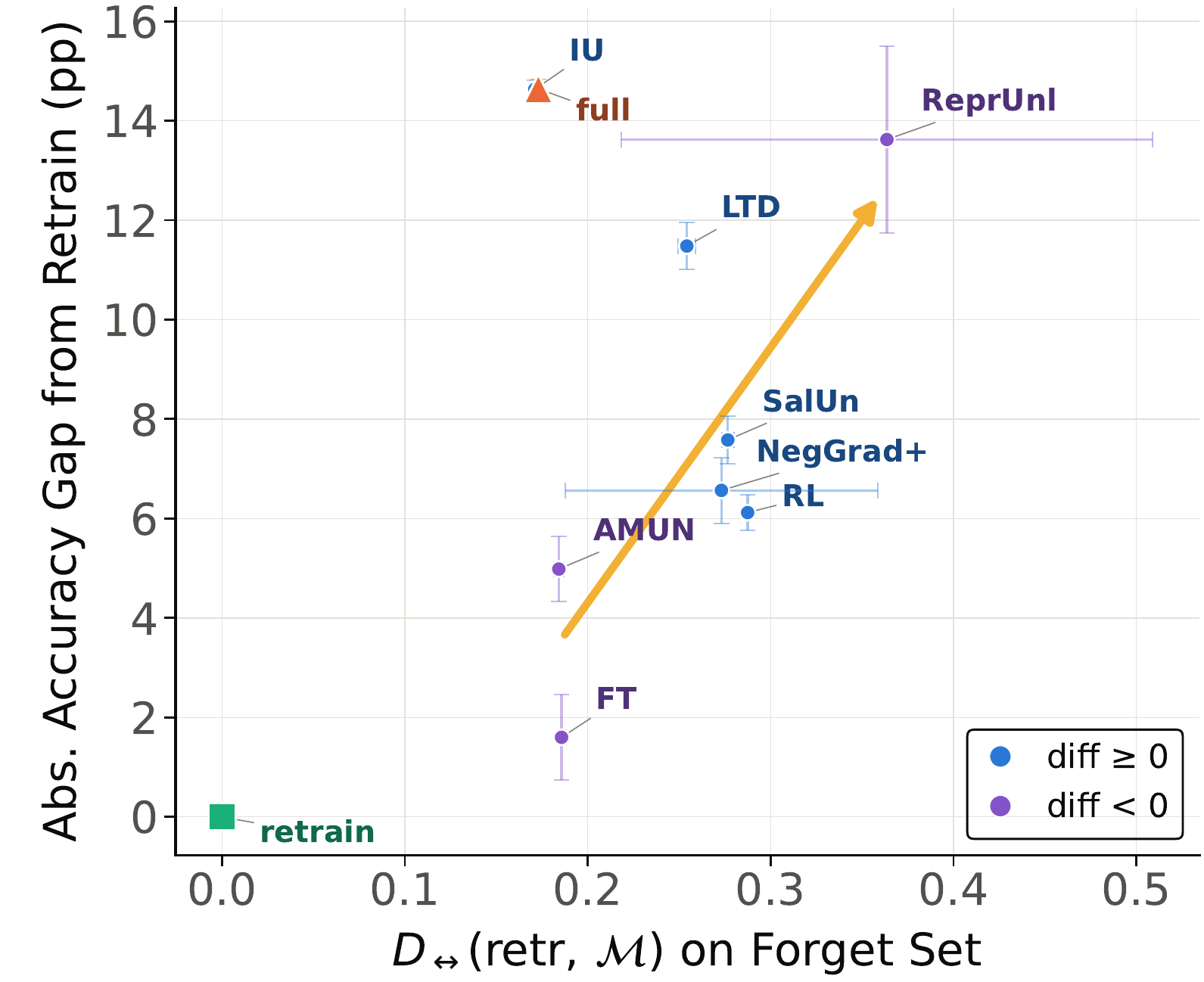}
        \caption{Partial relearning, Forget Set}
        \label{fig:feature-vs-halfforget}
    \end{subfigure}
    \hfill
    \begin{subfigure}{0.32\textwidth}
        \centering
        \includegraphics[width=\linewidth]{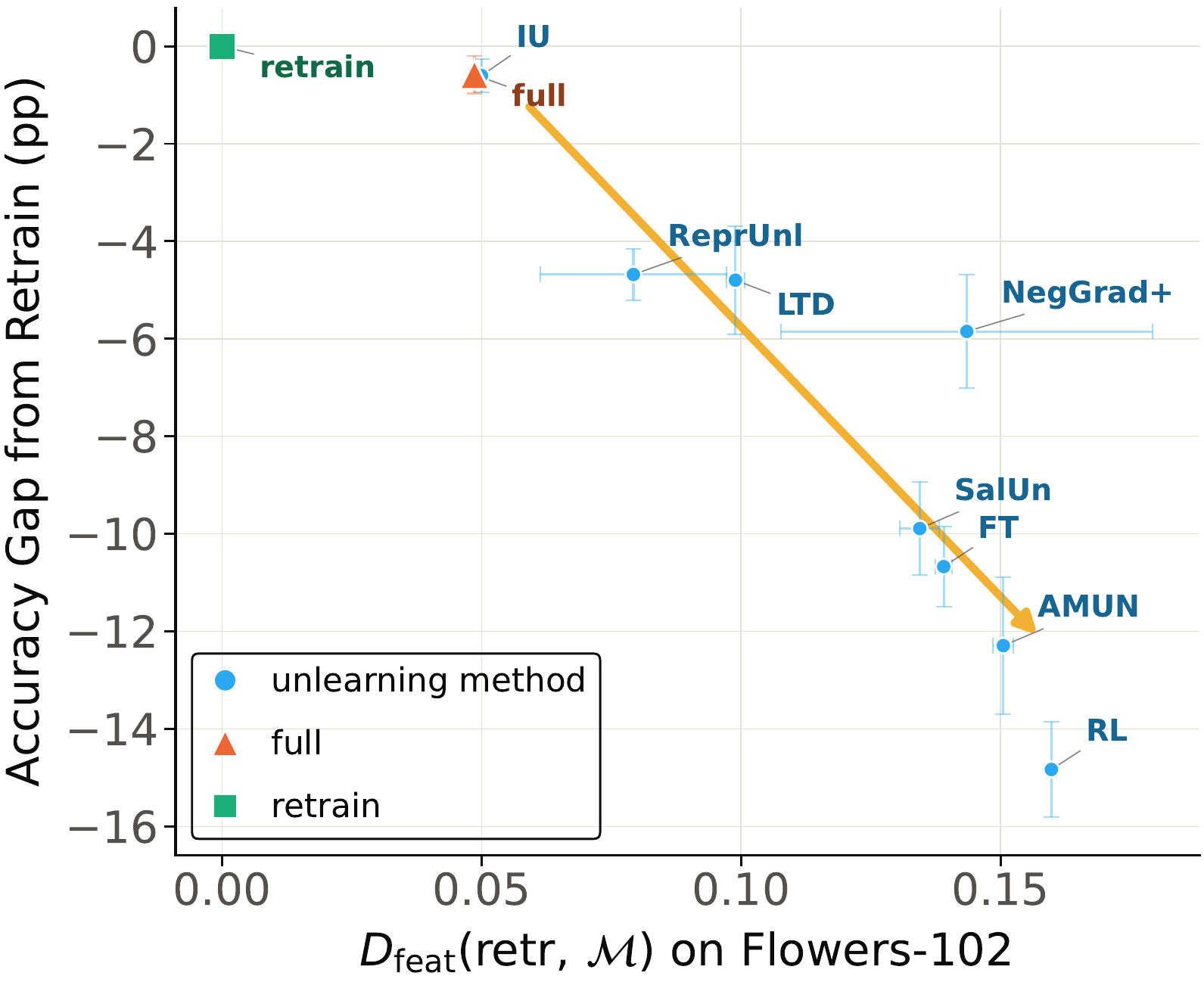}
        \caption{Flowers-102}
        \label{fig:feature-vs-flowers}
    \end{subfigure}
    \hfill
    \begin{subfigure}{0.32\textwidth}
        \centering
        \includegraphics[width=\linewidth]{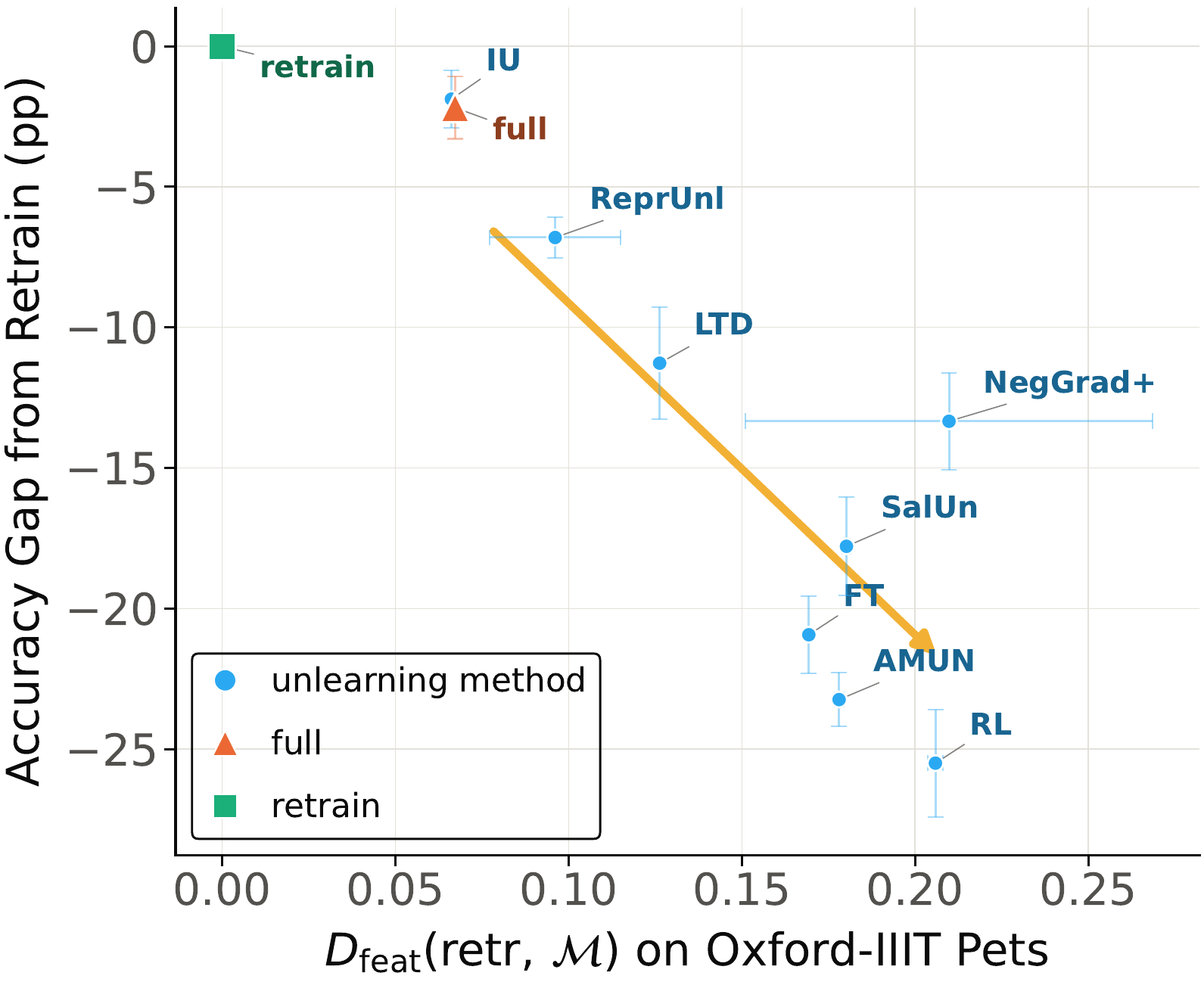}
        \caption{Oxford-IIIT Pets}
        \label{fig:feature-vs-pets}
    \end{subfigure}
    \caption{ViT-Tiny/CIFAR-100, Feature mismatch predicts behavior under subsequent adaptation. (a) Retraining-relative feature discrepancy on the forget distribution versus the absolute forget-accuracy gap after partial relearning; colors show the sign of the gap. (b--c) Missing retraining features on downstream inputs versus downstream accuracy gap on Flowers-102 and Oxford-IIIT Pets.\looseness=-1}
    \label{fig:feature-vs-adaptation}
    \vspace{-2mm}
\end{figure*}

Figure~\ref{fig:feature-vs-halfforget} shows a clear relationship: models farther from the retrained feature solution exhibit larger behavioral deviations after relearning. The full-data model and IU form a separate cluster: as Figure~\ref{fig:feature-recovery} shows, IU remains very close to the full-data feature solution. This is consistent with full-like representations retaining features that make the forgotten behavior easy to recover under partial relearning. Among the remaining methods, larger feature mismatch closely tracks larger deviation from retraining. The direction of the deviation varies across methods, but its magnitude grows with feature mismatch. Thus, representation differences that are hidden immediately after unlearning can be revealed by lightweight subsequent training, matching the mechanism identified in Section~\ref{sec:feature-mismatch-theory}.

\textbf{Missing Retraining Features Predict Downstream Loss.} 
Feature mismatch can also have the opposite consequence: an unlearned representation may lack capabilities that are retained by retraining. We test this by freezing each backbone and training a fresh head on two downstream classification datasets, Oxford-IIIT Pets~\citep{parkhi12a} and Flowers-102~\citep{Nilsback08}.

Here the directional discrepancy is the relevant quantity: unlearning may suppress features that retraining would retain, and we ask whether such missing features are associated with downstream degradation. On each downstream distribution, we measure \(D_{\mathrm{feat}}(\mathrm{retr},\mathcal M)\), capturing retraining features that are missing from the unlearned model on task-relevant inputs. We compare it with the downstream accuracy gap from retraining.

Across both tasks, missing retraining features closely track downstream performance loss: models with smaller directional discrepancy transfer similarly to retraining, while larger discrepancy corresponds to larger degradation.

\textbf{Correcting Feature Mismatch Moves Behavior Toward Retraining.}
The correlations above show that feature mismatch predicts subsequent behavior but do not establish that the identified mismatch is responsible for it. We therefore intervene directly on the features identified by our comparison: we restore retraining features missing from the unlearned solution and remove unlearned features that are absent from retraining. 

We apply these interventions at four increasing levels, progressively correcting the measured feature mismatch. After each intervention, we keep the modified representation fixed and train a fresh prediction head. We also include untargeted controls to distinguish correction of the identified mismatch from generic changes to the representation. Exact constructions are given in Appendix~\ref{app:feature-injection}.

\begin{figure*}[t]
    \centering
    \begin{subfigure}{0.32\textwidth}
        \centering
        \includegraphics[width=\linewidth]{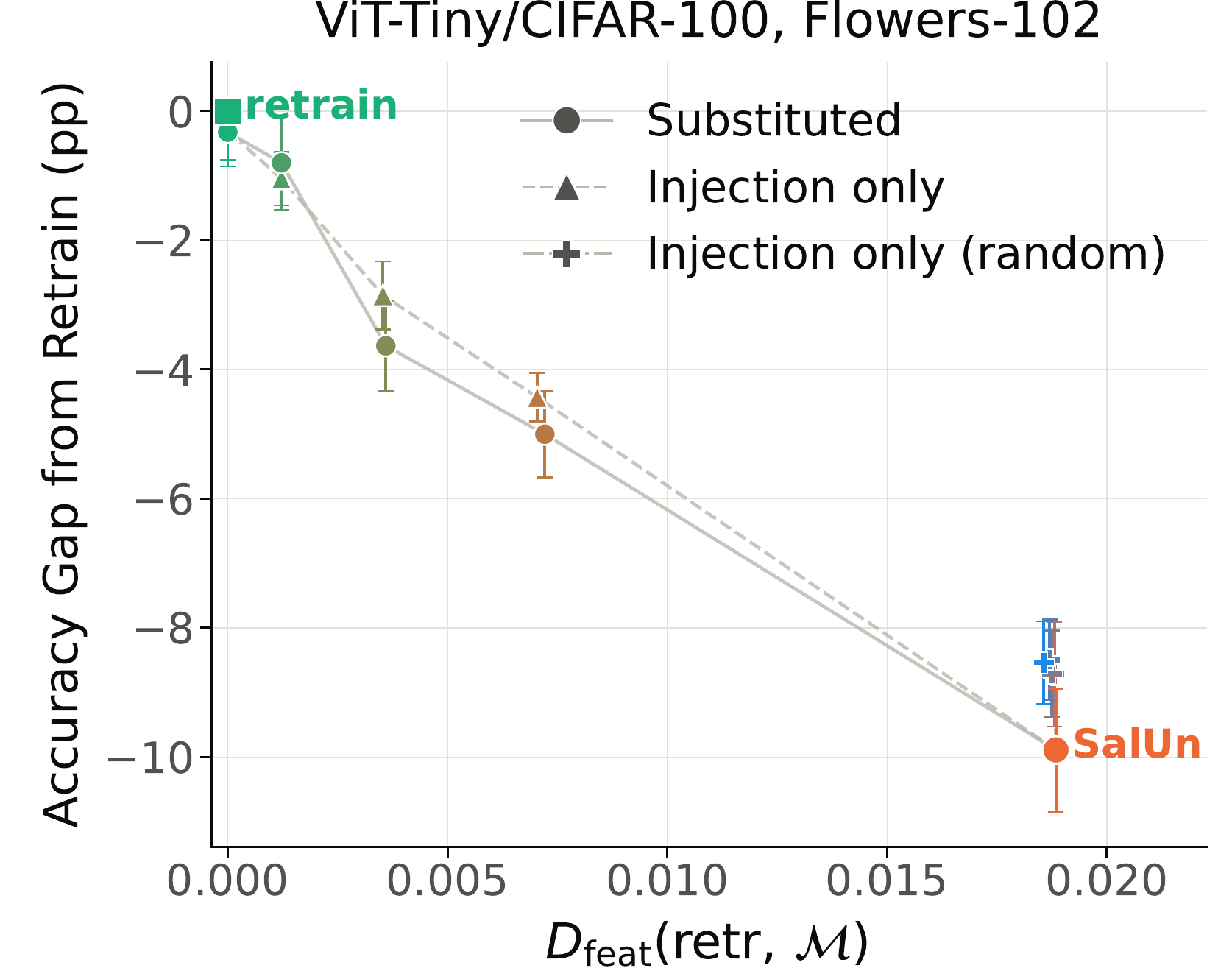}
        \label{fig:intervention-flowers}
    \end{subfigure}
    \hfill
    \begin{subfigure}{0.32\textwidth}
        \centering
        \includegraphics[width=\linewidth]{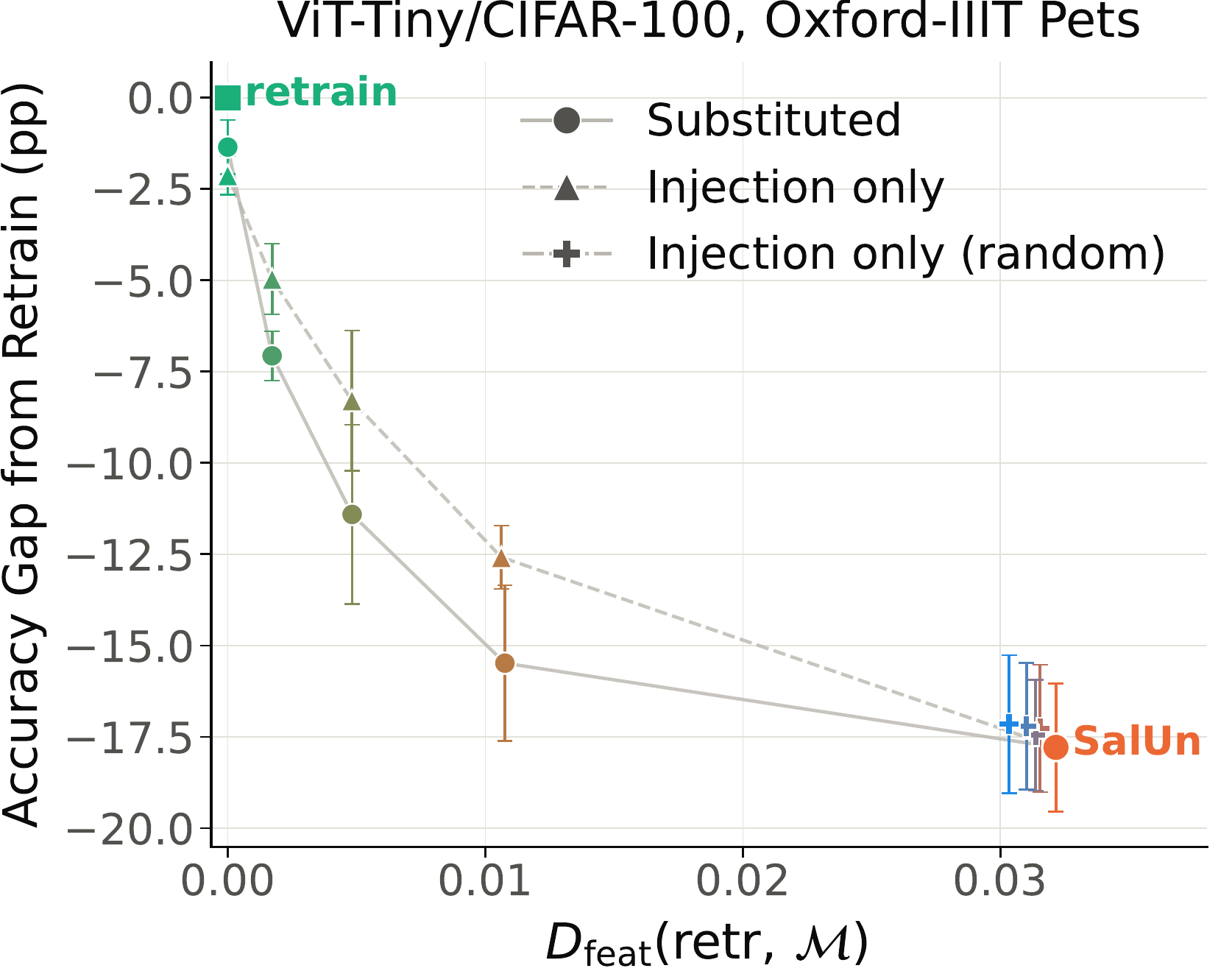}
        \label{fig:intervention-pets}
    \end{subfigure}
    \hfill
    \begin{subfigure}{0.32\textwidth}
        \centering
        \includegraphics[width=\linewidth]{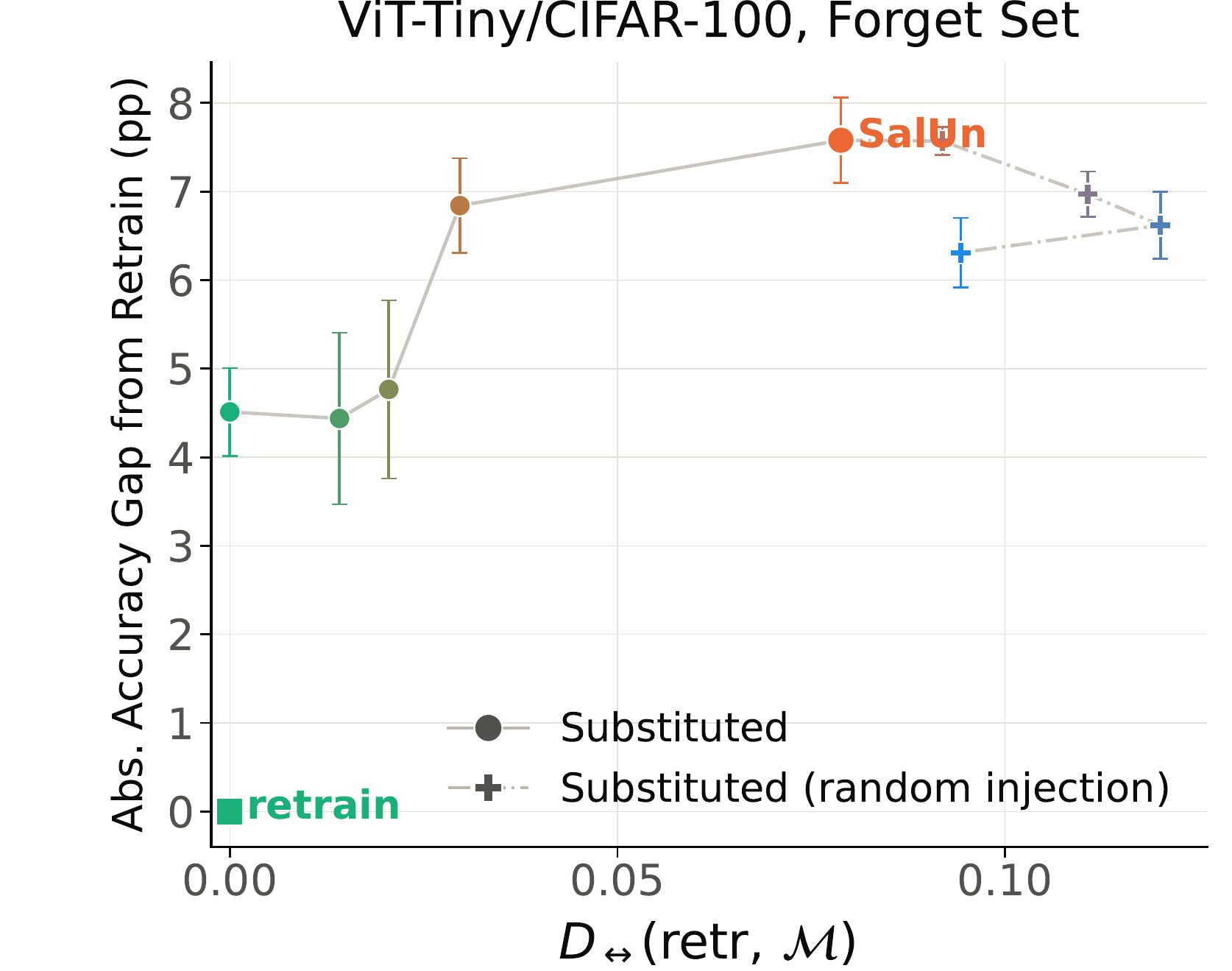}
        \label{fig:intervention-relearning}
    \end{subfigure}\vspace{-4mm}
    \caption{\textbf{Correcting retraining-relative feature mismatch moves behavior toward retraining.} For SalUn, restoring missing retraining features improves transfer, while additionally removing unmatched unlearned features further reduces the relearning gap.}
    \label{fig:feature-injection}
    \vspace{-3mm}
\end{figure*}

Figure~\ref{fig:feature-injection} shows a consistent movement toward retraining as the measured feature mismatch is corrected for SalUn (see Appendix~\ref{app:feature-injection-other} for other methods). On Flowers-102 and Oxford-IIIT Pets, progressively injecting missing retraining features recovers most of the downstream performance gap, while random injection does not produce the same effect. Under partial relearning, correcting the identified mismatch also substantially reduces the gap from retraining. A control that combines the same ablation with random feature injection also improves behavior, but to a smaller extent. This suggests that both removing unmatched features and recovering the specific features selected by retraining contribute to the behavioral change.

These interventions provide stronger evidence than correlation alone. The feature discrepancy does not merely identify representations that behave differently later: directly correcting the specific mismatch measured by our comparison systematically moves subsequent behavior toward retraining. Together with the predictive results above, this supports feature recovery as a behaviorally meaningful target for unlearning rather than merely a descriptive notion of representation similarity.

\section{Conclusion}
We study machine unlearning through feature recovery rather than static agreement with retraining. Taken together, our results suggest that successful unlearning is not only about reproducing retraining at the moment of deletion, but also about recovering a model state that behaves similarly during subsequent use. Output-equivalent models can encode different feature solutions, and these differences can become consequential once training resumes.

This is particularly relevant to continual unlearning, where the model produced after one deletion request becomes the starting point for the next. Feature differences left by one update may therefore matter for subsequent requests, even when the model initially agrees with retraining. This motivates evaluating unlearning not only immediately after deletion, but also under subsequent updates, with feature recovery as a complementary criterion.

\section*{Acknowledgements}
This work was supported by the Helmholtz Association’s Initiative and Networking Fund on the
HAICORE@FZJ partition. PD and SS gratefully acknowledge funding from the European Research Council (ERC) under the Horizon Europe Framework Programme (HORIZON) for proposal number 101170430 CollectiveMinds. Views and opinions expressed are however those of the authors only and do not necessarily reflect those of the European Union or the European Research Council. Neither the European Union nor the granting authority can be held responsible for them. AS also acknowledge support from VILLUM FONDEN via the Young Investigator program
(72069) and the Novo Nordisk Foundation for support via the Startup
grant (NNF24OC0087820).

\bibliography{iclr2027_conference}
\bibliographystyle{iclr2027_conference}

\appendix

\startcontents[appendix]

\newpage
\section*{Appendix Contents}
\printcontents[appendix]{}{1}{}{}

\newpage

\section{Additional Related Work}\label{app:related-work}

\citet{pmlr-v132-neel21a} distinguish statistical indistinguishability from retraining of the full maintained state from the weaker requirement of indistinguishability only at the observable output. Their setting concerns deletion guarantees for convex models; we instead study neural representations and ask when output agreement with retraining persists under subsequent head adaptation.

\paragraph{Approximate unlearning and evaluation.}
Approximate machine unlearning encompasses a broad range of practical approaches for modifying trained neural networks without retraining from scratch \citep{xu2023machineunlearningsurvey}. Representative methods include fine-tuning and relabeling \citep{WarneckePirchWressnegger2023_1000166047}, gradient-based forgetting \citep{kurmanji2023towards}, selective parameter updates \citep{fan2024salun}, adversarial unlearning \citep{ebrahimpourboroojeny2025amunadversarialmachineunlearning}, influence-based approximations~\citep{pmlr-v70-koh17a,jia2023model}, and representation-level objectives \citep{almudevar2026representation}. 

Evaluation is similarly diverse. Standard protocols report retain/test utility and forget-set behavior, often relative to a retrained reference. Privacy-oriented evaluations test whether deleted examples remain distinguishable, commonly through membership-inference attacks \citep{hayes2025inexact,tu2026reliable}, while other criteria assess residual forget-set information without explicit comparison to retraining \citep{chundawat2023can}. Other works also test the removal of specifically instilled undesirable behavior through incorrect labels or data poisoning~\citep{goel2022towards,goel2024corrective}.
Despite these different notions of successful deletion, evaluation remains predominantly snapshot-based, assessing the model immediately after unlearning rather than its agreement with retraining under subsequent use.

\section{Feature Construction Details}
\label{app:feature-construction}

\textbf{Correlation threshold.}
Following \citet{jiang2024on}, we determine the correlation threshold empirically from the reference ensemble. Let $\Lambda$ collect the Pearson correlations between every pair of PCA component responses from different models, evaluated on the same shared inputs. We set
\[
    \gamma_{\mathrm{corr}}
    =
    \operatorname{percentile}_{90}\!\left(|\Lambda|\right),
\]
using the same $90$th-percentile rule as \citet{jiang2024on}.

\textbf{Clustering.}
Components are matched using the absolute correlation of their response patterns. While \citet{jiang2024on} use greedy clustering, we use complete-linkage hierarchical clustering with threshold $\gamma_{\mathrm{corr}}$ to avoid dependence on component ordering and make the resulting clusters reproducible for a fixed ensemble. Thus, components belong to the same cluster only if every pair within the cluster has absolute response correlation at least $\gamma_{\mathrm{corr}}$.

\textbf{Feature centroids.}
Each cluster $C_t$ defines a model-defined feature. Since PCA response directions are defined only up to sign, we first align the signs of the normalized responses $u_{\theta,i}$ within each cluster and define its centroid as
\[
    c_t
    =
    \operatorname{unit}\!\left(
        \frac{1}{|C_t|}
        \sum_{(\theta,i)\in C_t}
        \epsilon_{\theta,i}u_{\theta,i}
    \right),
    \qquad
    \epsilon_{\theta,i}\in\{-1,+1\},
\]
where the signs \(\epsilon_{\theta,i}\) are aligned to the first member of \(C_t\) (ordered by model and component index): the anchor has \(\epsilon=+1\), and each remaining response is flipped if necessary so that its inner product with the anchor is nonnegative. The arbitrary sign of the anchor only determines the overall sign of \(c_t\), which is immaterial since subsequent comparisons use absolute correlation.

\textbf{Matching threshold for feature discrepancy.}
The matching threshold $\gamma$ used in Section~\ref{sec:feature-recovery} is calibrated from the clustering threshold rather than chosen independently. We set
\[
    \gamma = \min(1, \kappa\cdot\gamma_{\mathrm{corr}}),
\]
with $\kappa=2.0$ by default in our experiments. Components within a reference feature are themselves only required to have pairwise correlation at least \(\gamma_{\mathrm{corr}}\), so matching at the same threshold would not imply close agreement with that feature. Our default $\kappa=2$ avoids the strong compression observed at $\kappa=1$ while keeping the discrepancy between independent retraining groups small. We report ablations over $\kappa$ in Appendix~\ref{app:additional-results}.

\section{Harmlessness}\label{app:harmlessness}
\subsection{Proof of Proposition~\ref{prop:harmless-quantity}}

\begin{proof}
A gradient-descent step on $(x_i,y_i)$ of step size $\eta$ on the linear head gives
\[
    W_{\mathrm{retr}}^{+}
    =
    W_{\mathrm{retr}}
    -
    \eta\,\phi_{\mathrm{retr}}(x_i)g_i^{\top},
    \qquad
    W_{\mathrm{unl}}^{+}
    =
    W_{\mathrm{unl}}
    -
    \eta\,\phi_{\mathrm{unl}}(x_i)g_i^{\top}.
\]
For \(m\in\{\mathrm{retr},\mathrm{unl}\}\), let \(f_{m,i}^+(x):=(W_{m,i}^+)^\top\phi_m(x)\) denote the prediction at \(x\) after the head update on \((x_i,y_i)\). Thus, predictive equivalence before the update gives, for every $x_j\in\mathcal X$,
\[
    f_{\mathrm{retr},i}^+(x_j)-f_{\mathrm{unl},i}^+(x_j)
    =
    -\eta g_i
    \left(
        \phi_{\mathrm{retr}}(x_i)^\top\phi_{\mathrm{retr}}(x_j)
        -
        \phi_{\mathrm{unl}}(x_i)^\top\phi_{\mathrm{unl}}(x_j)
    \right).
\]
Since the term in parentheses is scalar,
\[
    \| f_{\mathrm{retr},i}^+(x_j)-f_{\mathrm{unl},i}^+(x_j) \|^2
    =
    \eta^2 \|g_i \|^2
    \left(
        \phi_{\mathrm{retr}}(x_i)^\top\phi_{\mathrm{retr}}(x_j)
        -
        \phi_{\mathrm{unl}}(x_i)^\top\phi_{\mathrm{unl}}(x_j)
    \right)^2.
\]
Summing over \(j\) therefore gives
\[
    \frac{
        \|F_{\mathrm{retr},i}^+
        -F_{\mathrm{unl},i}^+\|_F^2
    }{
        \eta^2 \|g_i\|_2^2
    }
    =
    \sum_{j=1}^N
    \left(
        \phi_{\mathrm{retr}}(x_i)^\top\phi_{\mathrm{retr}}(x_j)
        -
        \phi_{\mathrm{unl}}(x_i)^\top\phi_{\mathrm{unl}}(x_j)
    \right)^2.
\]
Summing over $i$ yields
\[
    \sum_{i=1}^N
    \frac{
        \|F_{\mathrm{retr},i}^+
        -F_{\mathrm{unl},i}^+\|_F^2
    }{
        \eta^2 \|g_i\|_2^2
    }
    =
    \left\|
        \Phi_{\mathrm{retr}}\Phi_{\mathrm{retr}}^\top
        -
        \Phi_{\mathrm{unl}}\Phi_{\mathrm{unl}}^\top
    \right\|_F^2.
\]
Let $H=I-\frac{1}{N}\mathbf 1\mathbf 1^\top$ denote the centering matrix. Since
\[
    \widetilde{\Phi}_{\mathrm{retr}}
    \widetilde{\Phi}_{\mathrm{retr}}^\top
    -
    \widetilde{\Phi}_{\mathrm{unl}}
    \widetilde{\Phi}_{\mathrm{unl}}^\top
    =
    H
    \left(
        \Phi_{\mathrm{retr}}\Phi_{\mathrm{retr}}^\top
        -
        \Phi_{\mathrm{unl}}\Phi_{\mathrm{unl}}^\top
    \right)
    H,
\]
and $H$ is an orthogonal projector,
\[
    \left\|
        \widetilde{\Phi}_{\mathrm{retr}}
        \widetilde{\Phi}_{\mathrm{retr}}^\top
        -
        \widetilde{\Phi}_{\mathrm{unl}}
        \widetilde{\Phi}_{\mathrm{unl}}^\top
    \right\|_F
    \leq
    \left\|
        \Phi_{\mathrm{retr}}\Phi_{\mathrm{retr}}^\top
        -
        \Phi_{\mathrm{unl}}\Phi_{\mathrm{unl}}^\top
    \right\|_F.
\]
Combining the two displays proves the claim.
\end{proof}

\subsection{Harmlessness Under a Shared Head}
\label{app:harmlessness-shared-head}
We formalize the fixed-head notion of harmlessness introduced in the main text. Fixing the unlearned head $W_{\mathrm{unl}}$, the representational difference is harmless when
\begin{equation}
\label{eq:shared-head-harmlessness}
W_{\mathrm{unl}}^\top
\bigl(
\phi_{\mathrm{retr}}(x)-\phi_{\mathrm{unl}}(x)
\bigr)
=0
\qquad \forall x \in \mathcal X.
\end{equation}
Thus, the two representations may differ in directions lying in $\ker(W_{\mathrm{unl}}^\top)$ without affecting the current predictions.

We now ask whether this property is preserved when the head is adapted. The following lemma characterizes exactly when~\eqref{eq:shared-head-harmlessness} survives a single gradient step.

\begin{lemma}
\label{lem:harmless-one-gd-condition}
Suppose~\eqref{eq:shared-head-harmlessness} holds. After one nonzero gradient step on $W_{\mathrm{unl}}$ using an example $(x_{\mathrm{ft}},y_{\mathrm{ft}})$ and the representation $\phi_{\mathrm{unl}}$, harmlessness is preserved if and only if
\[
    \bigl(
        \phi_{\mathrm{retr}}(x)-\phi_{\mathrm{unl}}(x)
    \bigr)^\top
    \phi_{\mathrm{unl}}(x_{\mathrm{ft}})
    =0
    \qquad \forall x \in \mathcal X.
\]
\end{lemma}

Hence, surviving one head update requires the representational difference to be orthogonal not only to the current head, but also to the representation of the adaptation point. Requiring this to hold across sufficiently many possible updates becomes highly restrictive.

\begin{proposition}
\label{prop:persistent-harmlessness}
Suppose there exist $d+1$ fine-tuning points $x_0,\ldots,x_d$ such that
\[
    \operatorname{affspan}
    \{\phi_{\mathrm{unl}}(x_0),\ldots,\phi_{\mathrm{unl}}(x_d)\}
    =
    \mathbb{R}^d.
\]
If the representational difference remains harmless after nonzero head updates on each of these points, then
\[
    \phi_{\mathrm{retr}}(x)
    =
    \phi_{\mathrm{unl}}(x)
    \qquad \forall x \in \mathcal X.
\]
In particular,
\[
    \widetilde{\Phi}_{\mathrm{retr}}
    =
    \widetilde{\Phi}_{\mathrm{unl}},
\]
so the two models have identical model-defined feature directions and strengths, up to the usual PCA sign and eigenspace ambiguities.
\end{proposition}

Thus, when the representation has full affine dimension, no nontrivial mismatch can remain harmless under sufficiently rich head adaptation. Importantly, full affine dimension is not an exceptional condition: it can be established for a broad class of feed-forward networks.

To make this explicit, consider a representation $\phi:\mathbb{R}^p\to\mathbb{R}^d$ defined by
\[
h_0(x)=x,
\qquad
h_\ell(x)=\sigma_\ell\!\left(W_\ell h_{\ell-1}(x)+b_\ell\right),
\quad
W_\ell\in\mathbb{R}^{m_\ell\times m_{\ell-1}},
\quad \ell=1,\ldots,L-1,
\]
and
\[
\phi(x)=W_L h_{L-1}(x)+b_L,
\qquad
W_L\in\mathbb{R}^{d\times m_{L-1}},
\]
with
\[
p=m_0\ge m_1\ge\cdots\ge m_{L-1}\ge d.
\]
We consider smooth activations with everywhere nonvanishing derivative, such as \(\tanh\), sigmoid, and softplus, together with Leaky-ReLU with nonzero negative slope.

\begin{proposition}
\label{prop:full-afine-dim-dl}
For the network above, suppose that every weight matrix has full rank. Then there exist $d+1$ inputs $x_0,\ldots,x_d$ such that
\[
\operatorname{affspan}\{\phi(x_0),\ldots,\phi(x_d)\}
=
\mathbb{R}^d.
\]
\end{proposition}

This provides a simple example in which harmlessness under the current head is maximally unstable: representational differences may be invisible before adaptation, yet requiring them to remain invisible across head updates forces the representations themselves to coincide.

Full affine dimension is not required to obtain a meaningful restriction. Even when nontrivial harmless directions remain, persistent harmlessness constrains the strengths of otherwise preserved features.

\begin{proposition}[Strength of preserved features]
\label{prop:strength-preserved-feat}
Suppose that the retrained and unlearned models have the same normalized response directions $u_i$, and that their representational difference remains harmless under the nonzero head updates on every $x \in \mathcal{X}$. Then, for every corresponding feature,
\[
    \lambda_i^{\mathrm{retr}}
    \geq
    \lambda_i^{\mathrm{unl}}.
\]
\end{proposition}

Taken together, these results show that equivalence under a fixed head can hide substantial representational mismatch, but this equivalence is generally unstable under subsequent adaptation.

\subsection{Proofs for Harmlessness Under a Shared Head}
\subsubsection{Proof of Lemma~\ref{lem:harmless-one-gd-condition}}
\begin{proof}
The argument is identical to the one-step calculation in the proof of Proposition~\ref{prop:harmless-quantity}. Using the initial harmlessness condition, after the update we obtain
\[
    (W_{\mathrm{unl}}^{+})^\top
    \bigl(\phi_{\mathrm{retr}}(x)-\phi_{\mathrm{unl}}(x)\bigr)
    =
    -\eta g_{\mathrm{ft}}\,
    \phi_{\mathrm{unl}}(x_{\mathrm{ft}})^\top
    \bigl(\phi_{\mathrm{retr}}(x)-\phi_{\mathrm{unl}}(x)\bigr).
\]
Since \(g_{\mathrm{ft}}\neq 0\) and the remaining factor is scalar, the left-hand side vanishes for all \(x\) if and only if the stated orthogonality condition holds.

\end{proof}

\subsubsection{Proof of Proposition~\ref{prop:persistent-harmlessness}}

\begin{proof}
By Lemma~\ref{lem:harmless-one-gd-condition},
\[
    \bigl(\phi_{\mathrm{retr}}(x)-\phi_{\mathrm{unl}}(x)\bigr)^\top
    \phi_{\mathrm{unl}}(x_i)=0
    \qquad \forall x,\; i=0,\ldots,d.
\]
Since $\{\phi_{\mathrm{unl}}(x_i)\}_{i=0}^d$ affinely spans $\mathbb{R}^d$, it also linearly spans $\mathbb{R}^d$. Hence $\phi_{\mathrm{retr}}(x)-\phi_{\mathrm{unl}}(x)=0$ for every $x$. The equality $\widetilde{\Phi}_{\mathrm{retr}}=\widetilde{\Phi}_{\mathrm{unl}}$ and the stated feature-level consequences follow immediately.

\end{proof}

\subsubsection{Proof of Proposition~\ref{prop:full-afine-dim-dl}}

\begin{lemma}
\label{lem:full-rank-local-image}
Let
\[
    \phi:\mathbb{R}^p\to\mathbb{R}^d,
    \qquad p\geq d,
\]
be continuously differentiable in a neighborhood of \(x^\star\), and suppose
\[
    \operatorname{rank} J_\phi(x^\star)=d.
\]
Then there exists a neighborhood \(U\) of \(x^\star\) such that
\(\phi(U)\) contains an open neighborhood of \(\phi(x^\star)\) in
\(\mathbb{R}^d\).
\end{lemma}

\begin{proof}
Since \(J_\phi(x^\star)\) has rank \(d\), some \(d\times d\) minor is nonzero at \(x^\star\). By continuity, this minor remains nonzero in a sufficiently small neighborhood \(U\) of \(x^\star\). Since \(d\) is the dimension of the codomain,
\[
    \operatorname{rank} J_\phi(x)=d,
    \qquad x\in U.
\]
The Constant Rank Theorem~\citep{book} therefore applies on \(U\). Since the rank equals
the dimension of the codomain, there exist local coordinates in which
\(\phi\) has the form
\[
    (u_1,\ldots,u_p)
    \longmapsto
    (u_1,\ldots,u_d).
\]
Hence \(\phi(U)\) contains an open neighborhood of
\(\phi(x^\star)\) in \(\mathbb{R}^d\).
\end{proof}

\begin{lemma}
\label{lem:leaky-full-rank}
Under the assumptions of Proposition~\ref{prop:full-afine-dim-dl}, if
$\sigma$ is Leaky ReLU with nonzero negative slope, then there exists
$x^\star\in\mathbb{R}^p$ such that no hidden pre-activation vanishes at
$x^\star$, $\phi$ is smooth in a neighborhood of $x^\star$, and
\[
    \operatorname{rank}J_\phi(x^\star)=d.
\]
\end{lemma}

\begin{proof}
We first construct a point away from all activation boundaries. For the first layer, each pre-activation
\[
    z_{1,j}(x)=w_{1,j}^\top x+b_{1,j}
\]
is a nonzero affine function, so its zero set is a proper hyperplane. Hence there exists a nonempty open set \(U_1\) on which all first-layer pre-activations have fixed nonzero signs.

Suppose we have a nonempty open set \(U_{\ell-1}\) on which all pre-activations in the first \(\ell-1\) layers are nonzero with fixed signs. Then \(h_{\ell-1}\), and hence every \(z_{\ell,j}\), is affine on \(U_{\ell-1}\). Under the full-rank assumption, none of these pre-activations is identically zero there, so their zero sets cannot cover \(U_{\ell-1}\). Thus there exists a nonempty open subset
\[
    U_\ell\subset U_{\ell-1}
\]
on which all layer-\(\ell\) pre-activations have fixed nonzero signs.

Iterating through the hidden layers yields a nonempty open set \(U\) on which the entire activation pattern is fixed. Hence the network is affine on \(U\), with
\[
    J_\phi(x)
    =
    W_LD_{L-1}W_{L-1}\cdots D_1W_1,
    \qquad x\in U,
\]
where each \(D_\ell\) is diagonal with entries in \(\{1,\alpha\}\) and is therefore invertible.

Under the full-rank and width decreasing assumptions of Proposition~\ref{prop:full-afine-dim-dl}, this product has rank \(d\). Thus any \(x^\star\in U\) is away from all activation boundaries, \(\phi\) is smooth in a neighborhood of \(x^\star\), and
\[
    \operatorname{rank}J_\phi(x^\star)=d.
\]
\end{proof}

\begin{proof}[Proof of Proposition~\ref{prop:full-afine-dim-dl}]
By Lemma~\ref{lem:full-rank-local-image}, it suffices to exhibit a point
$x^\star$ at which $\phi$ is smooth and
\[
    \operatorname{rank}J_\phi(x^\star)=d.
\]
For smooth activations with nonvanishing derivative, such a point exists
generically since all activation-derivative matrices are nonsingular. For
Leaky-ReLU networks, the same conclusion holds at a point away from all
activation boundaries; the existence of such a point with full-rank Jacobian
is established in Lemma~\ref{lem:leaky-full-rank}.
\end{proof}

\subsubsection{Proof of Proposition~\ref{prop:strength-preserved-feat}}
\begin{proof}
Let
\[
    D
    :=
    \widetilde{\Phi}_{\mathrm{unl}}
    -
    \widetilde{\Phi}_{\mathrm{retr}}.
\]
By Lemma~\ref{lem:harmless-one-gd-condition}, applied to all considered update points and then centered,
\[
    D\widetilde{\Phi}_{\mathrm{unl}}^\top=0.
\]
Since
\[
    \widetilde{\Phi}_{\mathrm{retr}}
    =
    \widetilde{\Phi}_{\mathrm{unl}}-D,
\]
we obtain
\[
    \widetilde{\Phi}_{\mathrm{retr}}
    \widetilde{\Phi}_{\mathrm{retr}}^\top
    =
    (\widetilde{\Phi}_{\mathrm{unl}}-D)
    (\widetilde{\Phi}_{\mathrm{unl}}-D)^\top 
    =
    \widetilde{\Phi}_{\mathrm{unl}}
    \widetilde{\Phi}_{\mathrm{unl}}^\top
    +
    DD^\top.
\]
Hence
\[
    \widetilde{\Phi}_{\mathrm{retr}}
    \widetilde{\Phi}_{\mathrm{retr}}^\top
    -
    \widetilde{\Phi}_{\mathrm{unl}}
    \widetilde{\Phi}_{\mathrm{unl}}^\top
    =
    DD^\top
    \succeq 0.
\]
Because the two models have the same normalized response directions \(u_i\),
\[
    \widetilde{\Phi}_{\mathrm{retr}}
    \widetilde{\Phi}_{\mathrm{retr}}^\top
    -
    \widetilde{\Phi}_{\mathrm{unl}}
    \widetilde{\Phi}_{\mathrm{unl}}^\top
    =
    \sum_i
    \bigl(
        \lambda_i^{\mathrm{retr}}
        -
        \lambda_i^{\mathrm{unl}}
    \bigr)
    u_i u_i^\top.
\]
Positive semidefiniteness therefore implies
\[
    \lambda_i^{\mathrm{retr}}
    \geq
    \lambda_i^{\mathrm{unl}}
\]
for every corresponding feature.
\end{proof}

\section{XOR Theory}\label{app:xor}
The model is trained for binary classification with logistic loss
\[
\mathcal{L}_S(\theta)
=
\mathbb{E}_{(x,y)\sim S}
\left[
\ell\bigl(m_\theta(x,y)\bigr)
\right],
\qquad
m_\theta(x,y):=yf_\theta(x),
\qquad
\ell(m)=\log(1+e^{-m}),
\]
where \(S=D\) for full-data training and \(S=D_r\) for retraining. Writing
\[
h(m):=-\ell'(m)=\frac{1}{1+e^m},
\]
full-batch gradient flow,
\[
\dot\theta_S(t)
=
-\nabla_\theta\mathcal{L}_S\bigl(\theta_S(t)\bigr),
\qquad
\theta_S(0)=\theta_0,
\]
satisfies
\begin{align}
\dot a_S(t)
&=
\mathbb{E}_{(x,y)\sim S}
\left[
h\bigl(m_{\theta_S(t)}(x,y)\bigr)\,ys
\right],
\label{eq:xor-gf-a}
\\
\dot u(t)
&=
\mathbb{E}_{(x,y)\sim S}
\left[
h\bigl(m_{\theta_S(t)}(x,y)\bigr)\,
y\bigl(u(t)^\top\widetilde z\bigr)\widetilde z
\right],
\label{eq:xor-gf-u}
\\
\dot v(t)
&=
-
\mathbb{E}_{(x,y)\sim S}
\left[
h\bigl(m_{\theta_S(t)}(x,y)\bigr)\,
y\bigl(v(t)^\top\widetilde z\bigr)\widetilde z
\right].
\label{eq:xor-gf-v}
\end{align}

\subsection{Connection to model features}
~\label{subsec:model-features-connection}
We take the activations immediately before the fixed linear output layer as the model representation,
\[
\phi_\theta(x)
=
\begin{pmatrix}
as\\
(u^\top\widetilde z)^2\\
(v^\top\widetilde z)^2
\end{pmatrix}.
\]
Let \(\xi:=\chi(z)=z_1z_2\in\{-1,1\}\). Since the representation depends only on whether an example is retained or forgotten and on \(\xi\), it takes four distinct values. Collecting them as rows gives
\[
\Phi =
\begin{pmatrix}
 ac
 & \varepsilon(u_1+u_2)^2
 & \varepsilon(v_1+v_2)^2
\\
-ac
 & \varepsilon(u_1-u_2)^2
 & \varepsilon(v_1-v_2)^2
\\
-ac
 & (1-\varepsilon)(u_1+u_2)^2
 & (1-\varepsilon)(v_1+v_2)^2
\\
 ac
 & (1-\varepsilon)(u_1-u_2)^2
 & (1-\varepsilon)(v_1-v_2)^2
\end{pmatrix},
\]
corresponding to
\[
(D_r,\xi=1),\qquad
(D_r,\xi=-1),\qquad
(D_f,\xi=1),\qquad
(D_f,\xi=-1),
\]
with probabilities
\[
\pi
=
\frac12
\begin{pmatrix}
1-\varepsilon\\
1-\varepsilon\\
\varepsilon\\
\varepsilon
\end{pmatrix}.
\]

We characterize the learned feature solution by PCA of the centered, \(\pi\)-weighted representation. Since \(\phi_\theta(x)\in\mathbb{R}^3\), there are at most three nonzero principal components. Each nonzero PCA direction induces a normalized response vector across the data, which is the quantity used to define model features in Section~\ref{sec:prelim_features}. The resulting principal directions and their ordering by eigenvalue depend on the particular model parameters $\theta$. We analyze these directions separately for each training and unlearning trajectory below.

\subsection{Proof roadmap}
\label{subsec:xor-proof-roadmap}

Although the derivations below are somewhat lengthy, they follow the same simple structure across all training and unlearning settings. For each trajectory---full-data training, retraining, random-label unlearning, random-label unlearning combined with fine-tuning, and gradient ascent---we first identify an invariant of the corresponding gradient-flow dynamics. In particular, retraining and all three unlearning trajectories preserve the same relation \(u=Jv\), for a matrix \(J\) specified below. We then use the invariant to reduce the parameter dynamics and determine the resulting coordinates of the feature representation. Finally, we compute the eigenvalues and eigenvectors of the corresponding centered covariance matrix, which gives the model-defined feature directions and strengths. Thus, the proofs differ mainly in the trajectory-specific algebra needed to establish the invariant and its limiting parameter values; the subsequent feature calculation is essentially the same.

\subsection{Full-data training dynamics}

We begin with the initial training stage, before any data are deleted, and characterize gradient flow on the full dataset \(D\). The chosen scaling makes the retain and forget contributions to the quadratic branch cancel, so training changes only the shortcut parameter.

For later use, define
\[
K:=
\begin{pmatrix}
0 & 1\\
1 & 0
\end{pmatrix}.
\]
Since \(\chi(z)=z_1z_2\) and \(z_i^2=1\),
\[
\mathbb{E}_{z}\!\left[\chi(z)zz^\top\right]=K,
\]
and therefore
\[
\mathbb{E}_{D_r}\!\left[y\widetilde z\widetilde z^\top\right]
=
\varepsilon K,
\qquad
\mathbb{E}_{D_f}\!\left[y\widetilde z\widetilde z^\top\right]
=
-(1 - \varepsilon) K.
\]
Hence the retain and forget contributions cancel exactly in the full-data gradient,
\begin{equation}
\mathbb{E}_{D}
\left[
y\widetilde z\widetilde z^\top
\right]
=
0.
\label{eq:xor-balanced-moment}
\end{equation}

\begin{lemma}[Frozen core under full-data training]
\label{lem:xor-full-core-frozen}
Under full-data gradient flow in the exactly balanced setting,
\[
u(t)=v(t)=\delta e_1
\qquad
\text{for all } t\geq 0.
\]
\end{lemma}

\begin{proof}
At initialization,
\[
u(0)=v(0)=\delta e_1.
\]
Suppose that at some point along the trajectory $u=v$.
 Then the two
quadratic terms cancel, and every training example has margin
\[
m_\theta(x,y)
=
yf_\theta(x)
=
yas
=
ac,
\]
where we used $ys=c$ for both retain and forget examples.

Hence the gradient-flow dynamics of the core parameters reduce to
\begin{align*}
\dot u
&=
h(ac)\,
\mathbb{E}_{D}
\left[
y\widetilde z\widetilde z^\top
\right]u,
\quad
\dot v
=
-h(ac)\,
\mathbb{E}_{D}
\left[
y\widetilde z\widetilde z^\top
\right]v.
\end{align*}
By~\eqref{eq:xor-balanced-moment},
\[
\mathbb{E}_{D}
\left[
y\widetilde z\widetilde z^\top
\right]
=0.
\]
Therefore $\dot u=\dot v=0$ whenever $u=v$.

Thus the trajectory
\[
u(t)=v(t)=\delta e_1
\]
is preserved by the gradient-flow vector field. By uniqueness of the
gradient-flow solution, it holds for all $t\geq0$.

\end{proof}

\begin{lemma}
\label{lem:xor-full-shortcut}
Under full-data gradient flow in the exactly balanced setting, $a(t)$ is increasing and
\[
a(t)\to+\infty
\qquad
\text{as } t\to\infty.
\]
\end{lemma}

\begin{proof}
By Lemma~\ref{lem:xor-full-core-frozen}, the quadratic branch cancels identically, so every example has margin \(a(t)c\). Thus
\[
\dot a(t)
=
c\,h(a(t)c)
=
\frac{c}{1+e^{ca(t)}}>0.
\]
Hence \(a(t)\) is increasing. If it were bounded, then \(a(t)\le M\) for some \(M\), implying
\[
\dot a(t)\ge \frac{c}{1+e^{cM}}>0,
\]
which contradicts boundedness. Therefore
\[
a(t)\to+\infty.
\]
\end{proof}

\begin{theorem}[Full-data feature solution]
Along the full-data training trajectory, the two nonzero normalized feature responses are, up to sign,
\[
u_{\mathrm{s}}
=
\frac{1}{\sqrt{2}}
\begin{pmatrix}
\sqrt{1-\varepsilon}\\
-\sqrt{1-\varepsilon}\\
-\sqrt{\varepsilon}\\
\sqrt{\varepsilon}
\end{pmatrix},
\qquad
u_{+}
=
\frac{1}{\sqrt{2}}
\begin{pmatrix}
\sqrt{\varepsilon}\\
\sqrt{\varepsilon}\\
-\sqrt{1-\varepsilon}\\
-\sqrt{1-\varepsilon}
\end{pmatrix},
\]
with corresponding PCA directions
\[
q_{\mathrm{s}}
=
\begin{pmatrix}
1\\0\\0
\end{pmatrix},
\qquad
q_{+}
=
\frac{1}{\sqrt{2}}
\begin{pmatrix}
0\\1\\1
\end{pmatrix},
\]
and strengths
\[
\lambda_{\mathrm{s}}(t)=a(t)^2c^2,
\qquad
\lambda_{+}
=
2\varepsilon(1-\varepsilon)(1-2\varepsilon)^2\delta^4.
\]
The orthogonal PCA direction
\[
q_{-}
=
\frac{1}{\sqrt{2}}
\begin{pmatrix}
0\\1\\-1
\end{pmatrix}
\]
has eigenvalue \(\lambda_-=0\), so there is no corresponding normalized feature response.
\end{theorem}

\begin{proof}
Recall the feature matrix from Section~\ref{subsec:model-features-connection}. By Lemmas~\ref{lem:xor-full-shortcut} and~\ref{lem:xor-full-core-frozen}, 
\[ u(t)=v(t)=\delta e_1 \] 
along the entire full-data training trajectory. 
Hence 
\[ u_2=v_2=0, \qquad u_1=v_1=\delta, \]
and the feature matrix reduces to 
\[
\Phi(t)
=
\begin{pmatrix}
ac & \varepsilon\delta^2 & \varepsilon\delta^2\\
-ac & \varepsilon\delta^2 & \varepsilon\delta^2\\
-ac & (1-\varepsilon)\delta^2 & (1-\varepsilon)\delta^2\\
ac & (1-\varepsilon)\delta^2 & (1-\varepsilon)\delta^2
\end{pmatrix}.
\]
Its probability-weighted mean is
\[
\bar\phi(t)
=
\begin{pmatrix}
0 &
2\varepsilon(1-\varepsilon)\delta^2 &
2\varepsilon(1-\varepsilon)\delta^2
\end{pmatrix}.
\]
Therefore the centered covariance matrix is
\[
C_{\mathrm{full}}
=
\begin{pmatrix}
a(t)^2c^2 & 0 & 0\\
0 & \gamma & \gamma\\
0 & \gamma & \gamma
\end{pmatrix},
\qquad
\gamma
=
\varepsilon(1-\varepsilon)(1-2\varepsilon)^2\delta^4.
\]
The first coordinate immediately gives \(q_{\mathrm{s}}\) with eigenvalue \(a(t)^2c^2\), while the remaining block
\[
\gamma
\begin{pmatrix}
1&1\\
1&1
\end{pmatrix}
\]
has normalized eigenvectors \((1,1)^\top/\sqrt{2}\) and \((1,-1)^\top/\sqrt{2}\) with eigenvalues \(2\gamma\) and \(0\), respectively. This yields the stated feature directions and strengths.

For each nonzero component, the corresponding normalized response vector is
\[
u_i
=
\frac{\operatorname{diag}(\pi)^{1/2}\widetilde\Phi q_i}{\sqrt{\lambda_i}}.
\]
Substituting \(q_{\mathrm{s}}\) and \(q_{+}\) gives the stated \(u_{\mathrm{s}}\) and \(u_{+}\); since \(\lambda_-=0\), no normalized response vector is defined for \(q_-\).
\end{proof}

\subsection{Retain-data training dynamics}

We next analyze retraining on the retain data after deletion. Unlike full-data training, deletion breaks the cancellation that kept the quadratic branch frozen, so the core parameters now evolve. We first show that the retraining trajectory preserves the symmetry \(u=Jv\), which reduces the dynamics to a low-dimensional form. We then characterize the resulting parameter growth and use the same invariant to derive the corresponding model features and their asymptotic ordering.

\begin{lemma}[Symmetry of the retraining trajectory]
\label{lem:xor-retrain-symmetry}
Consider gradient flow on the retain distribution $D_r$, and let
\[
J:=
\begin{pmatrix}
1&0\\
0&-1
\end{pmatrix}.
\]
Then
\[
u(t)=Jv(t)
\qquad
\text{for all }t\geq0.
\]
\end{lemma}

\begin{proof}
At initialization,
\[
u(0)=v(0)=\delta e_1,
\]
and hence $u(0)=Jv(0)$.

We show that the manifold
\[
\mathcal M:=\{(u,v):u=Jv\}
\]
is invariant under the retraining dynamics. Suppose $u=Jv$ and write
\[
u=
\begin{pmatrix}
p\\
s
\end{pmatrix},
\qquad
v=
\begin{pmatrix}
p\\
-s
\end{pmatrix}.
\]
For a retain example,
\[
\widetilde z=\sqrt{\varepsilon}\,z,
\qquad
y=\chi(z)=z_1z_2.
\]
Therefore, denoting the retain-data margin by \(m_R\),
\[
m_R
=
y\left(
as
+
\frac12(u^\top\widetilde z)^2
-
\frac12(v^\top\widetilde z)^2
\right)
=
ac+2\varepsilon ps,
\]
which is independent of the particular retain example.

Moreover,
\[
\mathbb E_{D_r}
\left[
y\widetilde z\widetilde z^\top
\right]
=
\varepsilon K,
\qquad
K:=
\begin{pmatrix}
0&1\\
1&0
\end{pmatrix}.
\]
Hence, on $\mathcal M$,
\[
\dot u
=
\varepsilon h(m_R)Ku,
\qquad
\dot v
=
-\varepsilon h(m_R)Kv.
\]
Since
\[
JK=-KJ,
\]
we obtain
\[
J\dot v
=
-\varepsilon h(m_R)JKv
=
\varepsilon h(m_R)KJv
=
\varepsilon h(m_R)Ku
=
\dot u.
\]
Thus the gradient-flow vector field is tangent to $\mathcal M$.
Since the gradient-flow solution is unique and the initialization lies
in $\mathcal M$, the entire trajectory remains in $\mathcal M$.
\end{proof}

\begin{lemma}
\label{lem:hyperbolic-time-change}
Let $p(t)$ and $s(t)$ satisfy
\[
\dot p(t)=\gamma(t)s(t),
\qquad
\dot s(t)=\gamma(t)p(t),
\]
for some continuous scalar function $\gamma(t)$, and define
\[
G(t):=\int_0^t \gamma(\tau)\,d\tau.
\]
Then
\begin{align*}
p(t)+s(t)
&=
\bigl(p(0)+s(0)\bigr)e^{G(t)},\\
p(t)-s(t)
&=
\bigl(p(0)-s(0)\bigr)e^{-G(t)}.
\end{align*}
In particular, if
\[
p(0)=\delta,
\qquad
s(0)=0,
\]
then
\[
p(t)=\delta\cosh G(t),
\qquad
s(t)=\delta\sinh G(t).
\]
\end{lemma}

\begin{proof}
Adding and subtracting the two differential equations gives
\[
\frac{d}{dt}(p+s)=\gamma(t)(p+s),
\qquad
\frac{d}{dt}(p-s)=-\gamma(t)(p-s).
\]
Integrating these scalar equations yields
\[
p(t)+s(t)
=
\bigl(p(0)+s(0)\bigr)e^{G(t)}
\]
and
\[
p(t)-s(t)
=
\bigl(p(0)-s(0)\bigr)e^{-G(t)}.
\]
The stated formulas follow immediately.
\end{proof}

\begin{lemma}
\label{lem:xor-retrain-dynamics}
Under retraining on $D_r$, define
\[
G(t)
:=
\int_0^t
\varepsilon h(m_R(\tau))\,d\tau.
\]
Then
\begin{align}
u(t)
&=
\delta
\begin{pmatrix}
\cosh G(t)\\
\sinh G(t)
\end{pmatrix},
&
v(t)
&=
\delta
\begin{pmatrix}
\cosh G(t)\\
-\sinh G(t)
\end{pmatrix},
\label{eq:xor-retrain-uv}
\\
a(t)
&=
a(0)+\frac{c}{\varepsilon}G(t).
\label{eq:xor-retrain-a}
\end{align}
Consequently,
\[
m_R(t)
=
ca(0)
+
\frac{c^2}{\varepsilon}G(t)
+
\varepsilon\delta^2\sinh\bigl(2G(t)\bigr).
\]
Moreover,
\[
G(t)\to\infty
\qquad
\text{as }t\to\infty.
\]
\end{lemma}

\begin{proof}
By Lemma~\ref{lem:xor-retrain-symmetry}, write
\[
u(t)=
\begin{pmatrix}
p(t)\\
s(t)
\end{pmatrix},
\qquad
v(t)=
\begin{pmatrix}
p(t)\\
-s(t)
\end{pmatrix}.
\]
The retraining dynamics therefore reduce to
\[
\dot p
=
\varepsilon h(m_R)s,
\qquad
\dot s
=
\varepsilon h(m_R)p.
\]
Applying Lemma~\ref{lem:hyperbolic-time-change} with $\gamma(t)=\varepsilon h(m_R(t))$ and using $p(0)=\delta$, $s(0)=0$ gives
\[
p(t)=\delta\cosh G(t),
\qquad
s(t)=\delta\sinh G(t),
\]
which proves~\eqref{eq:xor-retrain-uv}. For the shortcut parameter,
\[
\dot a(t)
=
c h(m_R(t))
=
\frac{c}{\varepsilon}\dot G(t).
\]
Hence
\[
a(t)
=
a(0)+\frac{c}{\varepsilon}G(t).
\]

Finally, since
\[
2p(t)s(t)
=
\delta^2\sinh\bigl(2G(t)\bigr),
\]
the retain margin becomes
\[
m_R(t)
=
ca(0)
+
\frac{c^2}{\varepsilon}G(t)
+
\varepsilon\delta^2\sinh\bigl(2G(t)\bigr).
\]

It remains to show that $G(t)$ is unbounded. If $G(t)\leq M$ for all $t\geq0$, then the expression above implies that $m_R(t)$ is uniformly bounded above by some finite $M_R$. Therefore,
\[
\dot G(t)
=
\varepsilon h(m_R(t))
\geq
\varepsilon h(M_R)
>0,
\]
which implies that $G(t)$ grows at least linearly in $t$, a
contradiction. Hence
\[
G(t)\to\infty.
\]
\end{proof}

\begin{proposition}[Feature solution under the $J$-invariant]
\label{lem:xor-j-invariant-pca}
Suppose that
\[
u=Jv,
\qquad
J=
\begin{pmatrix}
1&0\\
0&-1
\end{pmatrix},\qquad
u=(u_1,u_2)^\top.
\]
Then the centered representation has PCA directions
\[
q_{\mathrm{s}}
=
\begin{pmatrix}
1\\0\\0
\end{pmatrix},
\qquad
q_{+}
=
\frac{1}{\sqrt{2}}
\begin{pmatrix}
0\\1\\1
\end{pmatrix},
\qquad
q_{-}
=
\frac{1}{\sqrt{2}}
\begin{pmatrix}
0\\1\\-1
\end{pmatrix}.
\]
For each nonzero component, the corresponding normalized response vector is, up to sign,
\[
u_{\mathrm{s}}
=
\frac{1}{\sqrt{2}}
\begin{pmatrix}
\sqrt{1-\varepsilon}\\
-\sqrt{1-\varepsilon}\\
-\sqrt{\varepsilon}\\
\sqrt{\varepsilon}
\end{pmatrix},
\qquad
u_{+}
=
\frac{1}{\sqrt{2}}
\begin{pmatrix}
\sqrt{\varepsilon}\\
\sqrt{\varepsilon}\\
-\sqrt{1-\varepsilon}\\
-\sqrt{1-\varepsilon}
\end{pmatrix},
\qquad
u_{-}
=
\frac{1}{\sqrt{2}}
\begin{pmatrix}
\sqrt{\varepsilon}\\
-\sqrt{\varepsilon}\\
\sqrt{1-\varepsilon}\\
-\sqrt{1-\varepsilon}
\end{pmatrix}.
\]
Their corresponding strengths are
\[
\lambda_{\mathrm{s}}(t)=a(t)^2c^2,
\qquad
\lambda_{+}(t)
=
2\varepsilon(1-\varepsilon)(1-2\varepsilon)^2
\bigl(u_1(t)^2+u_2(t)^2\bigr)^2,
\]
and
\[
\lambda_{-}(t)
=
8\varepsilon(1-\varepsilon)
u_1(t)^2u_2(t)^2.
\]
\end{proposition}

\begin{proof}
Define
\[
s:=(u_1+u_2)^2,
\qquad
d:=(u_1-u_2)^2.
\]
Under the $J$-invariant, the feature matrix from Section~\ref{subsec:model-features-connection} becomes
\[
\Phi =
\begin{pmatrix}
 ac & \varepsilon s & \varepsilon d\\
-ac & \varepsilon d & \varepsilon s\\
-ac & (1-\varepsilon)s & (1-\varepsilon)d\\
 ac & (1-\varepsilon)d & (1-\varepsilon)s
\end{pmatrix}.
\]
Its probability-weighted mean is
\[
\bar\phi
=
\begin{pmatrix}
0 &
\varepsilon(1-\varepsilon)(s+d) &
\varepsilon(1-\varepsilon)(s+d)
\end{pmatrix}.
\]
For convenience, define
\[
A:=\varepsilon s-(1-\varepsilon)d,
\qquad
B:=\varepsilon d-(1-\varepsilon)s.
\]
Then the centered feature matrix is
\[
\widetilde{\Phi}
=
\begin{pmatrix}
 ac & \varepsilon A & \varepsilon B\\
-ac & \varepsilon B & \varepsilon A\\
-ac & -(1-\varepsilon)B & -(1-\varepsilon)A\\
 ac & -(1-\varepsilon)A & -(1-\varepsilon)B
\end{pmatrix}.
\]
Using the row probabilities $\pi$ from Section~\ref{subsec:model-features-connection}, the centered feature covariance is
\[
C
=
\widetilde{\Phi}^{\top}
\operatorname{diag}(\pi)
\widetilde{\Phi}
=
\begin{pmatrix}
a^2c^2 & 0 & 0\\
0 &
\dfrac{\varepsilon(1-\varepsilon)}{2}(A^2+B^2)
&
\varepsilon(1-\varepsilon)AB\\
0 &
\varepsilon(1-\varepsilon)AB
&
\dfrac{\varepsilon(1-\varepsilon)}{2}(A^2+B^2)
\end{pmatrix}.
\]
Thus \(q_{\mathrm{s}}\) is an eigenvector with eigenvalue \(a^2c^2\), while the symmetric \(2\times2\) block has eigenvectors \(q_+\) and \(q_-\) with eigenvalues
\[
\lambda_{+}
=
\frac{\varepsilon(1-\varepsilon)}{2}(A+B)^2,
\qquad
\lambda_{-}
=
\frac{\varepsilon(1-\varepsilon)}{2}(A-B)^2.
\]

Since
\[
A+B=-(1-2\varepsilon)(s+d),
\qquad
A-B=s-d,
\]
and
\[
s+d=2(u_1^2+u_2^2),
\qquad
s-d=4u_1u_2,
\]
the stated eigenvalues follow.

For the corresponding feature responses, multiplying the centered representation by the PCA directions and applying the probability weighting gives
\[
\operatorname{diag}(\pi)^{1/2}\widetilde{\Phi}q_{\mathrm{s}}
=
\frac{ac}{\sqrt{2}}
\begin{pmatrix}
\sqrt{1-\varepsilon}\\
-\sqrt{1-\varepsilon}\\
-\sqrt{\varepsilon}\\
\sqrt{\varepsilon}
\end{pmatrix},
\]
\[
\operatorname{diag}(\pi)^{1/2}\widetilde{\Phi}q_{+}
=
\frac{A+B}{2}
\begin{pmatrix}
\varepsilon\sqrt{1-\varepsilon}\\
\varepsilon\sqrt{1-\varepsilon}\\
-(1-\varepsilon)\sqrt{\varepsilon}\\
-(1-\varepsilon)\sqrt{\varepsilon}
\end{pmatrix},
\]
and
\[
\operatorname{diag}(\pi)^{1/2}\widetilde{\Phi}q_{-}
=
\frac{A-B}{2}
\begin{pmatrix}
\varepsilon\sqrt{1-\varepsilon}\\
-\varepsilon\sqrt{1-\varepsilon}\\
(1-\varepsilon)\sqrt{\varepsilon}\\
-(1-\varepsilon)\sqrt{\varepsilon}
\end{pmatrix}.
\]
Normalizing each nonzero response therefore gives, up to sign,
\[
u_{\mathrm{s}}
=
\frac{1}{\sqrt{2}}
\begin{pmatrix}
\sqrt{1-\varepsilon}\\
-\sqrt{1-\varepsilon}\\
-\sqrt{\varepsilon}\\
\sqrt{\varepsilon}
\end{pmatrix},
\qquad
u_{+}
=
\frac{1}{\sqrt{2}}
\begin{pmatrix}
\sqrt{\varepsilon}\\
\sqrt{\varepsilon}\\
-\sqrt{1-\varepsilon}\\
-\sqrt{1-\varepsilon}
\end{pmatrix},
\qquad
u_{-}
=
\frac{1}{\sqrt{2}}
\begin{pmatrix}
\sqrt{\varepsilon}\\
-\sqrt{\varepsilon}\\
\sqrt{1-\varepsilon}\\
-\sqrt{1-\varepsilon}
\end{pmatrix}.
\]
\end{proof}

\begin{theorem}[Asymptotic ordering of retraining feature directions]\label{lem:xor-retrain-pca-asymptotics}
Assume $0<\varepsilon<\frac12$. Along the retraining trajectory, the two core-feature eigenvalues asymptotically dominate the shortcut eigenvalue:
\[
\frac{\lambda_{\mathrm{s}}(t)}{\lambda_{+}(t)}
\longrightarrow 0,
\qquad
\frac{\lambda_{\mathrm{s}}(t)}{\lambda_{-}(t)}
\longrightarrow 0
\qquad
\text{as } t\to\infty.
\]
Moreover,
\[
\frac{\lambda_{+}(t)}{\lambda_{-}(t)}
\longrightarrow
(1-2\varepsilon)^2.
\]
Consequently, for all sufficiently large $t$,
\[
\lambda_{-}(t)>\lambda_{+}(t)>\lambda_{\mathrm{s}}(t),
\]
and the fraction of the total feature variance explained by the shortcut
direction vanishes:
\[
\frac{\lambda_{\mathrm{s}}(t)}
{\lambda_{\mathrm{s}}(t)+\lambda_{+}(t)+\lambda_{-}(t)}
\longrightarrow 0.
\]
\end{theorem}

\begin{proof}
By Lemma~\ref{lem:xor-retrain-dynamics}, along the retraining trajectory,
\[ u_1(t)=\delta\cosh G(t), \qquad u_2(t)=\delta\sinh G(t), \qquad a(t)=\frac{c}{\varepsilon}G(t), 
\]
where
\[
G(t)\to\infty \qquad \text{as }t\to\infty.
\]

Since retraining preserves the invariant $u(t)=Jv(t)$ by
Lemma~\ref{lem:xor-retrain-symmetry}, the principal feature directions and their eigenvalues are given directly by Proposition~\ref{lem:xor-j-invariant-pca}.
Substituting these expressions into the eigenvalues gives
\[
\lambda_{\mathrm{s}}(t)
=
\frac{c^4}{\varepsilon^2}G(t)^2. 
\]

Moreover, 
\[(u_1^2 + u_2^2) = \delta^2 \cosh\bigl(2G(t) \bigr) \qquad u_1 u_2 = \frac12 \delta^2 \sinh\bigl(2G(t) \bigr).\]

Therefore,
\[
\lambda_{+}(t) = 
2\varepsilon(1-\varepsilon)(1-2\varepsilon)^2 \delta^4 \cosh^2\bigl(2G(t) \bigr)
\]
and
\[
\lambda_{-}(t)
=
2\varepsilon(1-\varepsilon) \delta^4 \sinh^2\bigl(2G(t) \bigr).
\]

Since $G(t)\to\infty$, both $\cosh^2(2G(t))$ and $\sinh^2(2G(t))$ grow exponentially in $G(t)$, whereas $\lambda_{\mathrm{s}}(t)$ grows only quadratically.
Hence 
\[ 
\frac{\lambda_{\mathrm{s}}(t)}{\lambda_{+}(t)} \to 0, \qquad 
\frac{\lambda_{\mathrm{s}}(t)}{\lambda_{-}(t)} \to 0. 
\]

Furthermore,
\[
\frac{\lambda_{+}(t)}{\lambda_{-}(t)} = (1 - 2\varepsilon)^2 \coth^2\bigl(2G(t)\bigr) \longrightarrow (1 - 2\varepsilon)^2.
\]
For $0<\varepsilon<1/2$,
\[
(1 - 2\varepsilon)^2 < 1,
\]
and therefore, for all sufficiently large $t$, 
\[ 
\lambda_{-}(t)>\lambda_{+}(t)>\lambda_{\mathrm{s}}(t). 
\]
The vanishing fraction of variance explained by the shortcut direction
follows immediately from these asymptotic relations.
\end{proof}

\subsection{Random Labels Unlearning dynamics}

We next analyze Random Labels (RL) unlearning without retain fine-tuning, initialized from the full-data solution. By Lemma~\ref{lem:xor-full-core-frozen},
\[
a(0)=A,
\qquad
u(0)=v(0)=\delta e_1,
\]
where \(A>0\) is the shortcut coefficient at the start of unlearning.

RL replaces the forget labels by independent random labels
\[
\hat y\sim\operatorname{Unif}\{-1,1\},
\]
and performs gradient flow on
\[
\mathcal L_{\mathrm{RL}}(\theta)
=
\mathbb E_{(x, y)\sim {D_f}}
\mathbb E_{\hat y}
\left[
\ell\bigl(\hat y f_\theta(x)\bigr)
\right].
\]
The corresponding dynamics are
\begin{align}
\dot a(t)
&=
\mathbb E_{(x, y)\sim {D_f}}
\mathbb E_{\hat y}
\left[
h\bigl(\hat y f_\theta(x)\bigr)\hat y\,s
\right],
\\
\dot u(t)
&=
\mathbb E_{(x, y)\sim {D_f}}
\mathbb E_{\hat y}
\left[
h\bigl(\hat y f_\theta(x)\bigr)\hat y
\bigl(u(t)^\top\widetilde z\bigr)\widetilde z
\right],
\\
\dot v(t)
&=
-
\mathbb E_{(x, y)\sim {D_f}}
\mathbb E_{\hat y}
\left[
h\bigl(\hat y f_\theta(x)\bigr)\hat y
\bigl(v(t)^\top\widetilde z\bigr)\widetilde z
\right].
\end{align}
Since \(\hat y\) is uniform, define
\[
\psi(r)
:=
\mathbb E_{\hat y}
\left[
\hat y\,h(\hat y r)
\right]
=
\frac12\bigl(h(r)-h(-r)\bigr)
=
-\frac12\tanh\left(\frac r2\right).
\]
Averaging over the random labels gives
\begin{align}
\dot a(t)
&=
\mathbb E_{(x, y)\sim {D_f}}
\left[
\psi\bigl(f_\theta(x)\bigr)s
\right],
\\
\dot u(t)
&=
\mathbb E_{(x, y)\sim {D_f}}
\left[
\psi\bigl(f_\theta(x)\bigr)
\bigl(u(t)^\top\widetilde z\bigr)\widetilde z
\right],
\\
\dot v(t)
&=
-
\mathbb E_{(x, y)\sim {D_f}}
\left[
\psi\bigl(f_\theta(x)\bigr)
\bigl(v(t)^\top\widetilde z\bigr)\widetilde z
\right].
\end{align}

As in retraining, the RL trajectory preserves the invariant \(u(t)=Jv(t)\). We use this symmetry to reduce the dynamics, characterize their asymptotic behavior, and derive the resulting principal feature directions and strengths.

\begin{lemma}[Symmetry of the Random Labels trajectory]
\label{lem:xor-rl-symmetry}
Consider Random Labels gradient flow initialized from
\[
a(0)=A,
\qquad
u(0)=v(0)=\delta e_1.
\]
Let
\[
J=
\begin{pmatrix}
1&0\\
0&-1
\end{pmatrix}.
\]
Then the relation
\[
u(t)=Jv(t)
\]
is preserved along the entire Random Labels trajectory. 
\end{lemma}

\begin{proof}
The initialization satisfies \(u(0)=Jv(0)\). As in Lemma~\ref{lem:xor-retrain-symmetry}, we show that the manifold
\[
\mathcal M:=\{(u,v):u=Jv\}
\]
is invariant under the Random Labels dynamics. Suppose \(u=Jv\) and write
\[
u=
\begin{pmatrix}
p\\
s
\end{pmatrix},
\qquad
v=
\begin{pmatrix}
p\\
-s
\end{pmatrix}.
\]

For a forget example,
\[
f_\theta(x)
=
-a\chi(z)c
+
\frac{1-\varepsilon}{2}
\left[
(pz_1+sz_2)^2-(pz_1-sz_2)^2
\right]
=
\chi(z)\bigl(-ac+2(1-\varepsilon)ps\bigr).
\]
Since \(\psi\) is odd,
\[
\psi\bigl(f_\theta(x)\bigr)
=
\chi(z)\,
\psi\bigl(-ac+2(1-\varepsilon)ps\bigr).
\]

Then
\begin{align*}
\dot u(t)
&=
\mathbb E_{(x, y)\sim {D_f}}
\left[
\psi\bigl(f_\theta(x)\bigr)
\bigl(u(t)^\top\widetilde z\bigr)\widetilde z
\right]
\\
&=
\psi\bigl(-ac+2(1-\varepsilon)ps\bigr)
\mathbb E_{(x, y)\sim {D_f}}
\left[
\chi(z)\widetilde z\widetilde z^\top
\right]u(t)
\\
&=
(1-\varepsilon)
\psi\bigl(-ac+2(1-\varepsilon)ps\bigr)Ku(t),
\end{align*}
where
\[
\mathbb E_{(x, y)\sim {D_f}}
\left[
\chi(z)\widetilde z\widetilde z^\top
\right]
=
(1-\varepsilon)K.
\]
Similarly,
\[
\dot v(t)
=
-(1-\varepsilon)
\psi\bigl(-ac+2(1-\varepsilon)ps\bigr)Kv(t).
\]

Since \(JK=-KJ\), the same calculation as in Lemma~\ref{lem:xor-retrain-symmetry} gives
\[
J\dot v=\dot u.
\]
Thus the gradient-flow vector field is tangent to \(\mathcal M\). Since the initialization lies in \(\mathcal M\) and the solution is unique, the entire Random Labels trajectory remains in \(\mathcal M\).
\end{proof}

\begin{lemma}
\label{lem:xor-rl-dynamics}
Along the Random Labels trajectory, define
\[
G_{\mathrm{RL}}(t)
:=
\int_0^t
(1-\varepsilon)
\psi\bigl(-a(\tau)c
+2(1-\varepsilon)p(\tau)s(\tau)\bigr)
\,d\tau.
\]
Then
\[
u_1(t)=\delta\cosh G_{\mathrm{RL}}(t),
\qquad
u_2(t)=\delta\sinh G_{\mathrm{RL}}(t),
\qquad
a(t)
=
A-\frac{c}{1-\varepsilon}G_{\mathrm{RL}}(t).
\]
\end{lemma}

\begin{proof}
From the proof of Lemma~\ref{lem:xor-rl-symmetry}, the dynamics on the
invariant manifold $u=Jv$ satisfy
\[
\dot p
=
(1-\varepsilon)
\psi\bigl(-ac+2(1-\varepsilon)ps\bigr)s,
\qquad
\dot s
=
(1-\varepsilon)
\psi\bigl(-ac+2(1-\varepsilon)ps\bigr)p.
\]

Define
\[
G_{\mathrm{RL}}(t)
:=
\int_0^t
(1-\varepsilon)
\psi\bigl(-a(\tau)c
+2(1-\varepsilon)p(\tau)s(\tau)\bigr)
\,d\tau.
\]
Applying Lemma~\ref{lem:hyperbolic-time-change} and using
$p(0)=\delta$ and $s(0)=0$, we obtain
\[
p(t)=\delta\cosh G_{\mathrm{RL}}(t),
\qquad
s(t)=\delta\sinh G_{\mathrm{RL}}(t).
\]

For the shortcut parameter,
\[
\dot a
=
-c\,
\psi\bigl(-ac+2(1-\varepsilon)ps\bigr)
=
-\frac{c}{1-\varepsilon}\dot G_{\mathrm{RL}}.
\]
Since $a(0)=A$ and $G_{\mathrm{RL}}(0)=0$,
\[
a(t)
=
A-\frac{c}{1-\varepsilon}G_{\mathrm{RL}}(t).
\]
\end{proof}

\begin{lemma}[Convergence of the Random Labels dynamics]
\label{lem:xor-rl-convergence}
Along the Random Labels trajectory,
\[
G_{\mathrm{RL}}(t)\longrightarrow G_\star
\qquad
\text{as }t\to\infty,
\]
where $G_\star>0$ is the unique solution of
\[
-Ac
+
\frac{c^2}{1-\varepsilon}G_\star
+
(1-\varepsilon)\delta^2\sinh(2G_\star)
=
0.
\]
Equivalently, the limiting parameters satisfy
\[
a_\star c
=
(1-\varepsilon)\delta^2\sinh(2G_\star),
\]
where
\[
a_\star
=
A-\frac{c}{1-\varepsilon}G_\star.
\]
\end{lemma}

\begin{proof}
By Lemma~\ref{lem:xor-rl-dynamics},
\[
\dot G_{\mathrm{RL}}(t)
=
(1-\varepsilon)\psi\bigl(\rho(G_{\mathrm{RL}}(t))\bigr),
\]
where
\[
\rho(G)
:=
-Ac
+
\frac{c^2}{1-\varepsilon}G
+
(1-\varepsilon)\delta^2\sinh(2G).
\]

The function $\rho$ is strictly increasing, since
\[
\rho'(G)
=
\frac{c^2}{1-\varepsilon}
+
2(1-\varepsilon)\delta^2\cosh(2G)
>0.
\]
Moreover,
\[
\rho(0)=-Ac<0,
\qquad
\rho(G)\to\infty
\quad\text{as }G\to\infty.
\]
Hence, there exists a unique $G_\star>0$ such that
\[
\rho(G_\star)=0.
\]

Because \(\psi(r)>0\) for \(r<0\) and \(\psi(r)<0\) for \(r>0\),
\[
\dot G_{\mathrm{RL}}(t)>0
\quad\text{for }G_{\mathrm{RL}}(t)<G_\star,
\qquad
\dot G_{\mathrm{RL}}(t)<0
\quad\text{for }G_{\mathrm{RL}}(t)>G_\star.
\]

Since \(G_{\mathrm{RL}}(0)=0<G_\star\), uniqueness prevents the trajectory from crossing the equilibrium \(G_\star\). Hence \(G_{\mathrm{RL}}(t)\) is increasing and bounded above by \(G_\star\), and therefore converges. Its limit must be a fixed point of the dynamics, so by uniqueness it equals \(G_\star\).

Finally, \(\rho(G_\star)=0\) together with
\[
a_\star
=
A-\frac{c}{1-\varepsilon}G_\star
\]
gives
\[
a_\star c
=
(1-\varepsilon)\delta^2\sinh(2G_\star).
\]
\end{proof}

\begin{theorem}[Feature spectrum along the Random Labels trajectory]
\label{lem:xor-rl-pca}

Along the Random Labels trajectory, the normalized feature responses and corresponding PCA directions are those of Proposition~\ref{lem:xor-j-invariant-pca}. Their strengths are
\[
\lambda_{\mathrm{s}}(t)
=
a(t)^2c^2,
\]
\[
\lambda_{+}(t)
=
2\varepsilon(1-\varepsilon)(1-2\varepsilon)^2
\delta^4\cosh^2\bigl(2G_{\mathrm{RL}}(t)\bigr),
\]
and
\[
\lambda_{-}(t)
=
2\varepsilon(1-\varepsilon)
\delta^4\sinh^2\bigl(2G_{\mathrm{RL}}(t)\bigr).
\]
Moreover, for \(t>0\),
\[
\frac{\lambda_{\mathrm{s}}(t)}
{\lambda_{-}(t)}
>
\frac{1-\varepsilon}{2\varepsilon},
\qquad
\frac{\lambda_{\mathrm{s}}(t)}
{\lambda_{-}(t)}
\longrightarrow
\frac{1-\varepsilon}{2\varepsilon}
\]
monotonically from above as \(t\to\infty\).
\end{theorem}

\begin{proof}
By Lemma~\ref{lem:xor-rl-symmetry}, the Random Labels trajectory satisfies \(u(t)=Jv(t)\), so the principal directions and their eigenvalues are given by Proposition~\ref{lem:xor-j-invariant-pca}. By Lemma~\ref{lem:xor-rl-dynamics},
\[
p(t)=\delta\cosh G_{\mathrm{RL}}(t),
\qquad
s(t)=\delta\sinh G_{\mathrm{RL}}(t),
\]
and hence
\[
p(t)^2+s(t)^2
=
\delta^2\cosh\bigl(2G_{\mathrm{RL}}(t)\bigr),
\qquad
p(t)s(t)
=
\frac{\delta^2}{2}
\sinh\bigl(2G_{\mathrm{RL}}(t)\bigr).
\]
Substituting these expressions into Proposition~\ref{lem:xor-j-invariant-pca} gives the stated eigenvalues.

For the shortcut-to-core ratio,
\[
\frac{\lambda_{\mathrm{s}}(t)}
{\lambda_{-}(t)}
=
\frac{1-\varepsilon}{2\varepsilon}
\left(
\frac{a(t)c}
{(1-\varepsilon)\delta^2
\sinh\bigl(2G_{\mathrm{RL}}(t)\bigr)}
\right)^2.
\]

By Lemma~\ref{lem:xor-rl-convergence}, \(G_{\mathrm{RL}}(t)\) increases to \(G_\star\). For \(G_{\mathrm{RL}}(t)<G_\star\),
\[
\rho\bigl(G_{\mathrm{RL}}(t)\bigr)
=
-a(t)c
+
(1-\varepsilon)\delta^2
\sinh\bigl(2G_{\mathrm{RL}}(t)\bigr)
<0,
\]
so
\[
a(t)c
>
(1-\varepsilon)\delta^2
\sinh\bigl(2G_{\mathrm{RL}}(t)\bigr).
\]
Thus, for every finite $t>0$,
\[
\frac{\lambda_{\mathrm{s}}(t)}
{\lambda_{-}(t)}
>
\frac{1-\varepsilon}{2\varepsilon}.
\]

Moreover, \(a(t)\) is strictly decreasing while \(G_{\mathrm{RL}}(t)\), and hence \(\sinh(2G_{\mathrm{RL}}(t))\), is strictly increasing, so the ratio is strictly decreasing. Finally,
\[
a_\star c
=
(1-\varepsilon)\delta^2\sinh(2G_\star)
\]
by Lemma~\ref{lem:xor-rl-convergence}, which gives the stated limit.
\end{proof}

\begin{corollary}[Ordering of the core feature directions under Random Labels]
\label{lem:xor-rl-core-ordering}
Along the Random Labels trajectory,
\[
\frac{\lambda_{+}(t)}{\lambda_{-}(t)}
=
(1-2\varepsilon)^2
\coth^2\bigl(2G_{\mathrm{RL}}(t)\bigr),
\]
and this ratio is strictly decreasing for $t>0$.

In particular, at the beginning of unlearning,
\[
\lambda_{+}(0)>\lambda_{-}(0)=0.
\]
The two eigenvalues cross at most once, namely when
\[
\tanh\bigl(2G_{\mathrm{RL}}(t)\bigr)=1-2\varepsilon.
\]
\end{corollary}

\begin{proof}
The ratio follows directly from the expressions for
$\lambda_{+}$ and $\lambda_{-}$. Since $G_{\mathrm{RL}}(t)$ is strictly
increasing and $\coth(x)$ is strictly decreasing for $x>0$, the ratio is
strictly decreasing. At $t=0$, we have $G_{\mathrm{RL}}(0)=0$, so
$\lambda_{-}(0)=0$, while $\lambda_{+}(0)>0$ for
$0<\varepsilon<1/2$. The crossing condition follows by setting
$\lambda_{+}=\lambda_{-}$.
\end{proof}

\subsection{Random Labels with Retain Fine-Tuning unlearning dynamics}

We next consider Random Labels unlearning with simultaneous retain fine-tuning. The initialization and random relabeling of the forget set are unchanged; in addition, we minimize the standard training loss on the retain set. This corresponds to the objective
\[
L_{\mathrm{RL+FT}}(\theta)
=
\lambda L_{\mathrm{RL}}(\theta;F)
+
L_{\mathrm{ret}}(\theta;R), \qquad \lambda > 0,
\]
so the resulting gradient-flow vector field is the sum of the Random Labels and retraining vector fields. This setting reflects the common use of retain fine-tuning to preserve retained-data performance and also underlies several more elaborate unlearning methods.

\begin{lemma}[Symmetry of the RL+FT trajectory]
\label{lem:xor-rl-retain-symmetry}
Random Labels unlearning with simultaneous retain fine-tuning preserves
the invariant
\[
u(t)=Jv(t)
\qquad
\text{for all }t\geq0.
\]
\end{lemma}

\begin{proof}
The invariant holds at initialization since
\[
u(0)=v(0)=\delta e_1.
\]
By Lemma~\ref{lem:xor-retrain-symmetry}, the retraining vector field
preserves the manifold
\[
\mathcal M=\{(u,v):u=Jv\},
\]
and by Lemma~\ref{lem:xor-rl-symmetry}, the Random Labels vector field
preserves the same manifold. Since the combined dynamics are given by
the sum of these two vector fields, their vector field is also tangent
to $\mathcal M$. Hence,
\[
u(t)=Jv(t)
\]
along the entire trajectory.
\end{proof}

\begin{lemma}
\label{lem:xor-rl-retain-dynamics}
Define
\[
G_{\mathrm{RL+FT}}(t)
:=
\int_0^t
\left[
\varepsilon h\bigl(m_R(\tau)\bigr)
+
(1-\varepsilon)\lambda \,\psi\bigl(r_F(\tau)\bigr)
\right]\,d\tau,
\]
where
\[
m_R(t)=a(t)c+2\varepsilon p(t)s(t),
\qquad
r_F(t)=-a(t)c+2(1-\varepsilon)p(t)s(t).
\]
Then
\[
p(t)=\delta\cosh G_{\mathrm{RL+FT}}(t),
\qquad
s(t)=\delta\sinh G_{\mathrm{RL+FT}}(t),
\]
while the shortcut coefficient satisfies
\[
\dot a(t)
=
c\,h\bigl(m_R(t)\bigr)
-
c\,\lambda\,\psi\bigl(r_F(t)\bigr).
\]
\end{lemma}

\begin{proof}
By Lemma~\ref{lem:xor-rl-retain-symmetry}, write
\[
u(t)=
\begin{pmatrix}
p(t)\\s(t)
\end{pmatrix},
\qquad
v(t)=
\begin{pmatrix}
p(t)\\-s(t)
\end{pmatrix}.
\]
The retraining and Random Labels vector fields derived above contribute
\[
\dot p_R=\varepsilon h(m_R)s,
\qquad
\dot s_R=\varepsilon h(m_R)p,
\qquad
\dot a_R=c\,h(m_R),
\]
and
\[
\dot p_{\mathrm{RL}}=(1-\varepsilon)\psi(r_F)s,
\qquad
\dot s_{\mathrm{RL}}=(1-\varepsilon)\psi(r_F)p,
\qquad
\dot a_{\mathrm{RL}}=-c\,\psi(r_F).
\]
Since the simultaneous dynamics are the weighted sum of the two vector fields,
\[
\dot p
=
\left[
\varepsilon h(m_R)
+
(1-\varepsilon)\lambda \,\psi(r_F)
\right]s,
\qquad
\dot s
=
\left[
\varepsilon h(m_R)
+
(1-\varepsilon)\lambda \,\psi(r_F)
\right]p.
\]
Lemma~\ref{lem:hyperbolic-time-change}, together with \(p(0)=\delta\) and \(s(0)=0\), therefore gives
\[
p(t)=\delta\cosh G_{\mathrm{RL+FT}}(t),
\qquad
s(t)=\delta\sinh G_{\mathrm{RL+FT}}(t).
\]
Finally,
\[
\dot a
=
\dot a_R+\lambda\dot a_{\mathrm{RL}}
=
c\,h(m_R)-c\,\lambda\,\psi(r_F).
\]
\end{proof}

\begin{lemma}
\label{lem:xor-rl-retain-convergence}
Along the Random Labels trajectory with retain fine-tuning,
\[
G_{\mathrm{RL+FT}}(t)\to\infty,
\qquad
r_F(t)\to0
\qquad
\text{as }t\to\infty.
\]
\end{lemma}

\begin{proof}
Define
\[
S(t)
=
\frac{1-\varepsilon}{c}a(t)
+
G_{\mathrm{RL+FT}}(t).
\]
By Lemma~\ref{lem:xor-rl-retain-dynamics},
\[
\dot S(t)=h\bigl(m_R(t)\bigr)>0,
\]
so \(S(t)\) is strictly increasing.

We first note that both $a(t)$ and $G_{\mathrm{RL+FT}}(t)$ remain
non-negative along the entire trajectory. Indeed, since $S(0)>0$, if
$G_{\mathrm{RL+FT}}=0$, then $a>0$ and $r_F = - ac < 0$. Therefore $\psi(-ac) > 0$ and 
\[
\dot G_{\mathrm{RL+FT}}
=
\varepsilon h(ac)+(1-\varepsilon)\lambda\,\psi(-ac)>0.
\]
Similarly, if $a=0$, then $G_{\mathrm{RL+FT}}>0$ and $r_F > 0$. Therefore $\psi(r_F) < 0$ and
\[
\dot a
=
c\bigl(h(m_R)-\lambda \,\psi(r_F)\bigr)>0.
\]
Thus, the non-negative quadrant is preserved.

Suppose now that $S(t)$ is bounded. Since both $a(t)$ and
$G_{\mathrm{RL+FT}}(t)$ are non-negative, they are bounded as well.
Therefore,
\[
m_R
=
ac+\varepsilon\delta^2\sinh(2G_{\mathrm{RL+FT}})
\]
is bounded. Hence,
\[
\dot S(t)
=
h(m_R(t))
\ge
h\left(\sup_t m_R(t)\right)
>0,
\]
which implies that $S(t)$ is unbounded, contradicting our assumption.
Therefore,
\[
S(t)\to\infty
\qquad\text{as }t\to\infty.
\]

\textbf{Boundedness of \(r_F(t)\).} We next show that $r_F(t)$ remains bounded. Otherwise, there would exist
a sequence of times $t_k$ such that
\[
|r_F(t_k)|\to\infty.
\]
The Random Labels component of $L_{\mathrm{RL+FT}}$ then diverges, and
hence
\[
L_{\mathrm{RL+FT}}(t_k)\to\infty,
\]
contradicting the monotonicity of gradient flow,
\[
L_{\mathrm{RL+FT}}(t)\le L_{\mathrm{RL+FT}}(0).
\]
Thus, $|r_F(t)|\le C$ for some $C>0$. Boundedness of $r_F$, together with $S(t)\to\infty$, implies
\[
\boxed{
G_{\mathrm{RL+FT}}(t)\to\infty
\qquad\text{and}\qquad
a(t)\to\infty.
}
\]

\textbf{Convergence of $r_F$ to zero.} It remains to show that $r_F(t)\to0$. From the limits above,
\[
m_R(t)
=
a(t)c
+
\varepsilon\delta^2\sinh(2G_{\mathrm{RL+FT}}(t))
\to\infty,
\]
and therefore $h(m_R(t))\to0$. From the dynamics,
\begin{align}
\dot r_F(t)
&=
c^2\Bigl(\lambda \,\psi(r_F(t))-h(m_R(t))\Bigr)
\\
&\quad+
2(1-\varepsilon)\delta^2
\Bigl(
\varepsilon h(m_R(t))
+
(1-\varepsilon)\lambda \,\psi(r_F(t))
\Bigr)
\cosh(2G_{\mathrm{RL+FT}}(t)).
\end{align}

Fix any $\eta>0$. Since $h(m_R(t))\to0$, for all sufficiently large $t$,
\[
\frac\lambda2|\psi\left(\frac{\eta}{2}\right)| > |h(m_R(t))| \ge \frac{\varepsilon}{1 - \varepsilon}|h(m_R(t))|.
\]

From this time onward, whenever
\[
|r_F(t)|\ge \frac{\eta}{2},
\]
the sign of $\dot r_F(t)$ is determined by $\psi(r_F(t))$, and is
therefore opposite to the sign of $r_F(t)$. Moreover,
$|\dot r_F(t)|$ is bounded away from zero at such points. Therefore, for every $\eta>0$, there exists a time after which
$r_F(t)$ remains in $[-\eta,\eta]$. Hence,
\[
\boxed{
r_F(t)\to0.
}
\]
\end{proof}

\begin{theorem}[Asymptotic feature spectrum under RL+FT]
\label{lem:xor-rl-retain-spectrum}
Along the Random Labels trajectory with retain fine-tuning, the normalized feature responses and corresponding PCA directions are those of Proposition~\ref{lem:xor-j-invariant-pca}. Then their strengths satisfy
\[
\frac{\lambda_{\mathrm{s}}(t)}
{\lambda_-(t)}
\longrightarrow
\frac{1-\varepsilon}{2\varepsilon}
\qquad
\text{as }t\to\infty.
\]
\end{theorem}

\begin{proof}
By Proposition~\ref{lem:xor-j-invariant-pca},
\[
\lambda_{\mathrm{s}}
=
a^2c^2,
\qquad
\lambda_-
=
2\varepsilon(1-\varepsilon)\delta^4
\sinh^2(2G_{\mathrm{RL+FT}}),
\]
and hence
\[
\frac{\lambda_{\mathrm{s}}}{\lambda_-}
=
\frac{1-\varepsilon}{2\varepsilon}
\left(
\frac{ac}
{(1-\varepsilon)\delta^2
\sinh(2G_{\mathrm{RL+FT}})}
\right)^2.
\]

By Lemma~\ref{lem:xor-rl-retain-convergence},
\[
r_F
=
-ac
+
(1-\varepsilon)\delta^2
\sinh(2G_{\mathrm{RL+FT}})
\to0,
\qquad
G_{\mathrm{RL+FT}}\to\infty.
\]
Therefore,
\[
\frac{ac}
{(1-\varepsilon)\delta^2
\sinh(2G_{\mathrm{RL+FT}})}
\to1,
\]
which proves the claim.
\end{proof}

\subsection{Gradient Ascent unlearning dynamics}~\label{app:xor-learning-dynamics}

We next analyze Gradient Ascent (GA) without retain fine-tuning, initialized from the full-data solution,
\[
a(0)=A,
\qquad
u(0)=v(0)=\delta e_1,
\]
with \(A>0\). GA reverses the forget-set gradient, yielding
\begin{align}
\dot a(t)
&=
-\mathbb{E}_{(x,y)\sim D_f}
\left[
h\bigl(m_{\theta(t)}(x,y)\bigr)\,ys
\right],
\\
\dot u(t)
&=
-\mathbb{E}_{(x,y)\sim D_f}
\left[
h\bigl(m_{\theta(t)}(x,y)\bigr)\,
y\bigl(u(t)^\top\widetilde z\bigr)\widetilde z
\right],
\\
\dot v(t)
&=
\mathbb{E}_{(x,y)\sim D_f}
\left[
h\bigl(m_{\theta(t)}(x,y)\bigr)\,
y\bigl(v(t)^\top\widetilde z\bigr)\widetilde z
\right].
\end{align}
As in retraining and Random Labels, GA preserves \(u(t)=Jv(t)\). We use this invariant to characterize the trajectory and compare its asymptotic time to retraining.

\begin{lemma}[Symmetry of the Gradient Ascent trajectory]
\label{lem:xor-ga-symmetry}
Consider Gradient Ascent initialized from
\[
a(0)=A,
\qquad
u(0)=v(0)=\delta e_1,
\]
and let
\[
J=
\begin{pmatrix}
1&0\\
0&-1
\end{pmatrix}.
\]
Then
\[
u(t)=Jv(t)
\qquad
\text{for all }t\geq0.
\]
\end{lemma}

\begin{proof}
The relation holds at initialization. Suppose \(u=Jv\), and write
\[
u=
\begin{pmatrix}
p\\
s
\end{pmatrix},
\qquad
v=
\begin{pmatrix}
p\\
-s
\end{pmatrix}.
\]
For every forget example, the margin
\[
m_F
=
yf_\theta(x)
=
ac-2(1-\varepsilon)ps,
\]
so the margin is independent of the particular data point. Moreover,
\[
\mathbb E_{D_f}
\left[
y\widetilde z\widetilde z^\top
\right]
=
-(1-\varepsilon)K,
\qquad
K=
\begin{pmatrix}
0&1\\
1&0
\end{pmatrix}.
\]
Hence, on the manifold \(u=Jv\),
\[
\dot u
=
(1-\varepsilon)h(m_F)Ku,
\qquad
\dot v
=
-(1-\varepsilon)h(m_F)Kv.
\]
Since \(JK=-KJ\),
\[
J\dot v
=
-(1-\varepsilon)h(m_F)JKv
=
(1-\varepsilon)h(m_F)KJv
=
\dot u.
\]
Thus the Gradient Ascent vector field preserves the relation \(u=Jv\). Since it holds at initialization, it holds along the entire trajectory.
\end{proof}

\begin{proposition}[Feature spectrum along the GA trajectory]
\label{lem:xor-ga-pca}
Let
\[
G_{\mathrm{GA}}(t)
=
\int_0^t
(1-\varepsilon)h\bigl(m_F(\tau)\bigr)\,d\tau.
\]
Along the Gradient Ascent trajectory, the normalized feature responses and corresponding PCA directions are those of Proposition~\ref{lem:xor-j-invariant-pca}. Then their strengths are
\begin{align*}
\lambda_{\mathrm{s}}(t)
&=
c^2
\left(
A-\frac{c}{1-\varepsilon}G_{\mathrm{GA}}(t)
\right)^2,
\\
\lambda_{+}(t)
&=
2\varepsilon(1-\varepsilon)(1-2\varepsilon)^2
\delta^4
\cosh^2\bigl(2G_{\mathrm{GA}}(t)\bigr),
\\
\lambda_{-}(t)
&=
2\varepsilon(1-\varepsilon)
\delta^4
\sinh^2\bigl(2G_{\mathrm{GA}}(t)\bigr).
\end{align*}
\end{proposition}

\begin{proof}
By Lemma~\ref{lem:xor-ga-symmetry}, the Gradient Ascent trajectory
satisfies the $J$-invariant. Moreover, applying
Lemma~\ref{lem:hyperbolic-time-change} to the reduced GA dynamics gives
\[
p(t)=\delta\cosh G_{\mathrm{GA}}(t),
\qquad
s(t)=\delta\sinh G_{\mathrm{GA}}(t),
\]
while
\[
a(t)
=
A-\frac{c}{1-\varepsilon}G_{\mathrm{GA}}(t).
\]

The principal directions therefore follow from
Proposition~\ref{lem:xor-j-invariant-pca}. Using
\[
p(t)^2+s(t)^2
=
\delta^2\cosh\bigl(2G_{\mathrm{GA}}(t)\bigr),
\qquad
p(t)s(t)
=
\frac{\delta^2}{2}
\sinh\bigl(2G_{\mathrm{GA}}(t)\bigr),
\]
and substituting these expressions together with $a(t)$ into the
corresponding eigenvalues yields the result.
\end{proof}

Although Gradient Ascent can eventually make the core direction \(q_{-}\) dominate the shortcut, as retraining does, matching this qualitative behavior is not sufficient: unlearning should reach a retraining-like representation substantially faster than retraining itself.

To compare their timescales, let the full model be trained for time \(\tau\), reaching \(a(\tau)=A\), and retrain from the same initialization on \(D_r\) for time \(\tau\). We take its resulting feature ratio
\[
\left.
\frac{\lambda_{-}}{\lambda_{\mathrm{s}}}
\right|_{\mathrm{retr},\,t=\tau}
\]
as the target. Starting GA from the corresponding full model with \(a(0)=A\), let \(\tau_{\mathrm{GA}}\) be the time required to match this ratio:
\[
\left.
\frac{\lambda_{-}}{\lambda_{\mathrm{s}}}
\right|_{\mathrm{GA},\,t=\tau_{\mathrm{GA}}}
=
\left.
\frac{\lambda_{-}}{\lambda_{\mathrm{s}}}
\right|_{\mathrm{retr},\,t=\tau}.
\]
We compare \(\tau_{\mathrm{GA}}\) with \(\tau\) asymptotically as \(A\to\infty\).

\begin{lemma}
\label{lem:xor-full-time}
Let $\tau(A)$ be the time required by full-data training to reach
$a(\tau(A))=A$. Then
\[
\tau(A)
=
\frac{A}{c}
+
\frac{e^{cA}-1}{c^2}.
\]
\end{lemma}

\begin{proof}
By Lemma~\ref{lem:xor-full-core-frozen},
\[
\dot a=\frac{c}{1+e^{ca}},
\qquad
\frac{dt}{da}=\frac{1+e^{ca}}{c}.
\]
Integrating from $a(0)=0$ to $a(\tau(A))=A$ gives
\[
\tau(A)
=
\int_0^A\frac{1+e^{ca}}{c}\,da
=
\frac{A}{c}
+
\frac{e^{cA}-1}{c^2}.
\]
\end{proof}

\begin{lemma}[Unique Gradient Ascent matching point]
\label{lem:xor-ga-matching-point}
Let
\[
G_{\mathrm{retr}}^A
=
G_{\mathrm{retr}}(\tau(A)).
\]
There exists a unique
\[
G_{\mathrm{GA}}^A
\in
\left(
0,\frac{(1-\varepsilon)A}{c}
\right)
\]
such that Gradient Ascent matches the core-to-shortcut ratio achieved by retraining at time \(\tau(A)\). It is uniquely characterized by
\[
\frac{
\sinh\bigl(2G_{\mathrm{GA}}^A\bigr)
}{
A-\dfrac{c}{1-\varepsilon}G_{\mathrm{GA}}^A
}
=
\frac{
\sinh\bigl(2G_{\mathrm{retr}}^A\bigr)
}{
\dfrac{c}{\varepsilon}G_{\mathrm{retr}}^A
}.
\]
Consequently, there is a unique matching time \(\tau_{\mathrm{GA}}\) such that
\[
G_{\mathrm{GA}}(\tau_{\mathrm{GA}})
=
G_{\mathrm{GA}}^A.
\]
\end{lemma}

\begin{proof}
At time \(\tau(A)\), retraining achieves the core-to-shortcut ratio
\[
\mathcal R_{\mathrm{retr}}(A)
=
\frac{
2\varepsilon^3(1-\varepsilon)\delta^4
}{
c^4
}
\frac{
\sinh^2\bigl(2G_{\mathrm{retr}}^A\bigr)
}{
\bigl(G_{\mathrm{retr}}^A\bigr)^2
}.
\]

For Gradient Ascent,
\[
a(t)
=
A-\frac{c}{1-\varepsilon}G_{\mathrm{GA}}(t),
\]
and Proposition~\ref{lem:xor-ga-pca} gives
\[
\mathcal R_{\mathrm{GA}}(t)
=
\frac{
2\varepsilon(1-\varepsilon)\delta^4
\sinh^2\bigl(2G_{\mathrm{GA}}(t)\bigr)
}{
c^2
\left(
A-\dfrac{c}{1-\varepsilon}G_{\mathrm{GA}}(t)
\right)^2
}.
\]

Equating the two ratios and taking the positive square root yields
\begin{align}
\frac{
\sinh\bigl(2G_{\mathrm{GA}}^A\bigr)
}{
A-\dfrac{c}{1-\varepsilon}G_{\mathrm{GA}}^A
}
=
\frac{
\sinh\bigl(2G_{\mathrm{retr}}^A\bigr)
}{
\dfrac{c}{\varepsilon}G_{\mathrm{retr}}^A
}.
\label{eq:xor-ga-target-ratio}
\end{align}

It remains to show that this matching point is unique. Before the shortcut coefficient reaches zero,
\[
0
\leq
G
<
\frac{(1-\varepsilon)A}{c}.
\]
On this interval, define
\[
F_A(G)
:=
\frac{\sinh(2G)}
{
A-\dfrac{c}{1-\varepsilon}G
}.
\]
The numerator is strictly increasing and the denominator strictly decreasing and positive, so \(F_A\) is strictly increasing, with
\[
F_A(0)=0,
\qquad
F_A(G)\to\infty
\quad\text{as}\quad
G\to\frac{(1-\varepsilon)A}{c}^{-}.
\]
Since the right-hand side of~\eqref{eq:xor-ga-target-ratio} is finite and positive, there is a unique
\[
G_{\mathrm{GA}}^A
\in
\left(
0,\frac{(1-\varepsilon)A}{c}
\right)
\]
satisfying the matching condition.

Finally,
\[
\dot G_{\mathrm{GA}}(t)
=
(1-\varepsilon)h(m_F(t))
>0,
\]
so \(G_{\mathrm{GA}}(t)\) is strictly increasing and reaches \(G_{\mathrm{GA}}^A\) at a unique time \(\tau_{\mathrm{GA}}\).
\end{proof}

\begin{lemma}[Training-time representations]
\label{lem:xor-training-times}
Let $\tau(A)$ be the full-data training time required to reach shortcut coefficient $A$, let
\[
G_{\mathrm{retr}}^A
=
G_{\mathrm{retr}}(\tau(A)),
\]
and let $G_{\mathrm{GA}}^A$ be the unique Gradient Ascent matching point from Lemma~\ref{lem:xor-ga-matching-point}. Then
\[
\tau_{\mathrm{GA}}
=
\frac{1}{1-\varepsilon}
\int_0^{G_{\mathrm{GA}}^A}
\left(1+e^{m_{\mathrm{GA}}(G)}\right)\,dG,
\]
where
\[
m_{\mathrm{GA}}(G)
=
Ac
-
\frac{c^2}{1-\varepsilon}G
-
(1-\varepsilon)\delta^2\sinh(2G).
\]
Similarly,
\[
\tau(A)
=
\frac{1}{\varepsilon}
\int_0^{G_{\mathrm{retr}}^A}
\left(1+e^{m_{\mathrm{retr}}(G)}\right)\,dG,
\]
where
\[
m_{\mathrm{retr}}(G)
=
\frac{c^2}{\varepsilon}G
+
\varepsilon\delta^2\sinh(2G).
\]
\end{lemma}

\begin{proof}
Along the Gradient Ascent trajectory,
\[
a
=
A-\frac{c}{1-\varepsilon}G_{\mathrm{GA}},
\qquad
2ps
=
\delta^2\sinh(2G_{\mathrm{GA}}).
\]
Since the forget margin is
\[
m_F
=
ac-2(1-\varepsilon)ps,
\]
we obtain
\[
m_{\mathrm{GA}}(G)
=
Ac
-
\frac{c^2}{1-\varepsilon}G
-
(1-\varepsilon)\delta^2\sinh(2G).
\]
Therefore,
\[
\dot G_{\mathrm{GA}}
=
(1-\varepsilon)h(m_{\mathrm{GA}})
=
\frac{1-\varepsilon}{1+e^{m_{\mathrm{GA}}}},
\]
and hence
\[
\tau_{\mathrm{GA}}
=
\frac{1}{1-\varepsilon}
\int_0^{G_{\mathrm{GA}}^A}
\left(1+e^{m_{\mathrm{GA}}(G)}\right)\,dG.
\]

For retraining, Lemma~\ref{lem:xor-retrain-dynamics} gives
\[
m_{\mathrm{retr}}(G)
=
\frac{c^2}{\varepsilon}G
+
\varepsilon\delta^2\sinh(2G),
\]
and
\[
\dot G_{\mathrm{retr}}
=
\varepsilon h(m_{\mathrm{retr}})
=
\frac{\varepsilon}{1+e^{m_{\mathrm{retr}}}}.
\]
Integrating up to
\(G_{\mathrm{retr}}^A=G_{\mathrm{retr}}(\tau(A))\)
gives
\[
\tau(A)
=
\frac{1}{\varepsilon}
\int_0^{G_{\mathrm{retr}}^A}
\left(1+e^{m_{\mathrm{retr}}(G)}\right)\,dG.
\]
\end{proof}

\begin{lemma}
\label{lem:xor-ga-matching-divergence}
As $A\to\infty$,
\[
G_{\mathrm{GA}}^A\to\infty.
\]
\end{lemma}

\begin{proof}
By Lemma~\ref{lem:xor-full-time},
\[
\tau(A)\to\infty.
\]
Together with the retraining-time representation from Lemma~\ref{lem:xor-training-times}, this forces
\[
G_{\mathrm{retr}}^A\to\infty,
\]
since otherwise the corresponding integral would remain bounded. Hence
\[
\frac{
\sinh\bigl(2G_{\mathrm{retr}}^A\bigr)
}{
\dfrac{c}{\varepsilon}G_{\mathrm{retr}}^A
}
\to\infty.
\]
By the matching condition in Lemma~\ref{lem:xor-ga-matching-point},
\[
\frac{
\sinh\bigl(2G_{\mathrm{GA}}^A\bigr)
}{
A-\dfrac{c}{1-\varepsilon}G_{\mathrm{GA}}^A
}
=
\frac{
\sinh\bigl(2G_{\mathrm{retr}}^A\bigr)
}{
\dfrac{c}{\varepsilon}G_{\mathrm{retr}}^A
},
\]
so the left-hand side must also diverge.

If $G_{\mathrm{GA}}^A$ did not tend to infinity, there would be a subsequence along which $G_{\mathrm{GA}}^A\leq M$ for some finite $M$. Along this subsequence, the numerator remains bounded while
\[
A-\frac{c}{1-\varepsilon}G_{\mathrm{GA}}^A\to\infty,
\]
so the left-hand side tends to zero, a contradiction. Therefore
\[
G_{\mathrm{GA}}^A\to\infty.
\]
\end{proof}

\begin{theorem}[Asymptotic runtime of Gradient Ascent]
\label{thm:xor-ga-runtime}
Let $\tau(A)$ be the time required by full-data training to reach shortcut coefficient $A$, and let $\tau_{\mathrm{GA}}$ be the time required by Gradient Ascent to match the core-to-shortcut feature ratio achieved by retraining after time $\tau(A)$. Then
\[
\lim_{A\to\infty}
\frac{\tau_{\mathrm{GA}}}{\tau(A)}
=
R_{\mathrm{GA}}(\varepsilon,\delta,c),
\]
where
\[
R_{\mathrm{GA}}(\varepsilon,\delta,c)
:=
\int_0^\infty
\exp\left(
-z
-(1-\varepsilon)\delta^2
\sinh\left(
\frac{2(1-\varepsilon)}{c^2}z
\right)
\right)\,dz.
\]
For every fixed $\varepsilon\in(0,1)$, $\delta>0$, and $c>0$,
\[
0<R_{\mathrm{GA}}(\varepsilon,\delta,c)<1.
\]
Moreover,
\[
\lim_{\delta\to0}
R_{\mathrm{GA}}(\varepsilon,\delta,c)=1,
\qquad
\lim_{c\to\infty}
R_{\mathrm{GA}}(\varepsilon,\delta,c)=1.
\]
Thus, the asymptotic runtime of Gradient Ascent can be made arbitrarily close to that of retraining.
\end{theorem}

\begin{proof}
By Lemma~\ref{lem:xor-training-times},
\[
\tau_{\mathrm{GA}}
=
\frac{G_{\mathrm{GA}}^A}{1-\varepsilon}
+
\frac{e^{Ac}}{1-\varepsilon}
\int_0^{G_{\mathrm{GA}}^A}
\exp\left(
-\frac{c^2}{1-\varepsilon}G
-(1-\varepsilon)\delta^2\sinh(2G)
\right)\,dG.
\]
Define
\[
I_\infty
:=
\int_0^\infty
\exp\left(
-\frac{c^2}{1-\varepsilon}G
-(1-\varepsilon)\delta^2\sinh(2G)
\right)\,dG, \qquad I_\infty<\infty.
\]
Since \(G_{\mathrm{GA}}^A\to\infty\) by Lemma~\ref{lem:xor-ga-matching-divergence}, the remaining tail satisfies
\[
0
\leq
I_\infty
-
\int_0^{G_{\mathrm{GA}}^A}
\exp\left(
-\frac{c^2}{1-\varepsilon}G
-(1-\varepsilon)\delta^2\sinh(2G)
\right)\,dG
\leq
\frac{1-\varepsilon}{c^2}
\exp\left(
-\frac{c^2}{1-\varepsilon}G_{\mathrm{GA}}^A
\right)
=
o(1).
\]
Hence
\[
\int_0^{G_{\mathrm{GA}}^A}
\exp\left(
-\frac{c^2}{1-\varepsilon}G
-(1-\varepsilon)\delta^2\sinh(2G)
\right)\,dG
=
I_\infty+o(1).
\]
Moreover,
\[
G_{\mathrm{GA}}^A
<
\frac{1-\varepsilon}{c}A
=
o(e^{Ac}),
\]
and hence
\[
\tau_{\mathrm{GA}}
=
\frac{I_\infty}{1-\varepsilon}e^{Ac}(1+o(1)).
\]

Combining this with Lemma~\ref{lem:xor-full-time},
\[
\tau(A)
=
\frac{e^{Ac}}{c^2}(1+o(1)),
\]
gives
\[
\frac{\tau_{\mathrm{GA}}}{\tau(A)}
\longrightarrow
\frac{c^2}{1-\varepsilon}I_\infty.
\]
With the change of variables
\[
z=\frac{c^2}{1-\varepsilon}G,
\]
this is exactly \(R_{\mathrm{GA}}(\varepsilon,\delta,c)\).

The integrand defining \(R_{\mathrm{GA}}\) is strictly positive and, for \(z>0\),
\[
\exp\left(
-z
-(1-\varepsilon)\delta^2
\sinh\left(
\frac{2(1-\varepsilon)}{c^2}z
\right)
\right)
<
e^{-z}.
\]
Therefore
\[
0<R_{\mathrm{GA}}(\varepsilon,\delta,c)<\int_0^\infty e^{-z}\,dz=1.
\]
Finally, as either \(\delta\to0\) or \(c\to\infty\), the integrand converges pointwise to \(e^{-z}\) and remains bounded by \(e^{-z}\). The dominated convergence theorem therefore yields
\[
R_{\mathrm{GA}}(\varepsilon,\delta,c)\to1
\]
in both limits.
\end{proof}

Theorem~\ref{theorem:ga-time-main} follows by reparameterizing the limit in training time. By Lemma~\ref{lem:xor-full-shortcut}, \(a(T)\) is increasing and \(a(T)\to\infty\), so setting \(A=a(T)\) gives \(\tau(A)=T\) and \(A\to\infty\) as \(T\to\infty\). The claim then follows from Theorem~\ref{thm:xor-ga-runtime} and \(R_{\mathrm{GA}}(\varepsilon,\delta,c)\to1\) as \(\delta\to0\) or \(c\to\infty\).

\begin{corollary}[Fixed feature-ratio targets]
\label{cor:xor-ga-fixed-ratio}
Fix \(r>0\), and let \(\tau_{\mathrm{GA}}^{(r)}(A)\) be the first time such that
\[
\frac{\lambda_-}{\lambda_{\mathrm{s}}}=r.
\]
Then
\[
\frac{\tau_{\mathrm{GA}}^{(r)}(A)}{\tau(A)}
\longrightarrow
R_{\mathrm{GA}}(\varepsilon,\delta,c)
\qquad
\text{as }A\to\infty.
\]
Thus, the leading-order GA runtime is independent of the fixed target ratio \(r\).
\end{corollary}

\begin{proof}
The argument is identical to that of Theorem~\ref{thm:xor-ga-runtime}. For any fixed \(r>0\), the corresponding matching point \(G_r^A\) satisfies \(G_r^A\to\infty\) and \(G_r^A=O(A)\), so the same asymptotic estimates apply.
\end{proof}

\section{Additional Results}\label{app:additional-results}
\subsection{Experimental Details}\label{app:experimental-details}

Full and retrained models use the same training recipe and differ only in the training data. For CIFAR-100~\citep{krizhevsky2009learning}, we fine-tune an ImageNet-21k-pretrained ViT-Tiny~\citep{DBLP:journals/corr/abs-2010-11929} for 50 epochs with AdamW ($\mathrm{lr}=5\times10^{-5}$, weight decay $0.05$), 5-epoch warmup, cosine decay, and batch size 128. On SVHN, ViT-Tiny is trained from scratch for 100 epochs with AdamW ($\mathrm{lr}=10^{-3}$, weight decay $0.05$), cosine decay, and batch size 256. ResNet-56~\citep{He2015DeepRL} is trained on CIFAR-100 for 200 epochs with SGD ($\mathrm{lr}=0.1$, momentum $0.9$, weight decay $5\times10^{-4}$), cosine decay, and batch size 256.

For each method, we tune only the main hyperparameters with a small manual sweep. Across FT, AMUN, NegGrad+, RL, SalUn, LTD, and ReprUnl, learning rates are swept over method-specific grids within $10^{-4}$--$10^{-1}$; AMUN, RL, and SalUn use $\lambda\in[1,5]$, while NegGrad+ uses $\lambda\in[0.01,1]$. IU sweeps $\alpha\in[0.3,50]$, and ReprUnl additionally sweeps $\beta\in[1,15]$. For epoch-based methods, we use 5 unlearning epochs on ViT-Tiny/CIFAR-100 (versus 50 training epochs) and 10 epochs in the remaining settings; ReprUnl uses 300 update steps.

Each model set contains 20 independently trained seeds. We compute feature metrics for three independently constructed sets and report the mean and standard deviation across them in the scatter plots.

For each setting, we choose \(K\) so that the top-\(K\) principal components capture at least \(80\%\) of the total feature mass. This gives \(K=75\) for ViT-Tiny/CIFAR-100, \(K=20\) for ViT-Tiny/SVHN, and \(K=30\) for both ResNet-56 settings. For head fine-tuning, we freeze the model body and train a fresh linear head for 30 epochs with Adam (\(\mathrm{lr}=10^{-3}\), no weight decay).

\subsection{Standard Metrics}\label{app:standard-metrics}\label{app:head-reset}
Table~\ref{tab:standard-accuracy} reports standard post-unlearning accuracies for the main ViT-Tiny/CIFAR-100 setting. Most methods perform well under these metrics: they reduce forget accuracy close to the retraining baseline while largely preserving performance on the retained data. IU is the clear exception, leaving both forget and test accuracy almost unchanged from the full-data model. Thus, methods that perform well on these standard metrics can still differ substantially in how closely their learned features match those of retraining (Figure~\ref{fig:feature-recovery}).

The adaptation results in Figure~\ref{fig:feature-vs-adaptation} further illustrate this distinction. ReprUnl has the smallest mean absolute accuracy gap from retraining across the three standard metrics, yet exhibits the largest feature discrepancy on the forget set and one of the largest accuracy gaps after partial relearning. Conversely, IU has the largest mean absolute gap under the standard metrics but remains closest to retraining among unlearning methods in downstream accuracy on both transfer tasks.

\begin{table}[h!]
\caption{Standard post-unlearning accuracy. Values are mean $\pm$ std over 5 runs, in $\%$. Blue values in brackets show deviation from Retrain in percentage points.}
\label{tab:standard-accuracy}
\centering
\small
\setlength{\tabcolsep}{6pt}
\scalebox{0.95}{
\begin{tabular}{lccc}
\multicolumn{1}{c}{\bf Method} &
\multicolumn{1}{c}{\bf Retain Acc.} &
\multicolumn{1}{c}{\bf Forget Acc.} &
\multicolumn{1}{c}{\bf Test Acc.}
\\ \toprule
Retrain          & $100.0 \pm 0.0$ & $0.0 \pm 0.0$ & $70.8 \pm 0.1$ \\
\midrule
FT               & $94.1 \pm 0.7$ \diff{-5.9}   & $0.3 \pm 0.2$ \diff{+0.3}     & $65.0 \pm 0.6$ \diff{-5.8} \\
AMUN             & $93.7 \pm 0.4$ \diff{-6.2}   & $2.1 \pm 1.4$ \diff{+2.1}     & $65.4 \pm 0.5$ \diff{-5.4} \\
NegGrad+               & $97.2 \pm 0.2$ \diff{-2.8}   & $0.0 \pm 0.0$ \diff{0.0}      & $67.4 \pm 0.3$ \diff{-3.4} \\
IU               & $100.0 \pm 0.0$ \diff{0.0}   & $100.0 \pm 0.0$ \diff{+100.0} & $86.7 \pm 0.2$ \diff{+15.9} \\
LTD              & $97.4 \pm 0.3$ \diff{-2.5}   & $0.0 \pm 0.0$ \diff{0.0}      & $65.9 \pm 0.2$ \diff{-4.9} \\
RL               & $93.3 \pm 0.3$ \diff{-6.7}   & $0.4 \pm 0.1$ \diff{+0.4}     & $65.1 \pm 0.3$ \diff{-5.7} \\
Rep. Unl.        & $100.0 \pm 0.0$ \diff{0.0}   & $2.8 \pm 0.5$ \diff{+2.8}     & $70.6 \pm 0.2$ \diff{-0.2} \\
SalUn            & $94.9 \pm 0.4$ \diff{-5.0}   & $0.9 \pm 0.2$ \diff{+0.9}     & $66.3 \pm 0.3$ \diff{-4.5} \\
\bottomrule
\end{tabular}}
\end{table}

To separate the learned representation from the particular prediction head produced by unlearning, we discard the original head, freeze the representation, and train a fresh head on the retain set. Table~\ref{tab:head-reset-accuracy} shows that this largely removes the remaining output-level differences on the forget set: all methods achieve essentially zero forget accuracy, including IU, whose original head preserves the forgotten behavior almost entirely. Retain accuracy is also nearly saturated across methods, although some differences remain on the test set.

\begin{table}[h!]
\caption{Accuracy after training a fresh prediction head on the frozen representation. Values are mean $\pm$ std, in $\%$. Blue values in brackets show deviation from Retrain in percentage points.}
\label{tab:head-reset-accuracy}
\centering
\small
\setlength{\tabcolsep}{6pt}
\scalebox{0.95}{
\begin{tabular}{lccc}
\multicolumn{1}{c}{\bf Method} &
\multicolumn{1}{c}{\bf Retain Acc.} &
\multicolumn{1}{c}{\bf Forget Acc.} &
\multicolumn{1}{c}{\bf Test Acc.}
\\ \toprule
Retrain   & $100.0 \pm 0.0$ & $0.0 \pm 0.0$ & $70.8 \pm 0.1$ \\
\midrule
FT        & $99.7 \pm 0.0$ \diff{-0.3} & $0.0 \pm 0.0$ \diff{0.0} & $66.7 \pm 0.4$ \diff{-4.1} \\
AMUN      & $99.6 \pm 0.1$ \diff{-0.4} & $0.0 \pm 0.0$ \diff{0.0} & $66.7 \pm 0.3$ \diff{-4.2} \\
NegGrad+        & $99.8 \pm 0.1$ \diff{-0.1} & $0.0 \pm 0.0$ \diff{0.0} & $68.8 \pm 0.3$ \diff{-2.0} \\
IU        & $100.0 \pm 0.0$ \diff{0.0} & $0.0 \pm 0.0$ \diff{0.0} & $70.3 \pm 0.2$ \diff{-0.5} \\
LTD       & $100.0 \pm 0.0$ \diff{0.0} & $0.0 \pm 0.0$ \diff{0.0} & $68.1 \pm 0.3$ \diff{-2.7} \\
RL        & $99.0 \pm 0.2$ \diff{-1.0} & $0.0 \pm 0.0$ \diff{0.0} & $66.6 \pm 0.3$ \diff{-4.2} \\
Rep. Unl. & $100.0 \pm 0.0$ \diff{0.0} & $0.0 \pm 0.0$ \diff{0.0} & $70.4 \pm 0.2$ \diff{-0.4} \\
SalUn     & $99.6 \pm 0.1$ \diff{-0.4} & $0.0 \pm 0.0$ \diff{0.0} & $67.4 \pm 0.2$ \diff{-3.4} \\
\bottomrule
\end{tabular}}
\end{table}

More generally, retain-only head training makes the methods appear nearly identical to retraining on the forget set, despite the feature mismatches identified in our analysis. This motivates the subsequent-use experiments in Section~\ref{sec:consequences}, where partial relearning and downstream adaptation expose differences encoded in the frozen representation.

\subsection{Directional Discrepancies}\label{app:directional-dicrepancy}
\begin{figure}[h!]
    \centering
    \begin{subfigure}{0.45\linewidth}
        \centering
        \includegraphics[width=\linewidth]{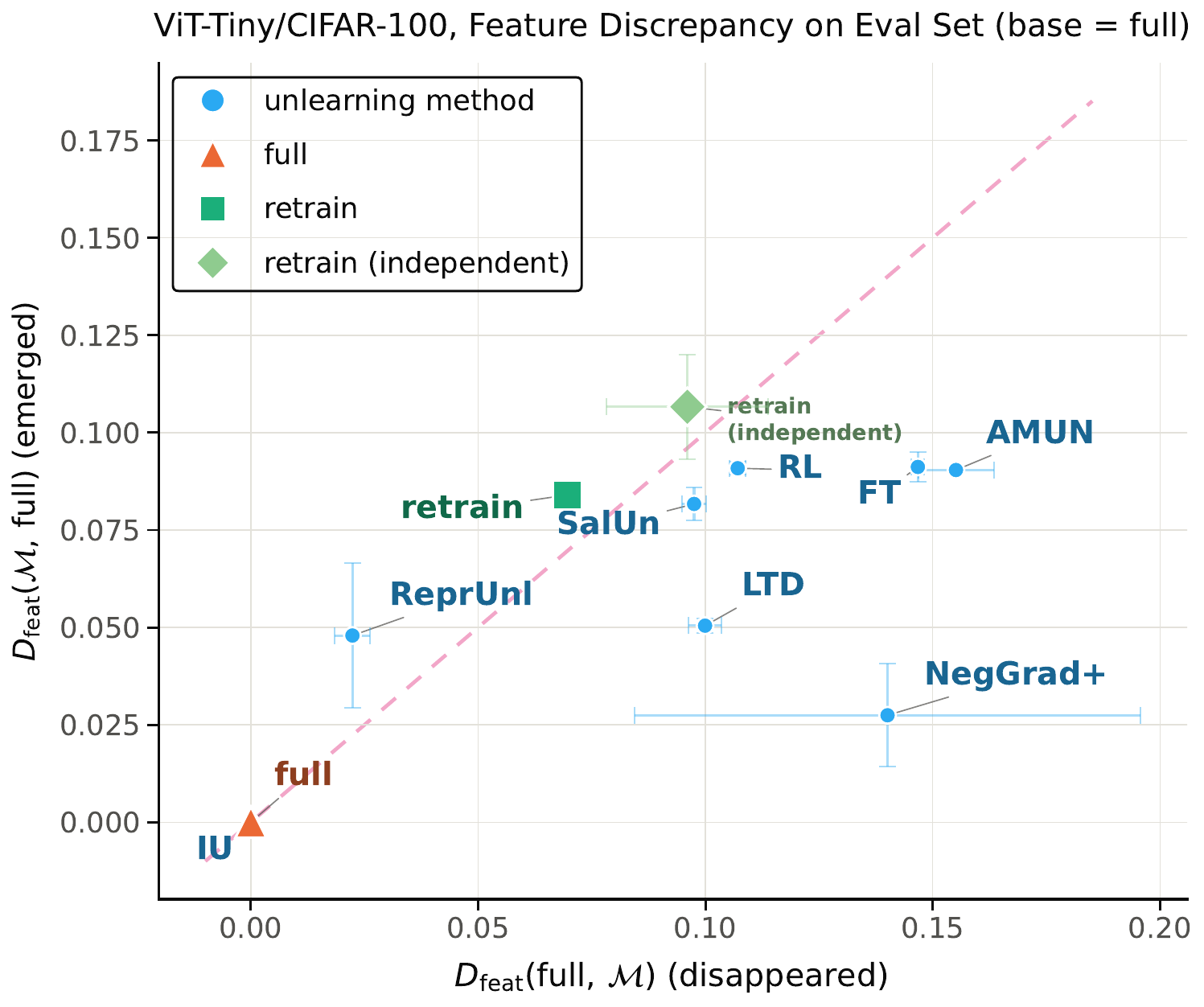}
    \end{subfigure}
    \hfill
    \begin{subfigure}{0.45\linewidth}
        \centering
        \includegraphics[width=\linewidth]{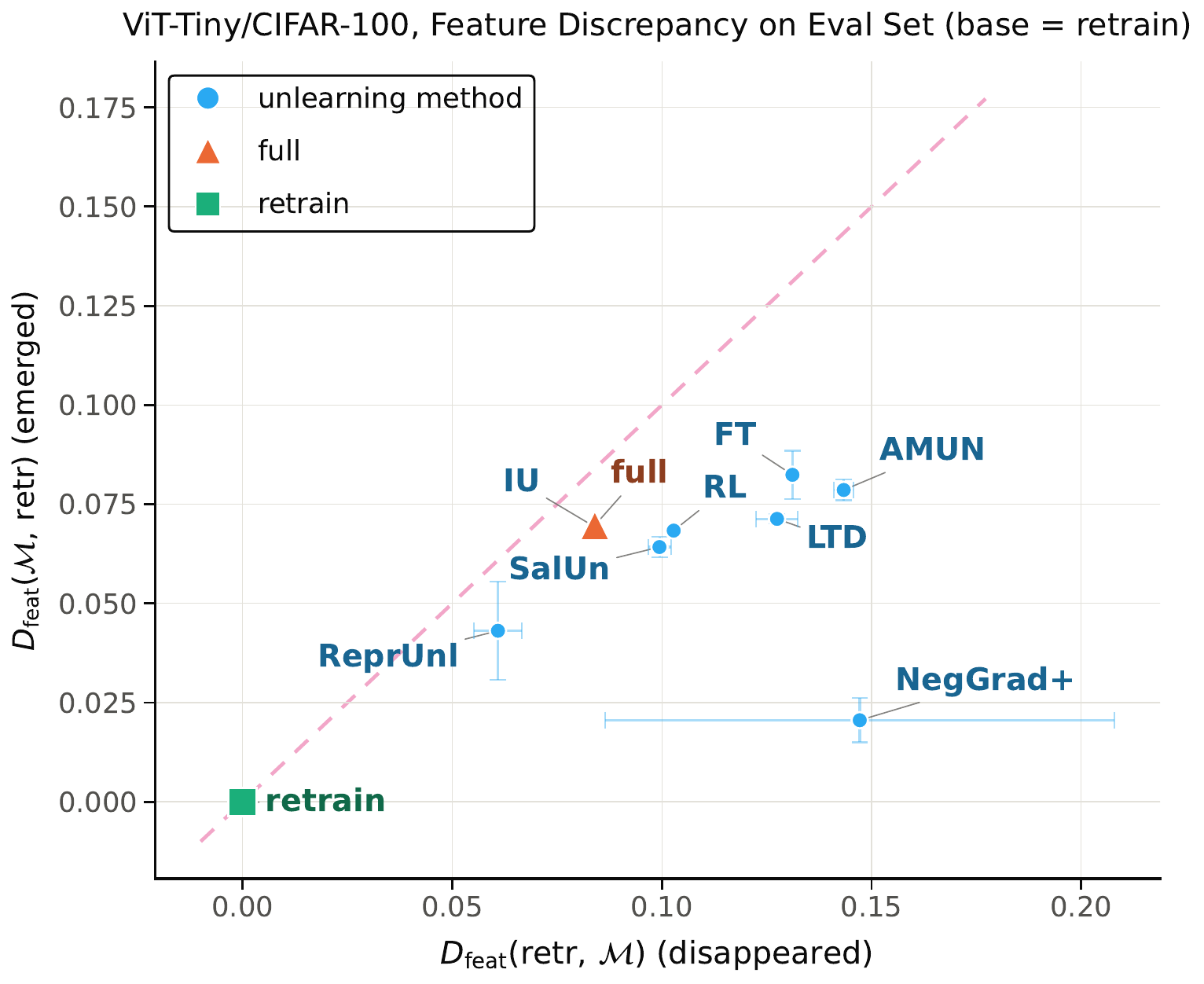}
    \end{subfigure}
     \caption{Directional feature discrepancies on ViT-Tiny/CIFAR-100, using the full-data (left) and retrained (right) solutions as reference.}
    \label{fig:directional-discrepancy}
\end{figure}

Figure~\ref{fig:directional-discrepancy} separates the two directional terms of our feature discrepancy. For the full and retrained solutions, the two directions are comparable: retraining both loses features present in the full model and gains features absent from it. Thus, the feature change induced by deletion is genuinely bidirectional.

\subsection{Additional Settings}\label{app:other-settings}
We complement the main-text analysis with three additional settings spanning a different architecture, a smaller feature-targeted deletion set, and a different dataset. These experiments test whether the qualitative patterns observed in the main setting persist beyond the specific ViT-Tiny/CIFAR-100 configuration. We do not assume that retraining changes the feature solution for every deletion request; some deletions may leave it essentially unchanged. Rather, the settings below test whether nontrivial full-to-retraining feature differences arise beyond the main configuration, and whether the same qualitative consequences persist when they do.

\begin{figure}[h]
    \centering
    \begin{subfigure}{0.45\linewidth}
        \centering
        \includegraphics[width=\linewidth]{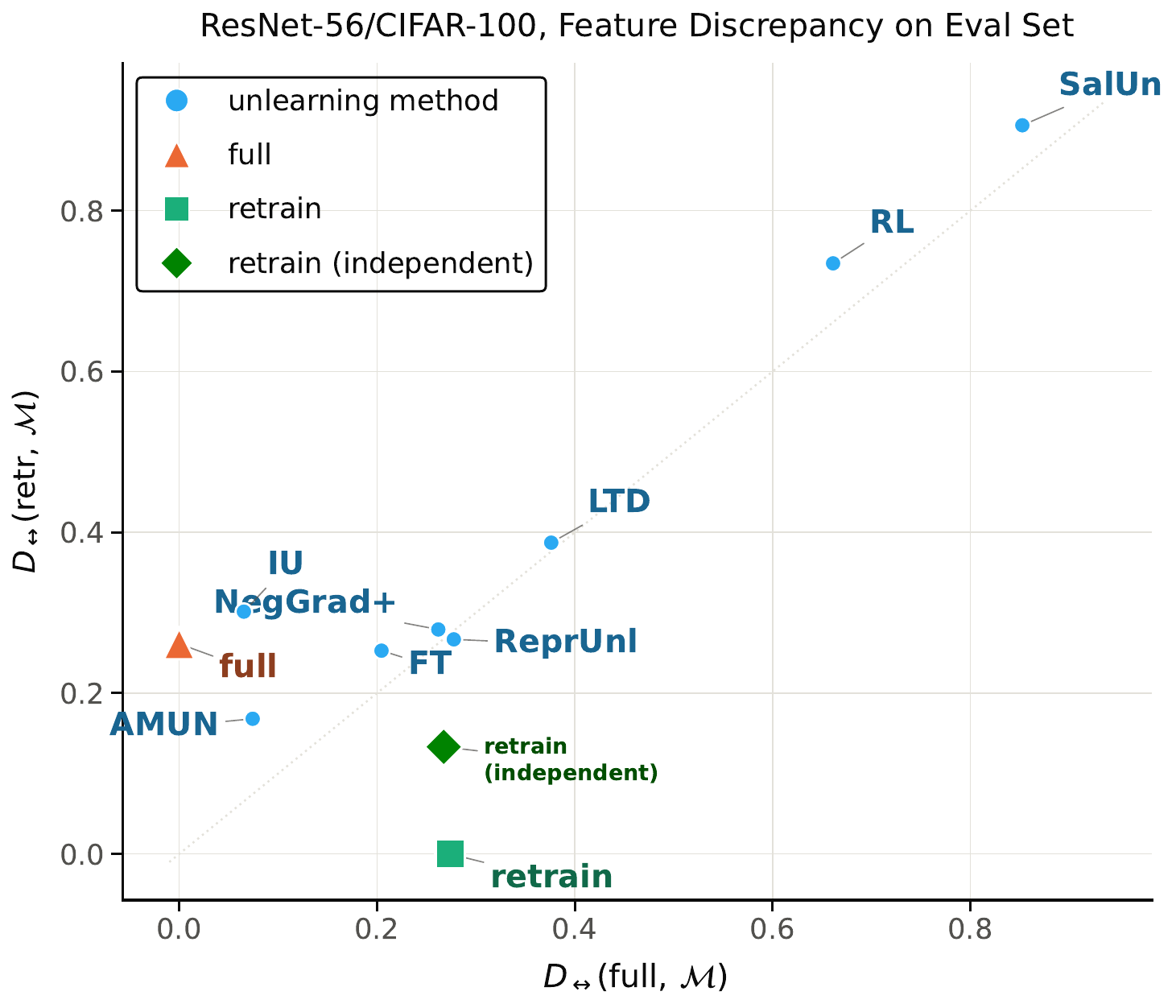}
        \caption{Test Set}
    \end{subfigure}
    \hfill
    \begin{subfigure}{0.45\linewidth}
        \centering
        \includegraphics[width=\linewidth]{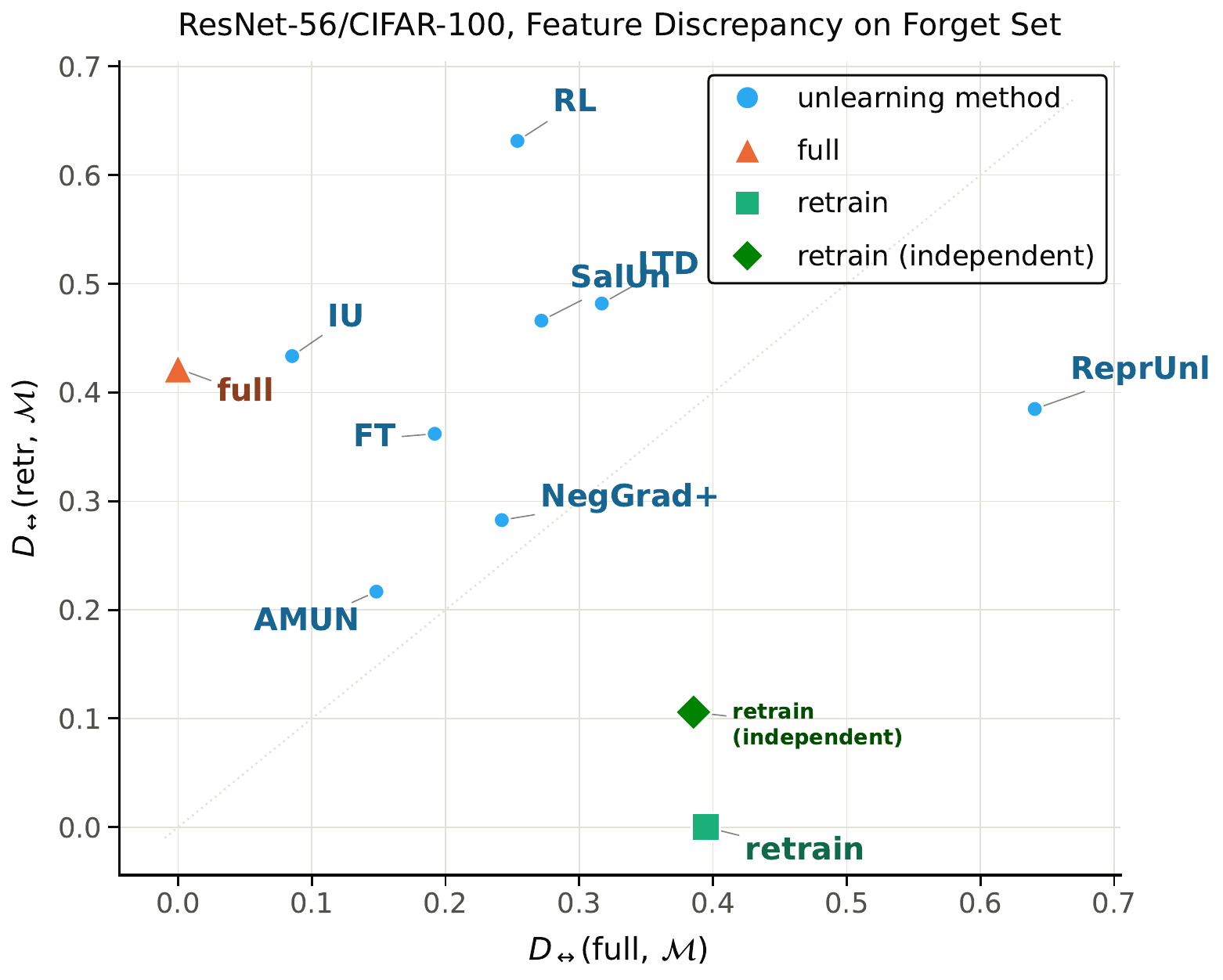}
        \caption{Forget Set}
    \end{subfigure}
    \caption{Feature discrepancy for ResNet-56 on CIFAR-100 under class-wise deletion. Each panel compares unlearned models with the full-data and retrained feature solutions on (a) the test distribution and (b) the forget distribution.}
    \label{fig:feature-recovery-resnet-perclass}
\end{figure}

\begin{figure*}[h!]
    \centering
    \begin{subfigure}{0.32\textwidth}
        \centering
        \includegraphics[width=\linewidth]{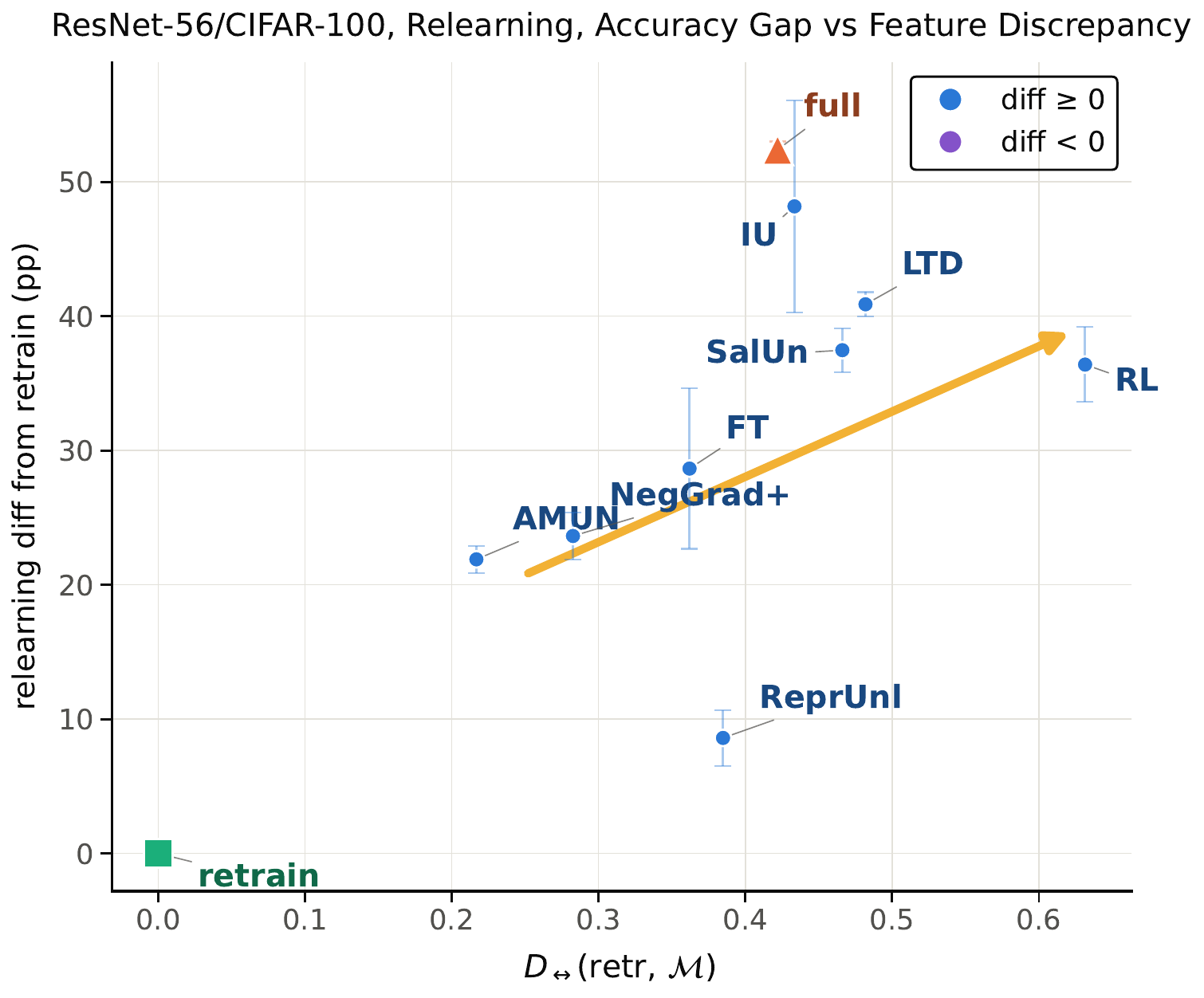}
        \caption{Partial relearning}
    \end{subfigure}
    \hfill
    \begin{subfigure}{0.32\textwidth}
        \centering
        \includegraphics[width=\linewidth]{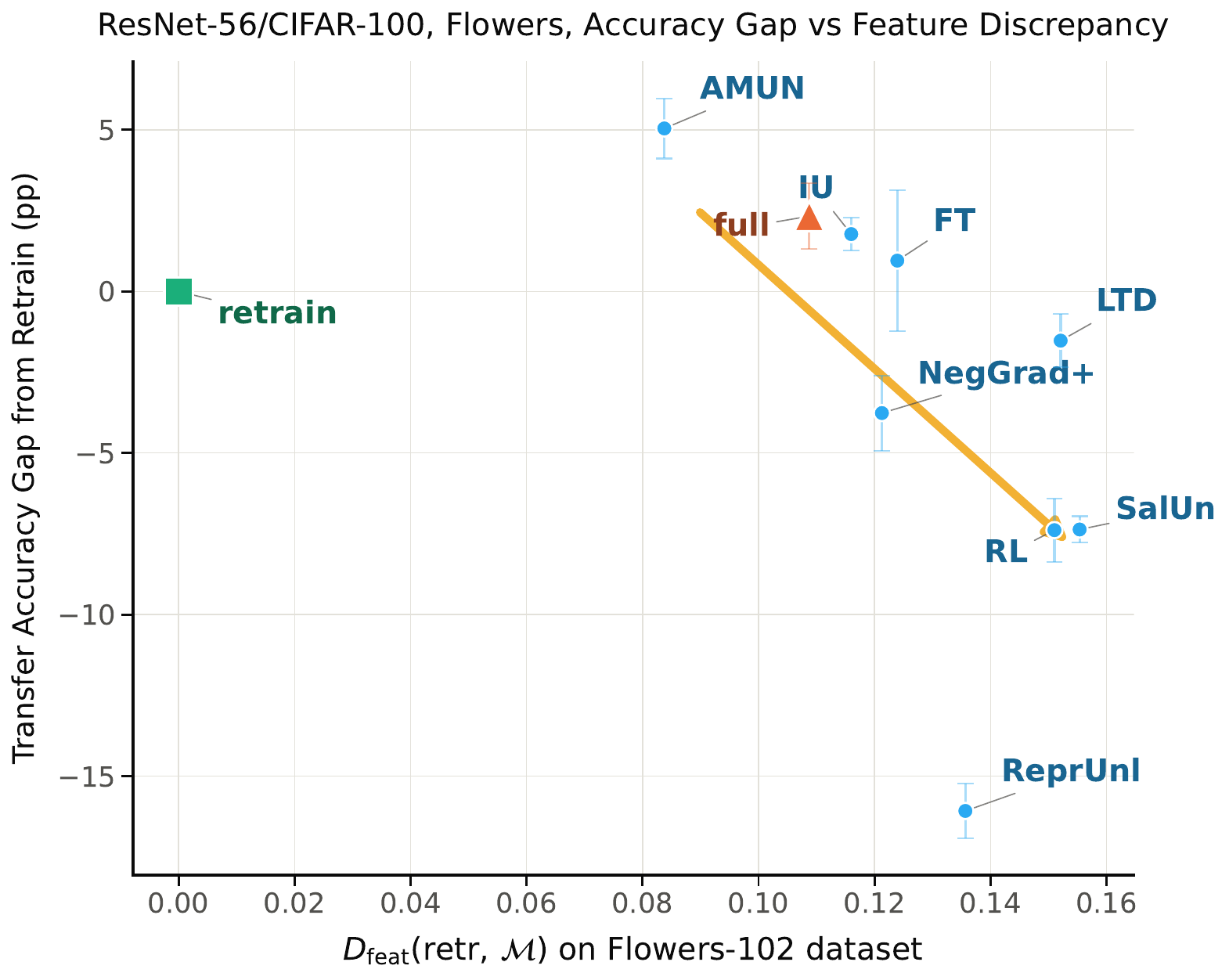}
        \caption{Flowers-102}
    \end{subfigure}
    \hfill
    \begin{subfigure}{0.32\textwidth}
        \centering
        \includegraphics[width=\linewidth]{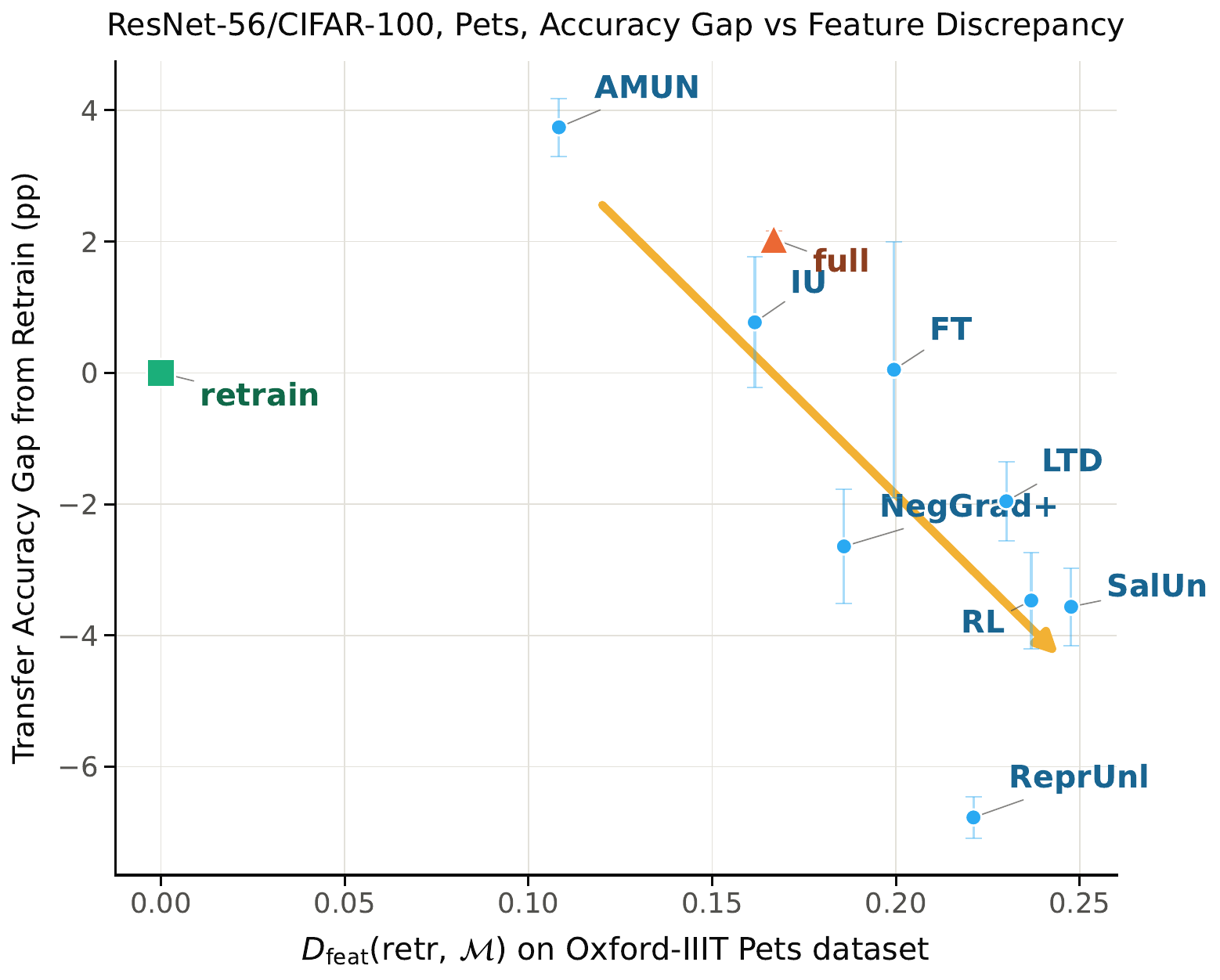}
        \caption{Oxford-IIIT Pets}
    \end{subfigure}
    \caption{Feature mismatch predicts behavior under subsequent adaptation. (a) Retraining-relative feature discrepancy on the forget distribution versus the absolute forget-accuracy gap after partial relearning; colors show the sign of the gap. (b--c) Missing retraining features on downstream inputs versus downstream performance on Flowers-102 and Oxford-IIIT Pets.}
    \label{fig:feature-vs-adaptation-resnet-perclass}
\end{figure*}

\paragraph{ResNet-56 on CIFAR-100: class-wise deletion.} We first repeat the main CIFAR-100 deletion setting with a ResNet-56, removing one fine-grained class from each of the 20 superclasses. Figure~\ref{fig:feature-recovery-resnet-perclass} shows the feature discrepancies on the test and forget distributions. As in the main ViT-Tiny experiment, methods with similar standard unlearning performance can occupy markedly different positions relative to the full-data and retrained feature solutions. Figure~\ref{fig:feature-vs-adaptation-resnet-perclass} reports the corresponding head fine-tuning results.

\paragraph{ResNet-56 on CIFAR-100: blue-green shortcut deletion.} 
We next consider a substantially smaller, feature-targeted deletion set. We select 20 fine-grained CIFAR-100 classes belonging to four superclasses: aquatic mammals and fish (the \emph{marine} group), and trees and flowers (the \emph{plant} group). Within these classes, an example is placed in the forget set only when its image border exhibits the corresponding color shortcut: blue for marine examples and green for plant examples. Concretely, we compute the mean RGB values over a five-pixel image border and remove marine examples whose blue channel exceeds both red and green by more than 10, and plant examples whose green channel exceeds both red and blue by more than 10. 

\begin{figure}[t]
    \centering
    \begin{subfigure}{0.45\linewidth}
        \centering
        \includegraphics[width=\linewidth]{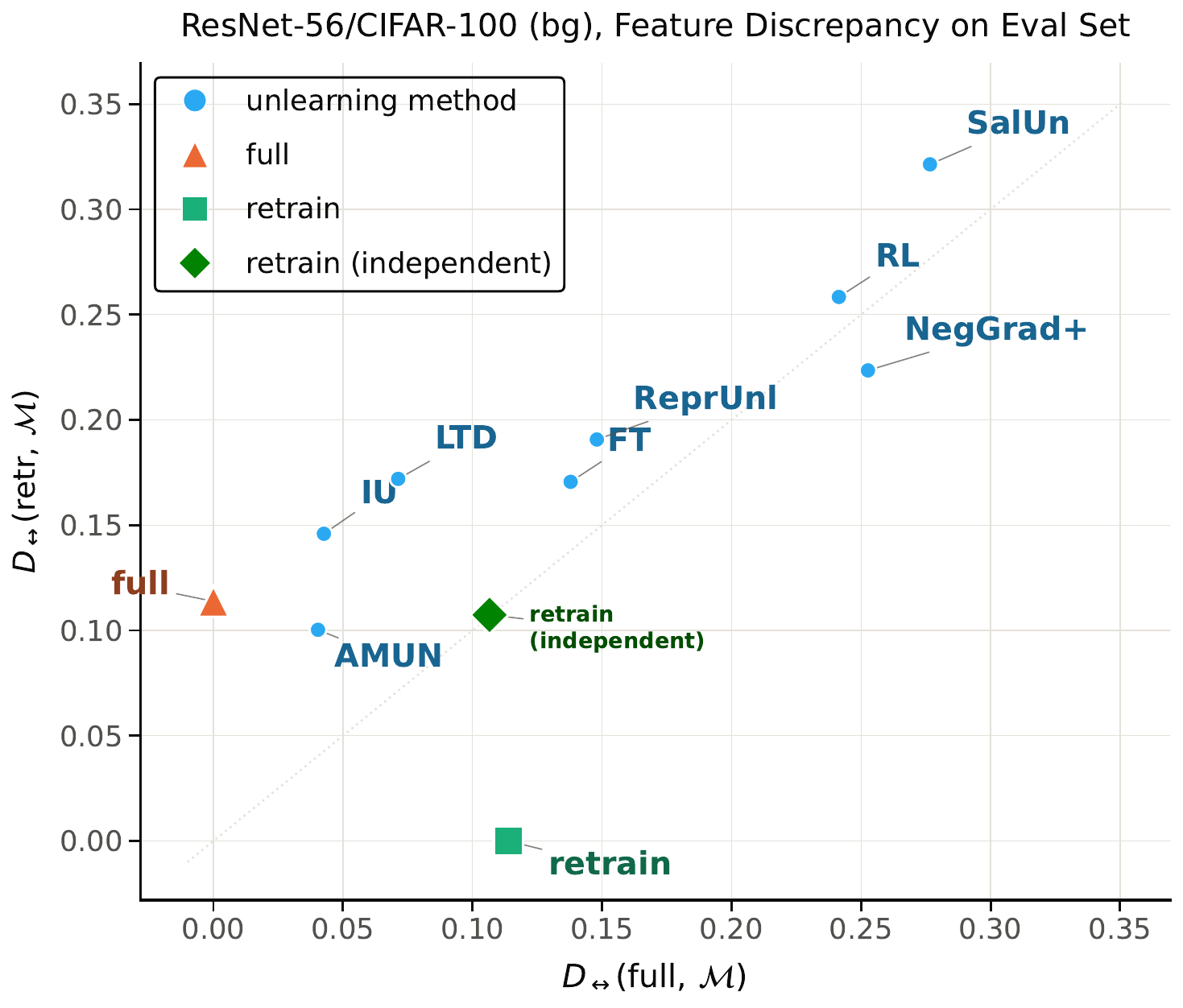}
        \caption{Test Set}
    \end{subfigure}
    \hfill
    \begin{subfigure}{0.45\linewidth}
        \centering
        \includegraphics[width=\linewidth]{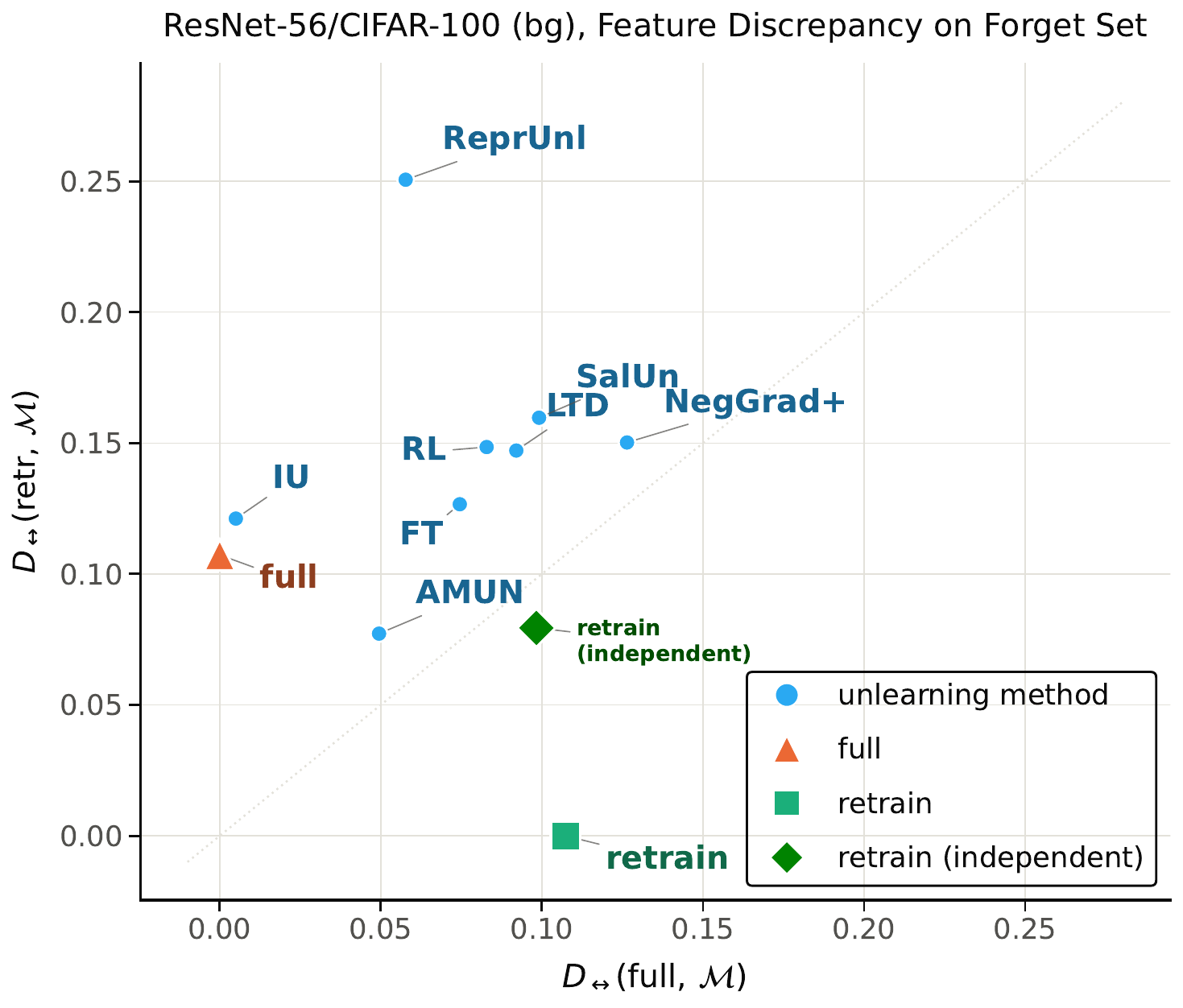}
        \caption{Forget Set}
    \end{subfigure}
    \caption{Feature discrepancy for ResNet-56 on CIFAR-100 under blue-green deletion. Each panel compares unlearned models with the full-data and retrained feature solutions on (a) the test distribution and (b) the forget distribution.}
    \label{fig:feature-recovery-resnet-bluegreen}
\end{figure}

\begin{figure*}[t]
    \centering
    \begin{subfigure}{0.32\textwidth}
        \centering
        \includegraphics[width=\linewidth]{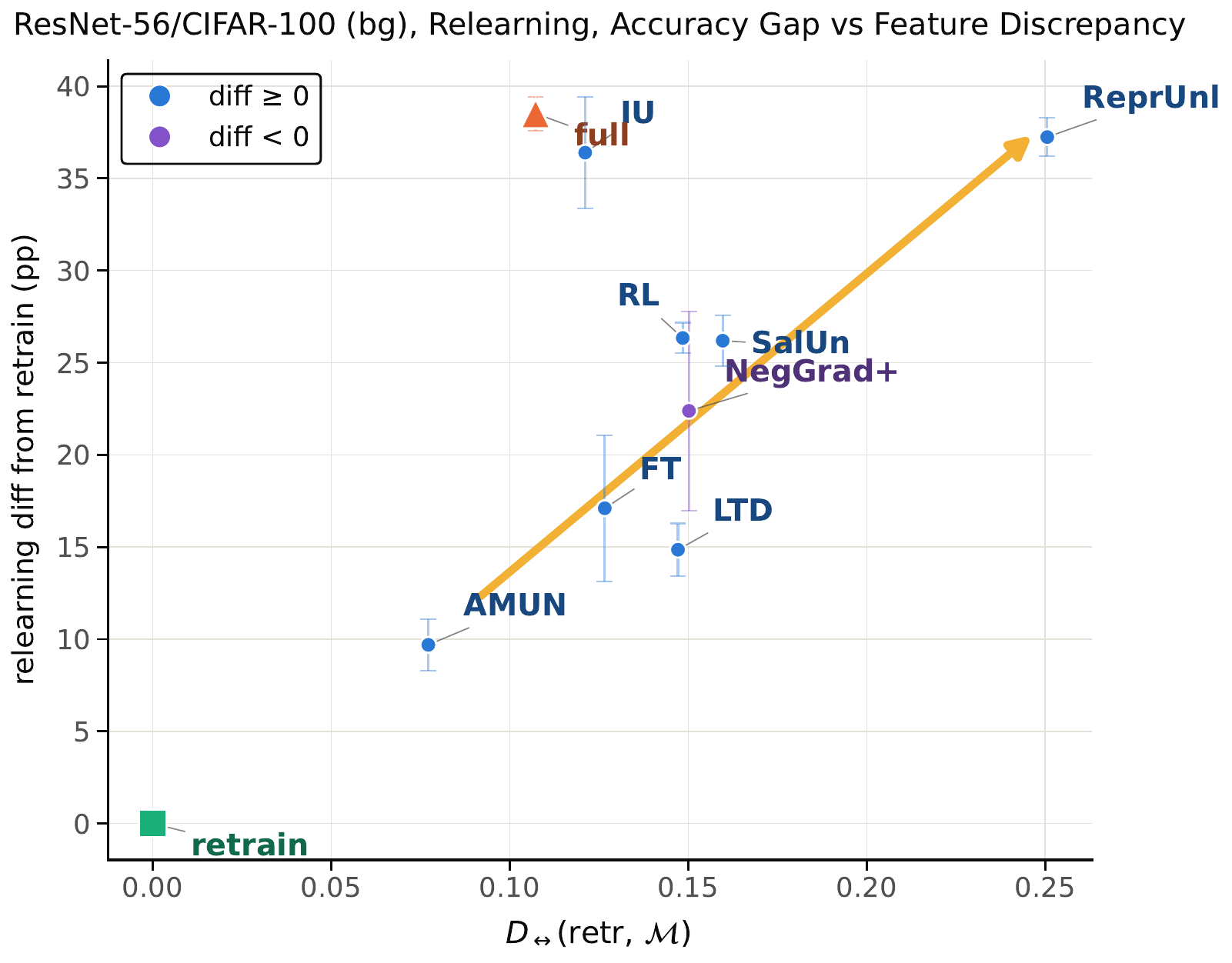}
        \caption{Partial relearning}
        %\label{fig:feature-vs-halfforget}
    \end{subfigure}
    \hfill
    \begin{subfigure}{0.32\textwidth}
        \centering
        \includegraphics[width=\linewidth]{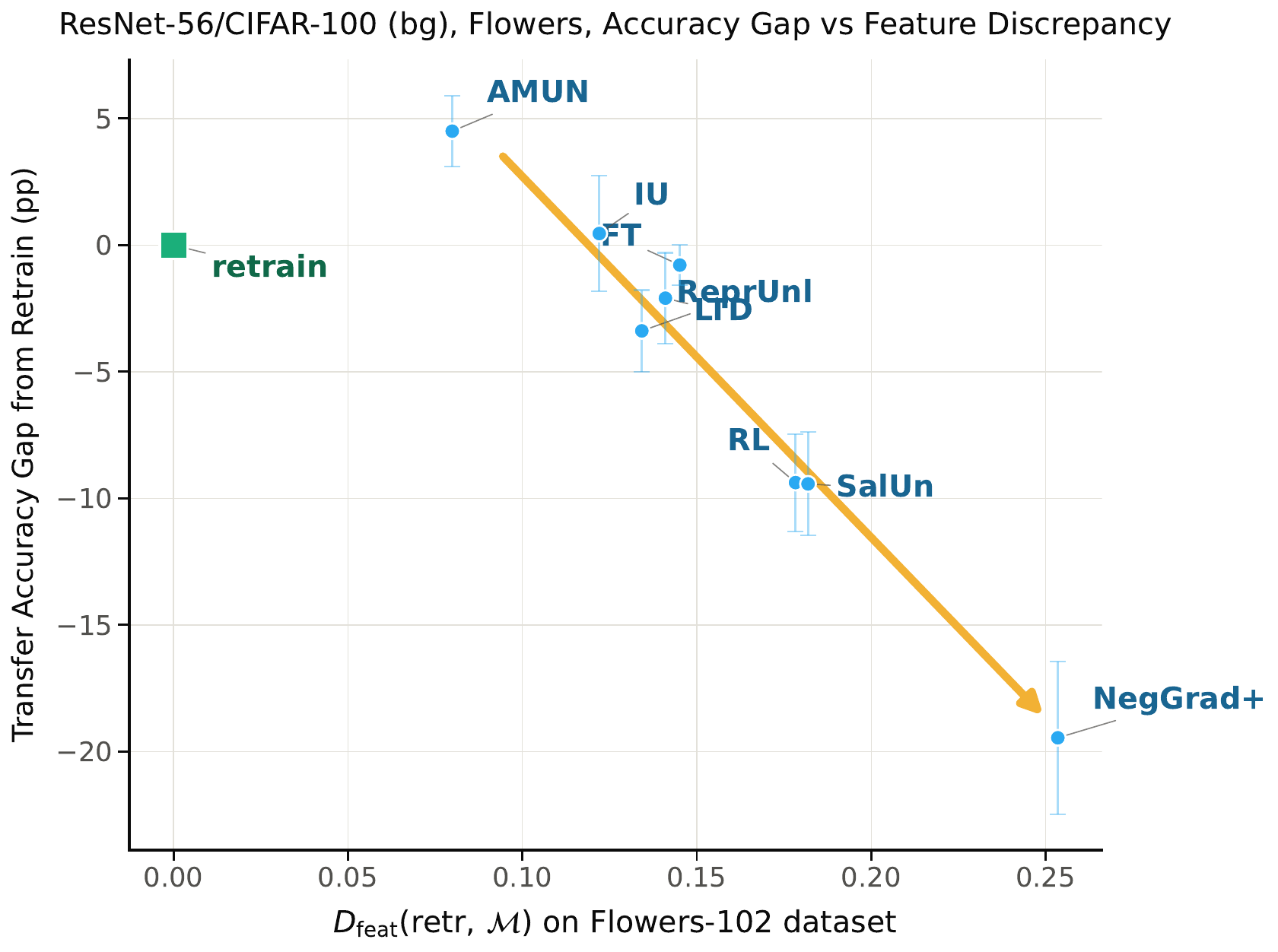}
        \caption{Flowers-102}
        %\label{fig:feature-vs-flowers}
    \end{subfigure}
    \hfill
    \begin{subfigure}{0.32\textwidth}
        \centering
        \includegraphics[width=\linewidth]{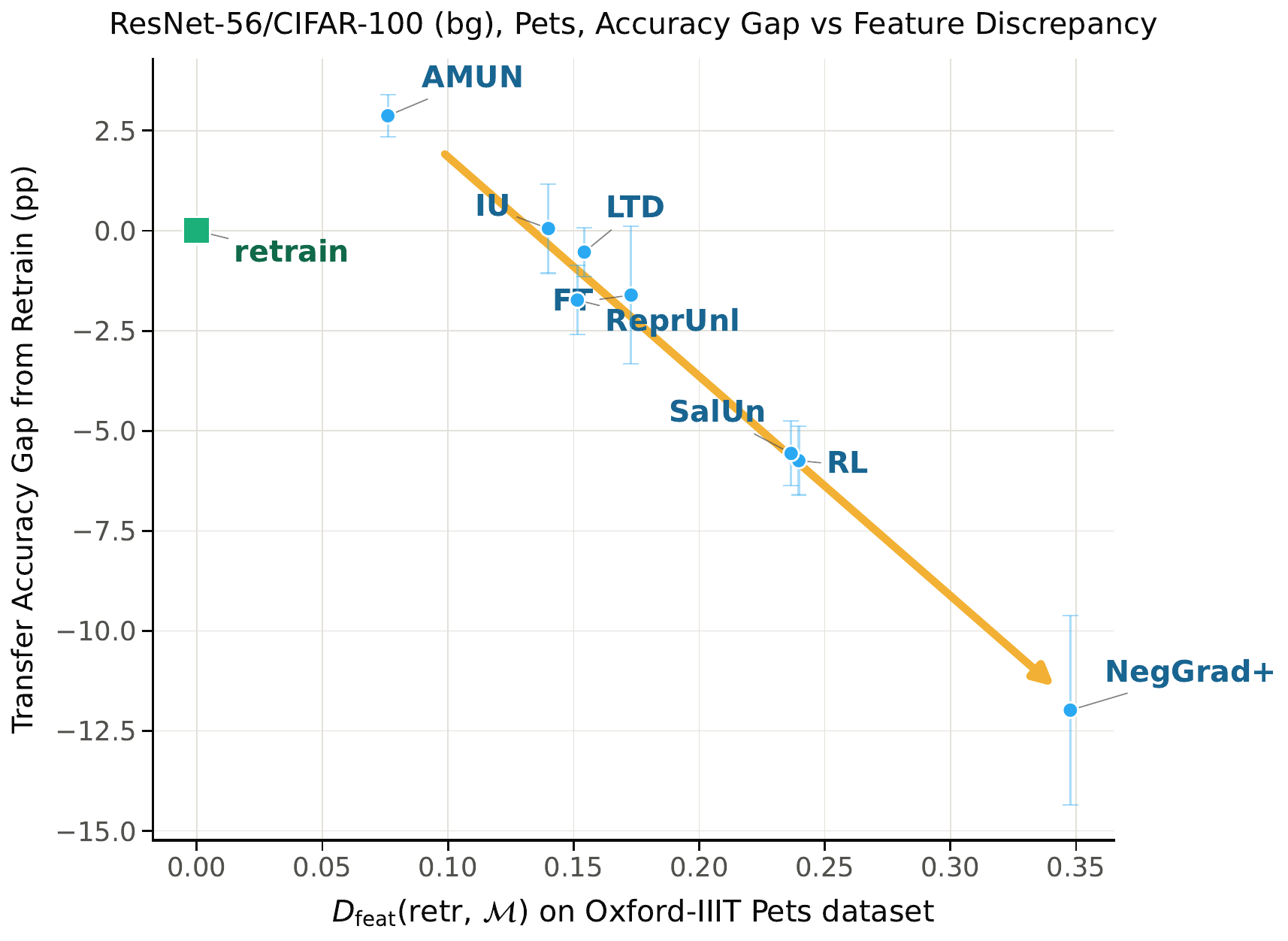}
        \caption{Oxford-IIIT Pets}
        %\label{fig:feature-vs-pets}
    \end{subfigure}
    \caption{Feature mismatch predicts behavior under subsequent adaptation. (a) Retraining-relative feature discrepancy on the forget distribution versus the absolute forget-accuracy gap after partial relearning; colors show the sign of the gap. (b--c) Missing retraining features on downstream inputs versus downstream performance on Flowers-102 and Oxford-IIIT Pets.}
    \label{fig:feature-vs-adaptation-resnet-bluegreen}
\end{figure*}

This yields a forget set containing only 4.64\% of the 50,000 training examples, distributed across the 20 classes. Unlike class-wise deletion, this setting targets a small set of examples sharing a simple, human-interpretable feature, allowing us to ask whether unlearning removes the corresponding feature structure rather than merely changing class-level behavior. Figure~\ref{fig:feature-recovery-resnet-bluegreen} shows the feature discrepancies on the test and forget distributions, and Figure~\ref{fig:feature-vs-adaptation-resnet-bluegreen} reports the head fine-tuning results.

\begin{figure}[t]
    \centering
    \begin{subfigure}{0.45\linewidth}
        \centering
        \includegraphics[width=\linewidth]{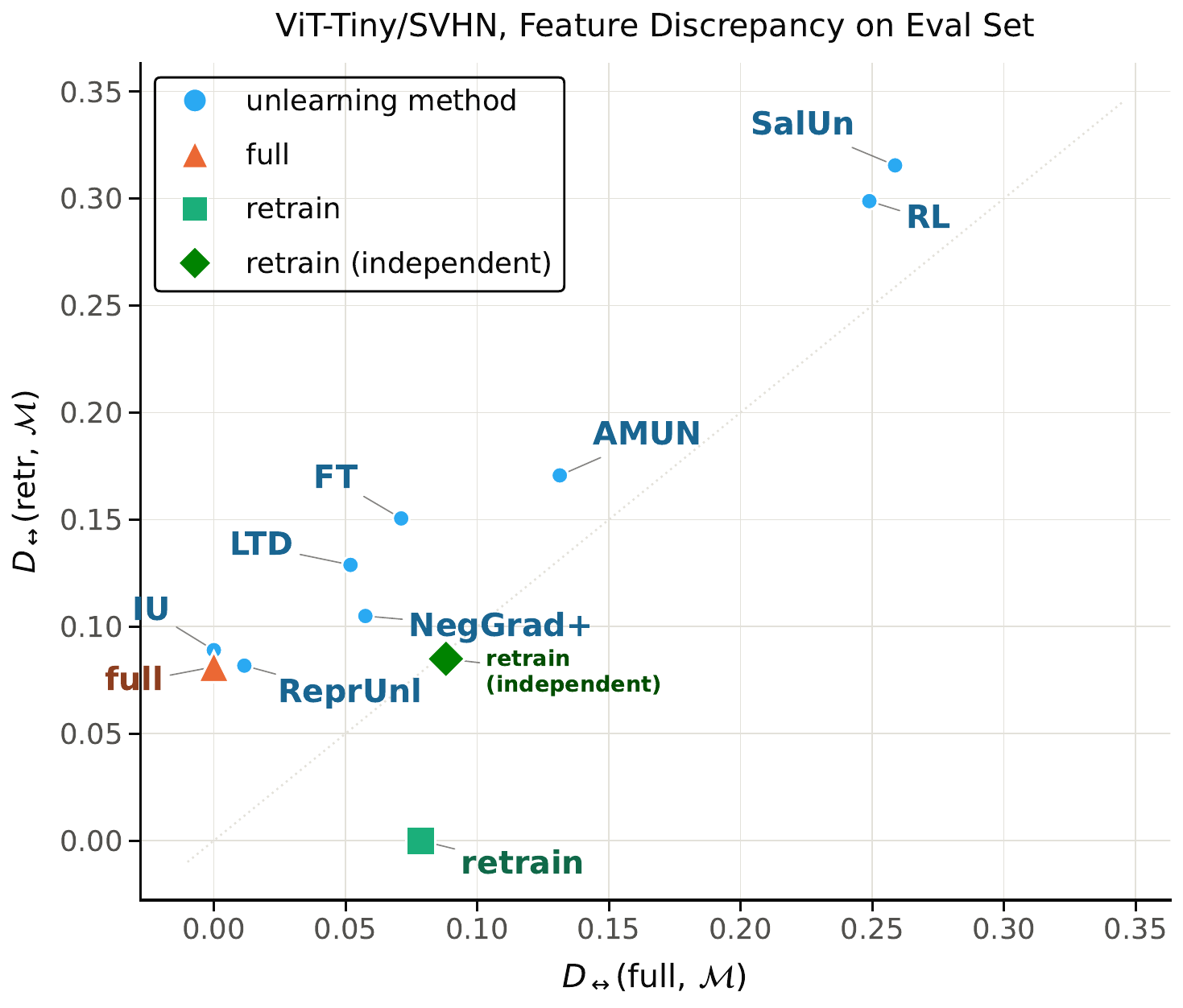}
        \caption{Test Set}
    \end{subfigure}
    \hfill
    \begin{subfigure}{0.45\linewidth}
        \centering
        \includegraphics[width=\linewidth]{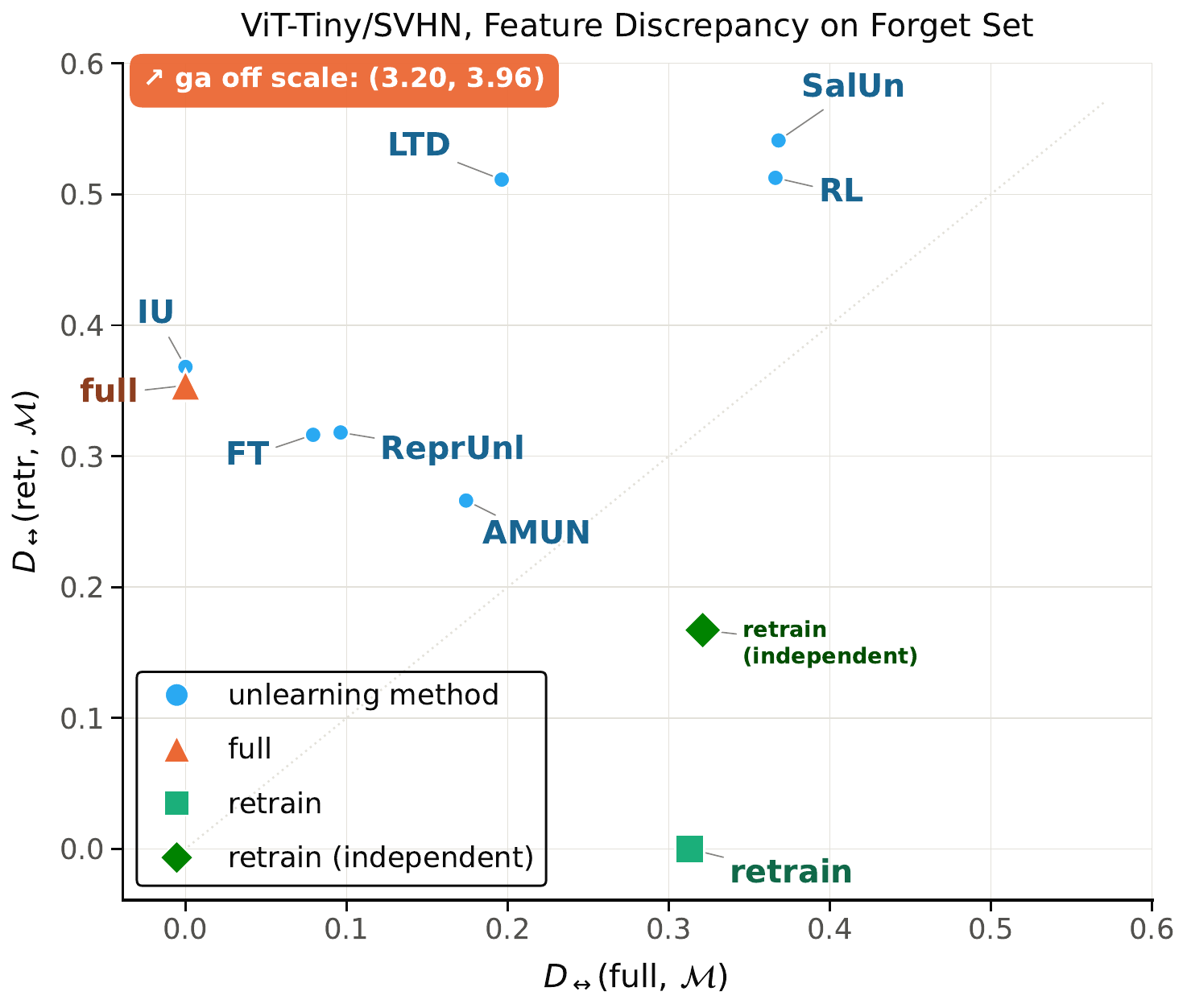}
        \caption{Forget Set}
    \end{subfigure}
    \caption{Feature discrepancy for ViT-Tiny on SVHN class-wise deletion. Each panel compares unlearned models with the full-data and retrained feature solutions on (a) the test distribution and (b) the forget distribution.}
    \label{fig:feature-recovery-svhn}
\end{figure}

\begin{figure}[t]
    \centering
    \begin{subfigure}{0.45\linewidth}
        \centering
        \includegraphics[width=\linewidth]{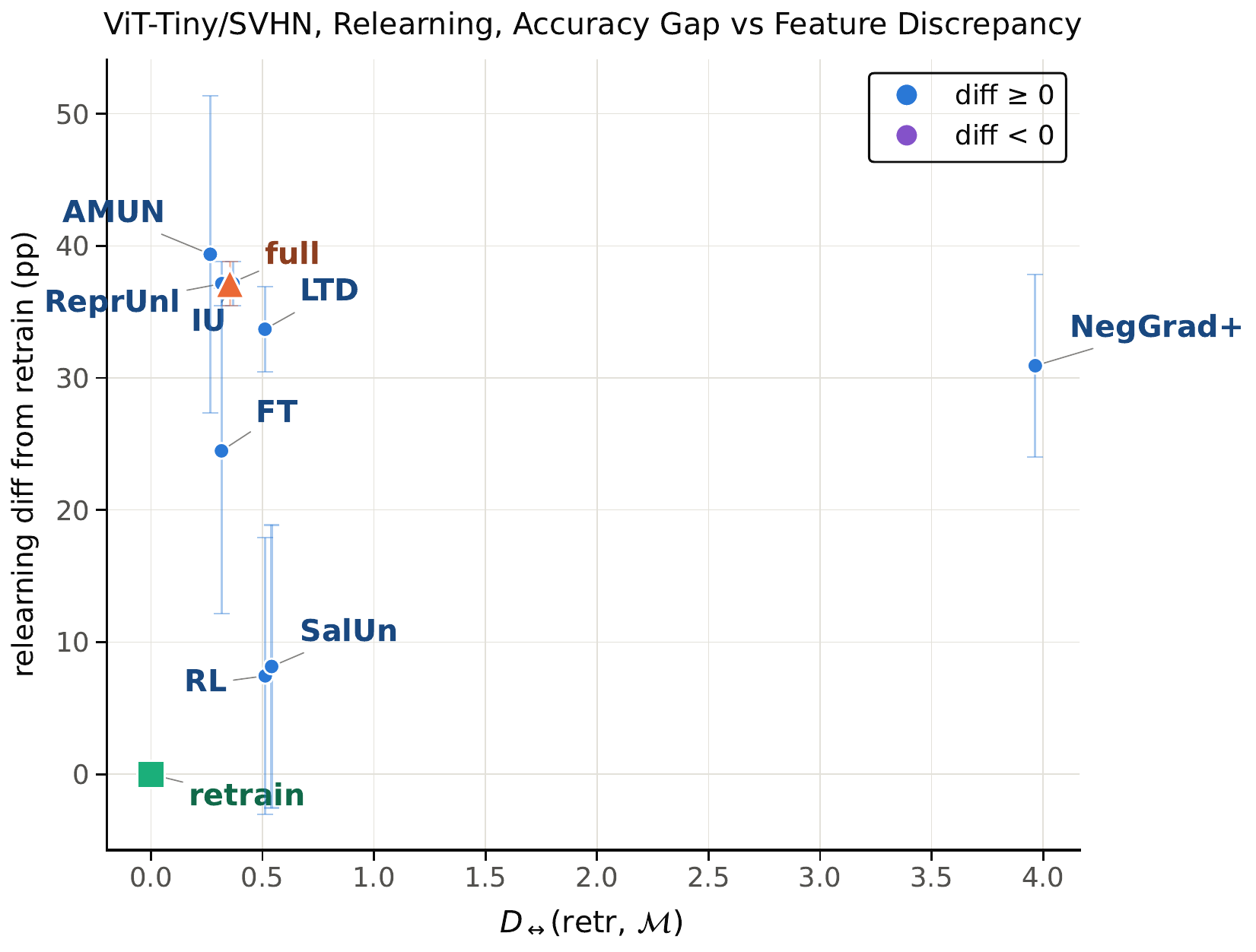}
        \caption{Forget Set}
    \end{subfigure}
    \hfill
    \begin{subfigure}{0.45\linewidth}
        \centering
        \includegraphics[width=\linewidth]{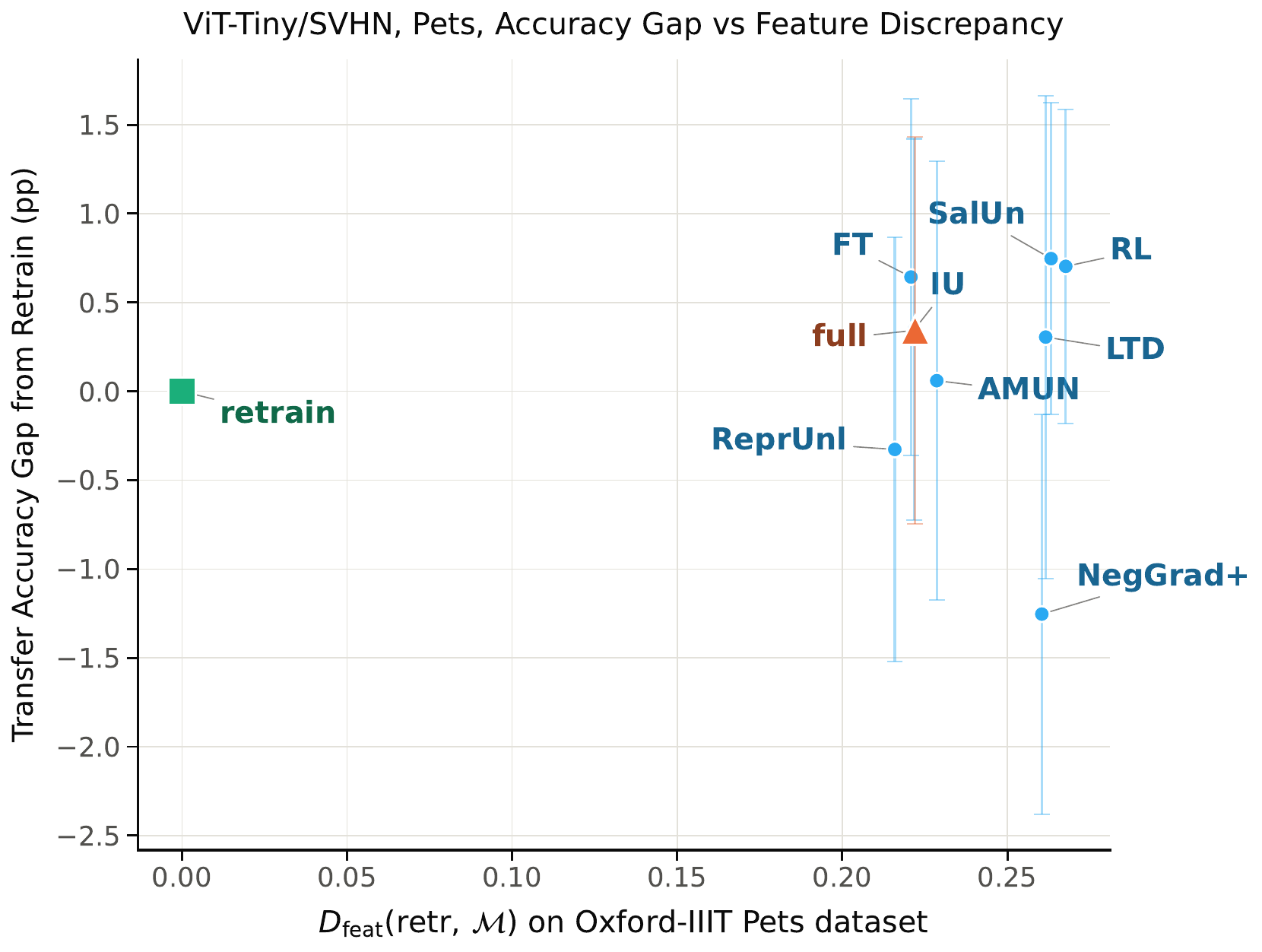}
        \caption{Oxford-IIIT Pets}
    \end{subfigure}
    \caption{Feature mismatch and behavior under subsequent adaptation for ViT-Tiny/SVHN.}
    \label{fig:feature-vs-adaptation-svhn}
\end{figure}

\paragraph{ViT-Tiny on SVHN: single-class deletion.} Finally, we consider ViT-Tiny on SVHN with one class removed. On both the test and forget distributions, most unlearned models remain concentrated near the full-data feature solution, providing an additional example of feature persistence (Figure~\ref{fig:feature-recovery-svhn}). Because the separation between methods is comparatively small in this setting, we avoid drawing fine-grained conclusions about their relative ordering. Figure~\ref{fig:feature-vs-adaptation-svhn} reports the corresponding head fine-tuning result.

Overall, these results show that the main feature-recovery picture is not specific to the ViT-Tiny/CIFAR-100 setting. Across architectures and deletion mechanisms, standard behavioral success can coexist with substantial mismatch from retraining, while some settings---such as SVHN---are dominated instead by persistence of the full-data feature solution.

\subsection{Hyperparameter ablations}\label{app:param-ablations}
\paragraph{Matching threshold $\gamma$.} We set the matching threshold as 
\[
    \gamma = \kappa\cdot\gamma_{\mathrm{corr}},
\] 
where $\gamma_{\mathrm{corr}}$ is the correlation threshold used to construct recurring feature clusters and $\kappa$ controls the sensitivity of the discrepancy. At $\kappa=1$, many moderately similar components already count as full matches, compressing the range of discrepancies. Increasing $\kappa$ makes the metric more sensitive to imperfect feature recovery.

\begin{figure*}[h]
    \centering

    % Row 1
    \begin{subfigure}{0.32\textwidth}
        \centering
        \includegraphics[width=\linewidth]{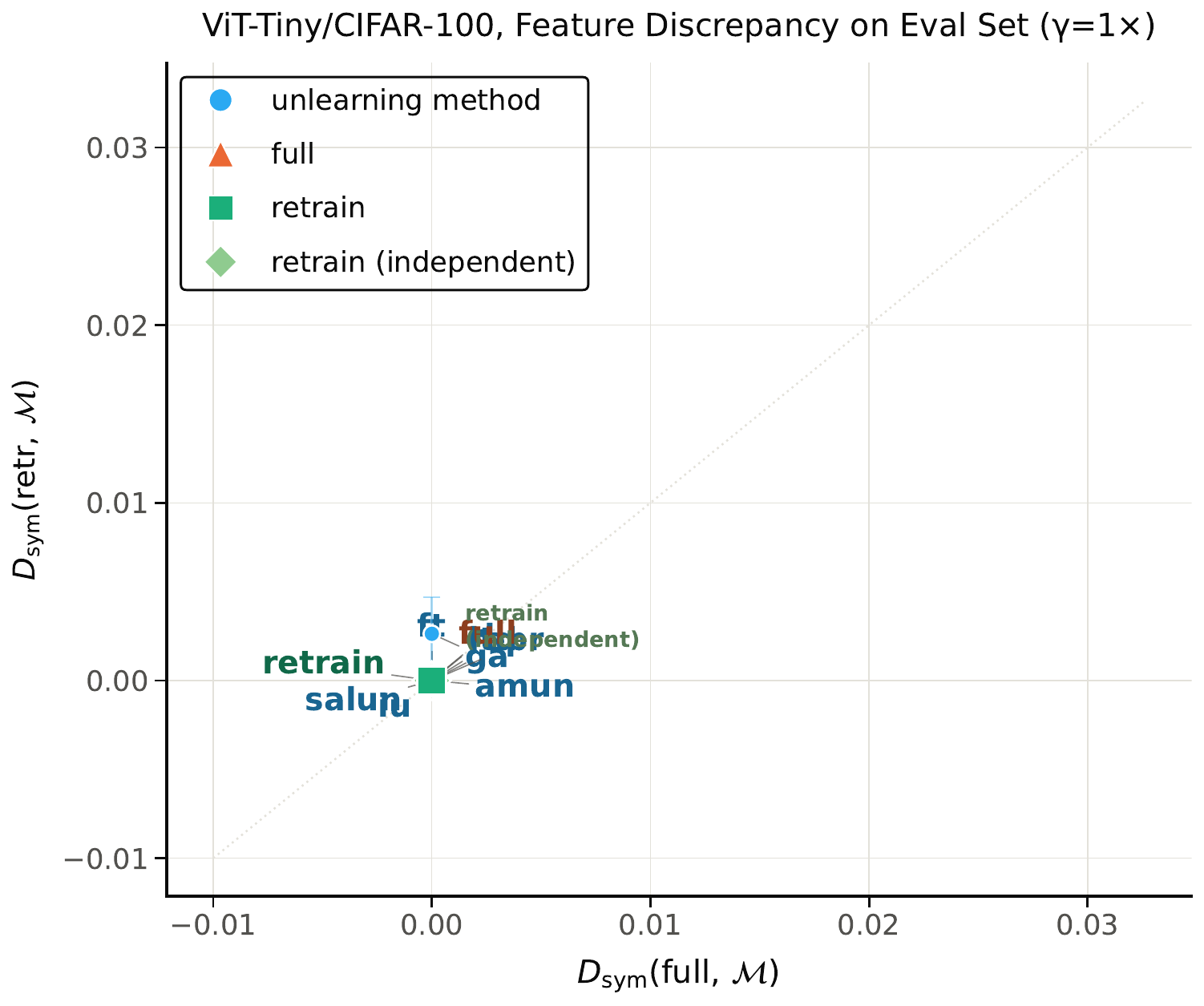}
        \caption{$\kappa=1$}
    \end{subfigure}
    \hfill
    \begin{subfigure}{0.32\textwidth}
        \centering
        \includegraphics[width=\linewidth]{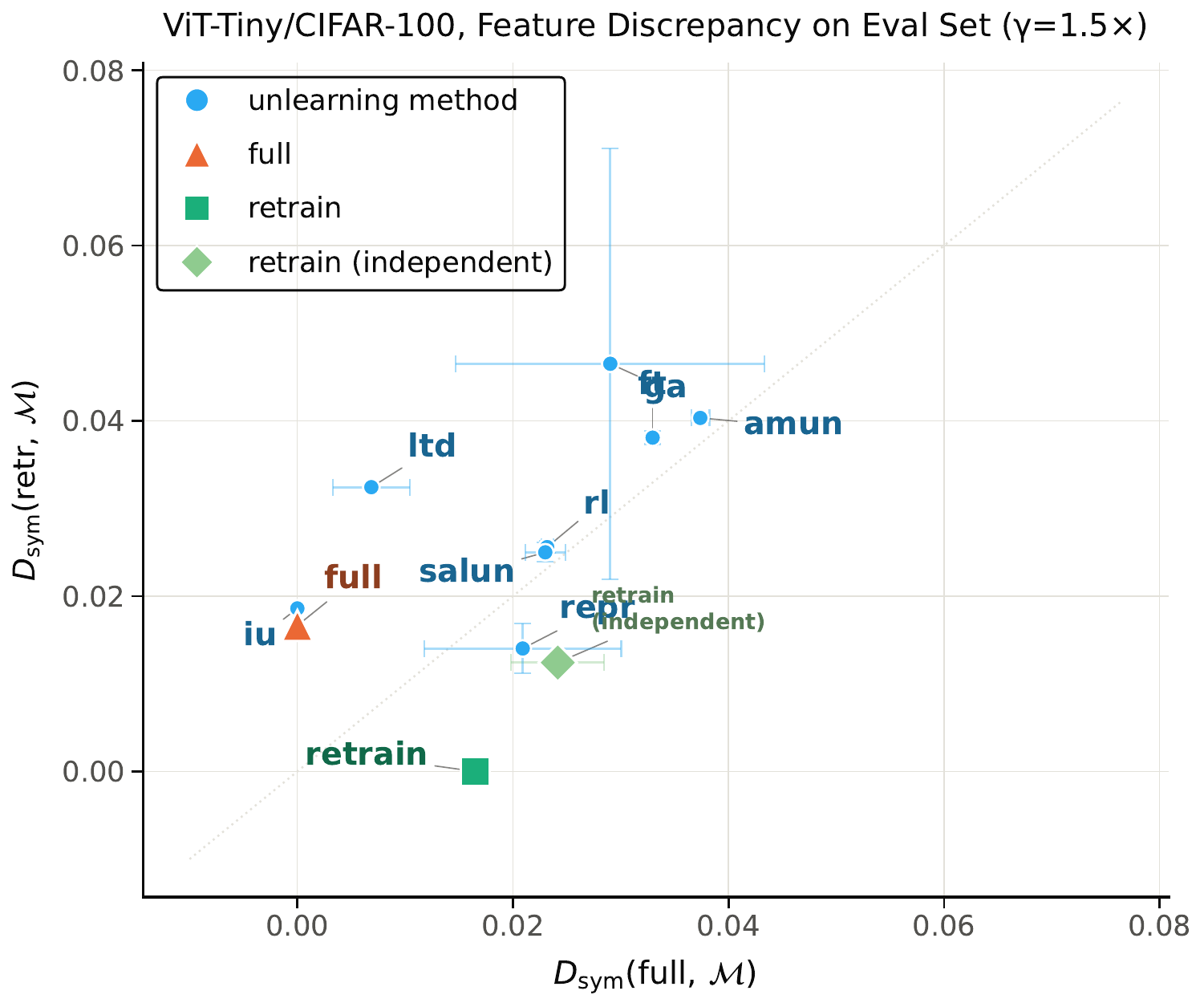}
        \caption{$\kappa=1.5$}
    \end{subfigure}
    \hfill
    \begin{subfigure}{0.32\textwidth}
        \centering
        \includegraphics[width=\linewidth]{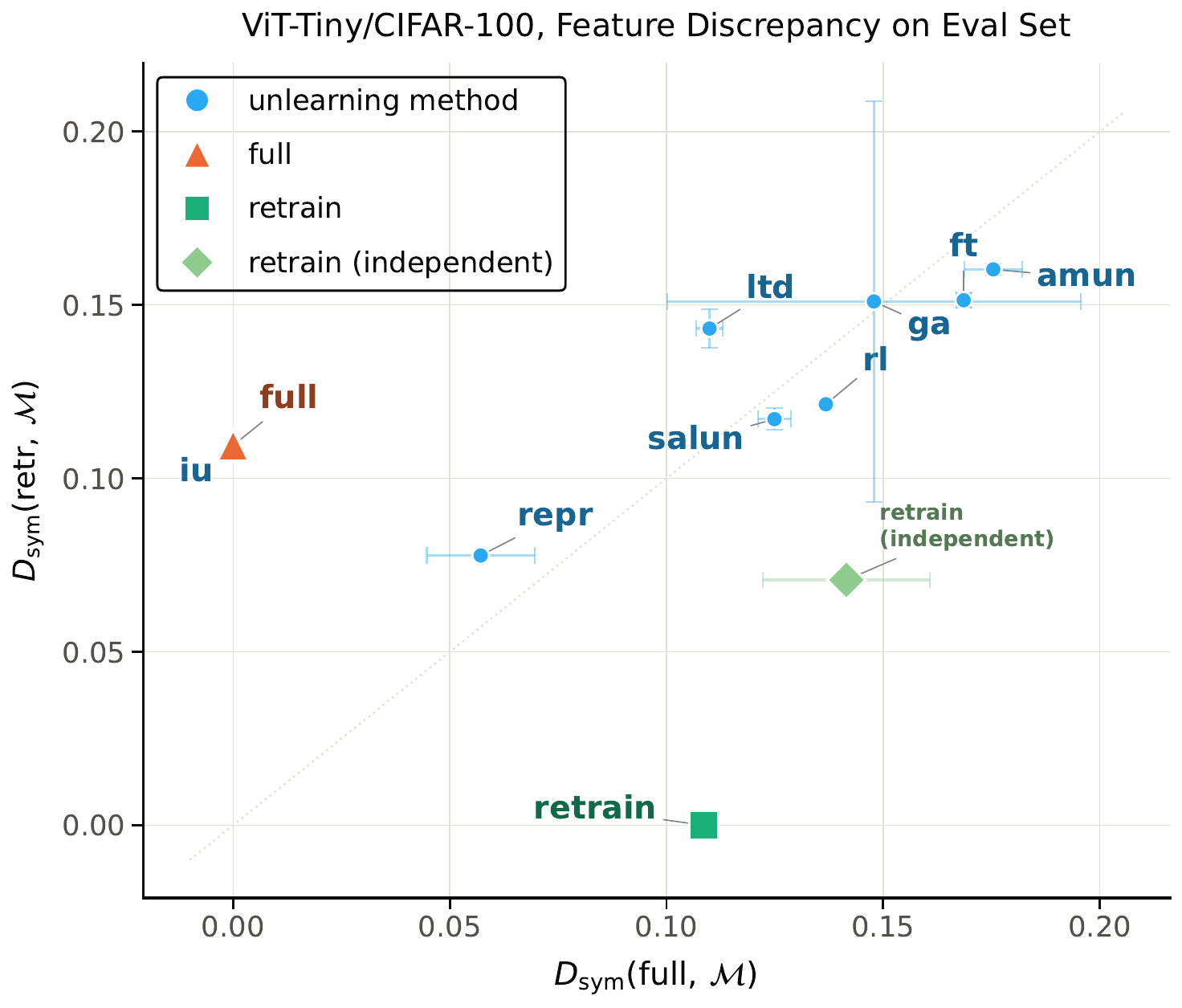}
        \caption{$\kappa=2$ (default)}
    \end{subfigure}

    \vspace{0.8em}

    % Row 2
    \begin{subfigure}{0.32\textwidth}
        \centering
        \includegraphics[width=\linewidth]{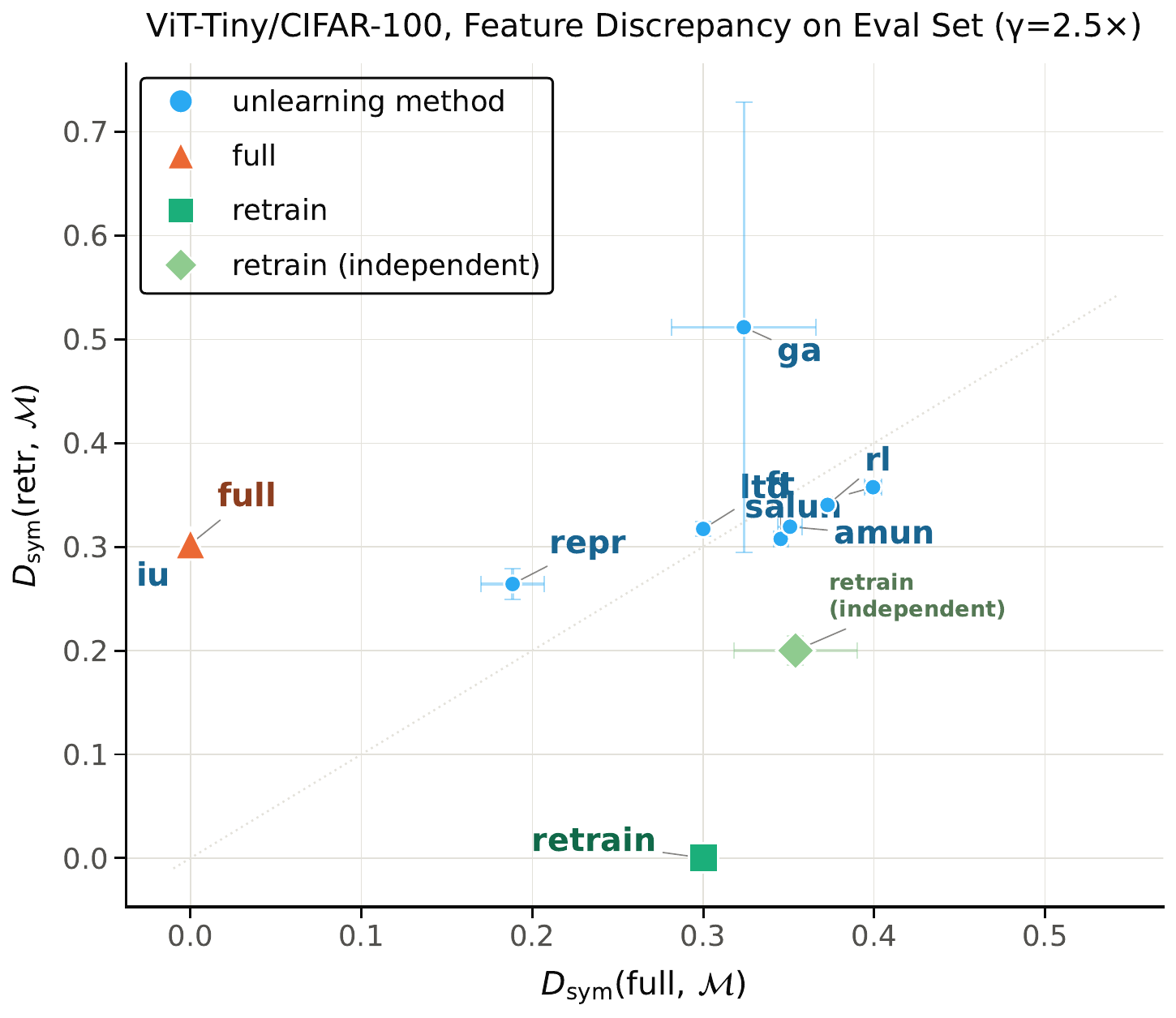}
        \caption{$\kappa=2.5$}
    \end{subfigure}
    \hspace{0.04\textwidth}
    \begin{subfigure}{0.32\textwidth}
        \centering
        \includegraphics[width=\linewidth]{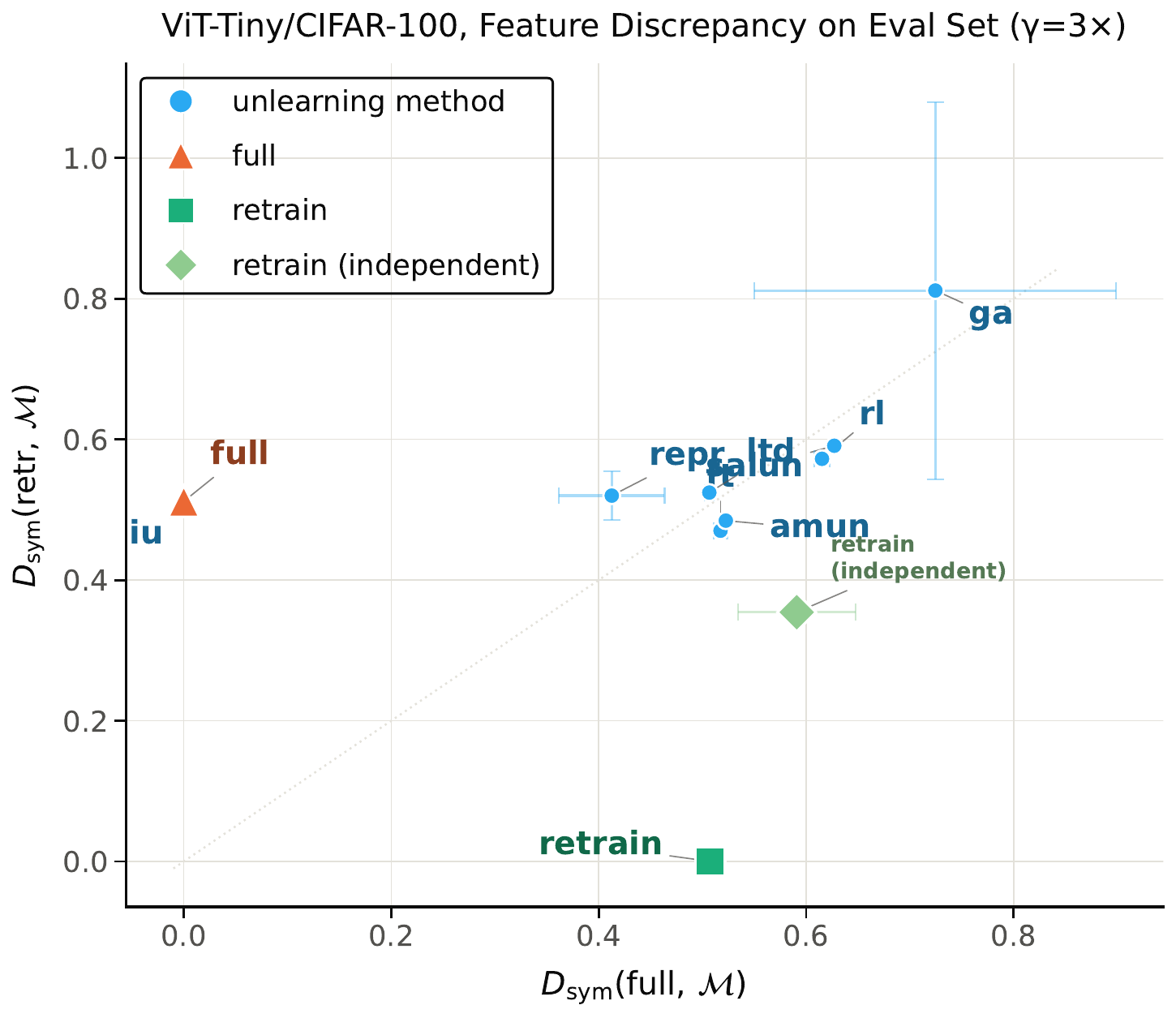}
        \caption{$\kappa=3$}
    \end{subfigure}

    \caption{Sensitivity of the feature discrepancy to the matching threshold
    $\gamma=\kappa\,\gamma_{\mathrm{corr}}$. Larger $\kappa$ increases sensitivity to imperfect feature matches.}
    \label{fig:gamma-ablation}
\end{figure*}

\begin{figure*}[h!]
    \centering

    % Row 1
    \begin{subfigure}{0.32\textwidth}
        \centering
        \includegraphics[width=\linewidth]{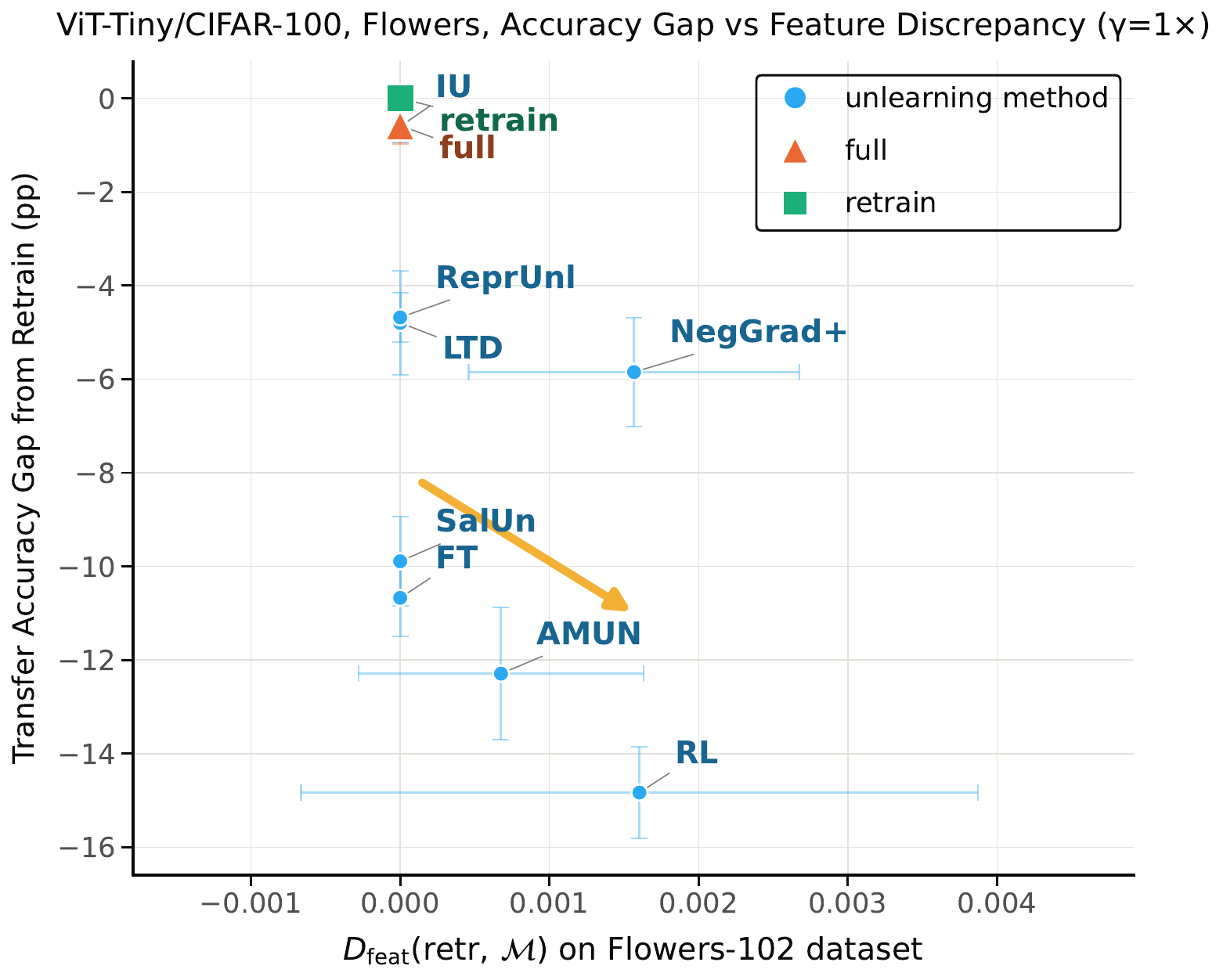}
        \caption{$\kappa=1$}
    \end{subfigure}
    \hfill
    \begin{subfigure}{0.32\textwidth}
        \centering
        \includegraphics[width=\linewidth]{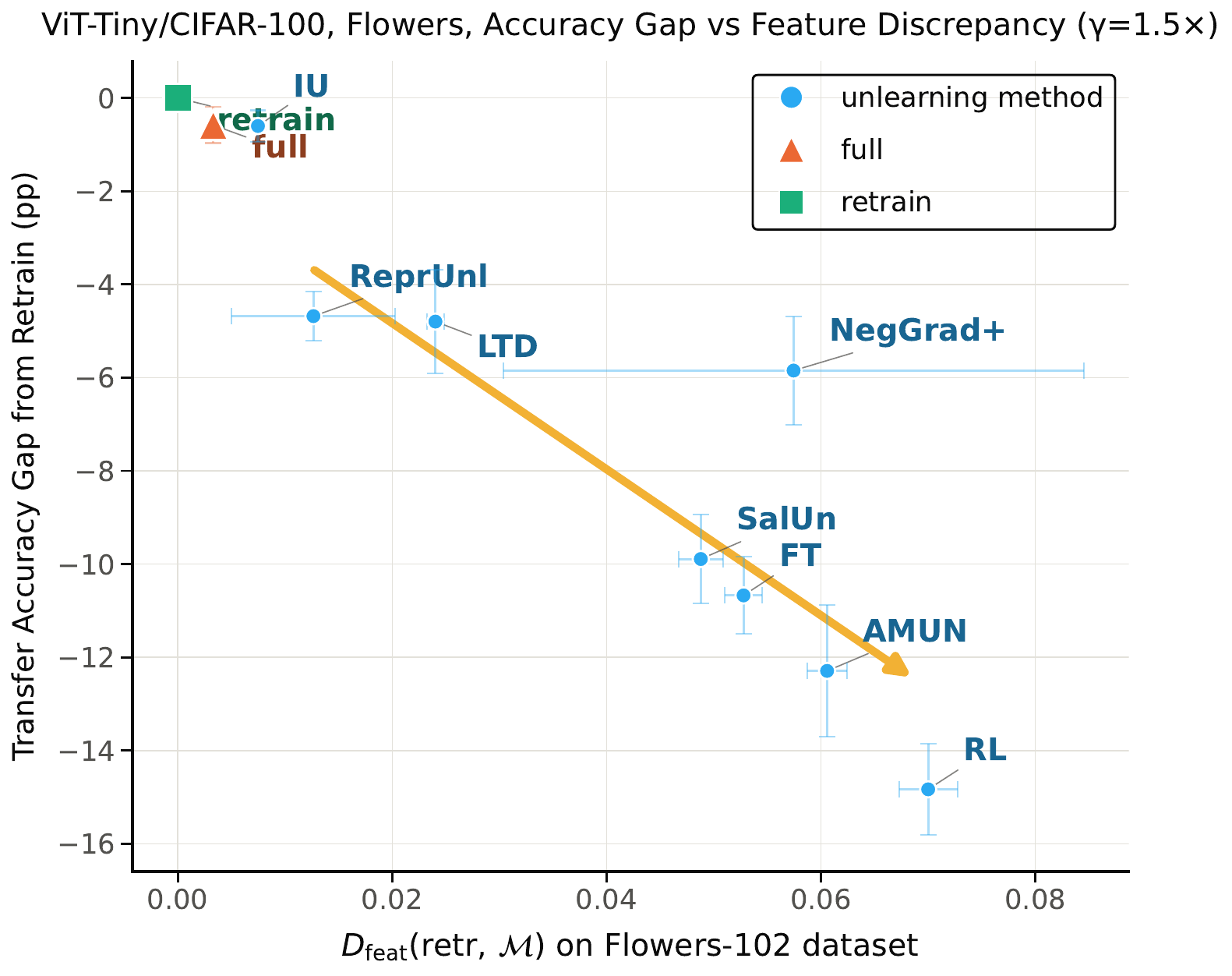}
        \caption{$\kappa=1.5$}
    \end{subfigure}
    \hfill
    \begin{subfigure}{0.32\textwidth}
        \centering
        \includegraphics[width=\linewidth]{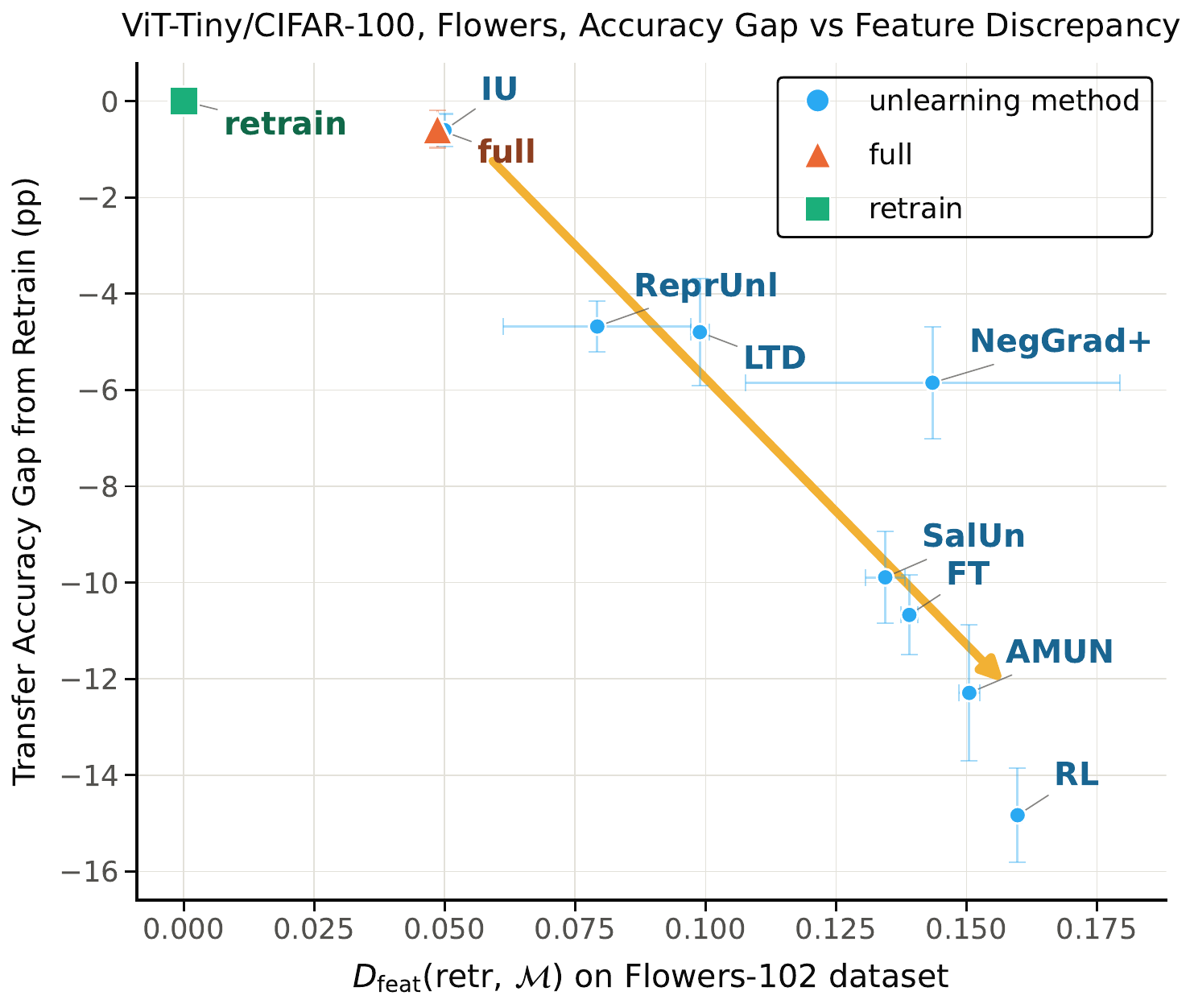}
        \caption{$\kappa=2$ (default)}
    \end{subfigure}

    \vspace{0.8em}

    % Row 2
    \begin{subfigure}{0.32\textwidth}
        \centering
        \includegraphics[width=\linewidth]{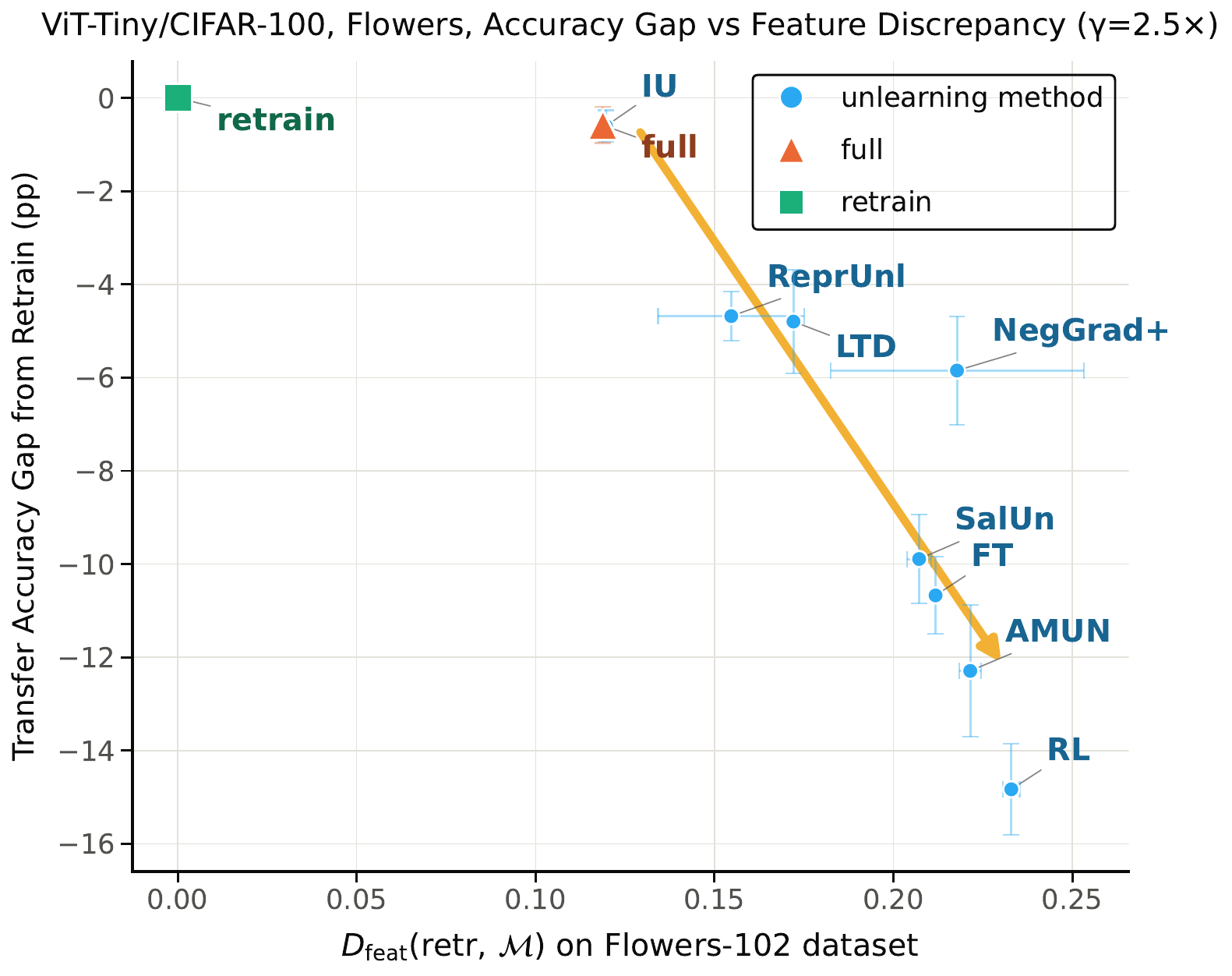}
        \caption{$\kappa=2.5$}
    \end{subfigure}
    \hspace{0.04\textwidth}
    \begin{subfigure}{0.32\textwidth}
        \centering
        \includegraphics[width=\linewidth]{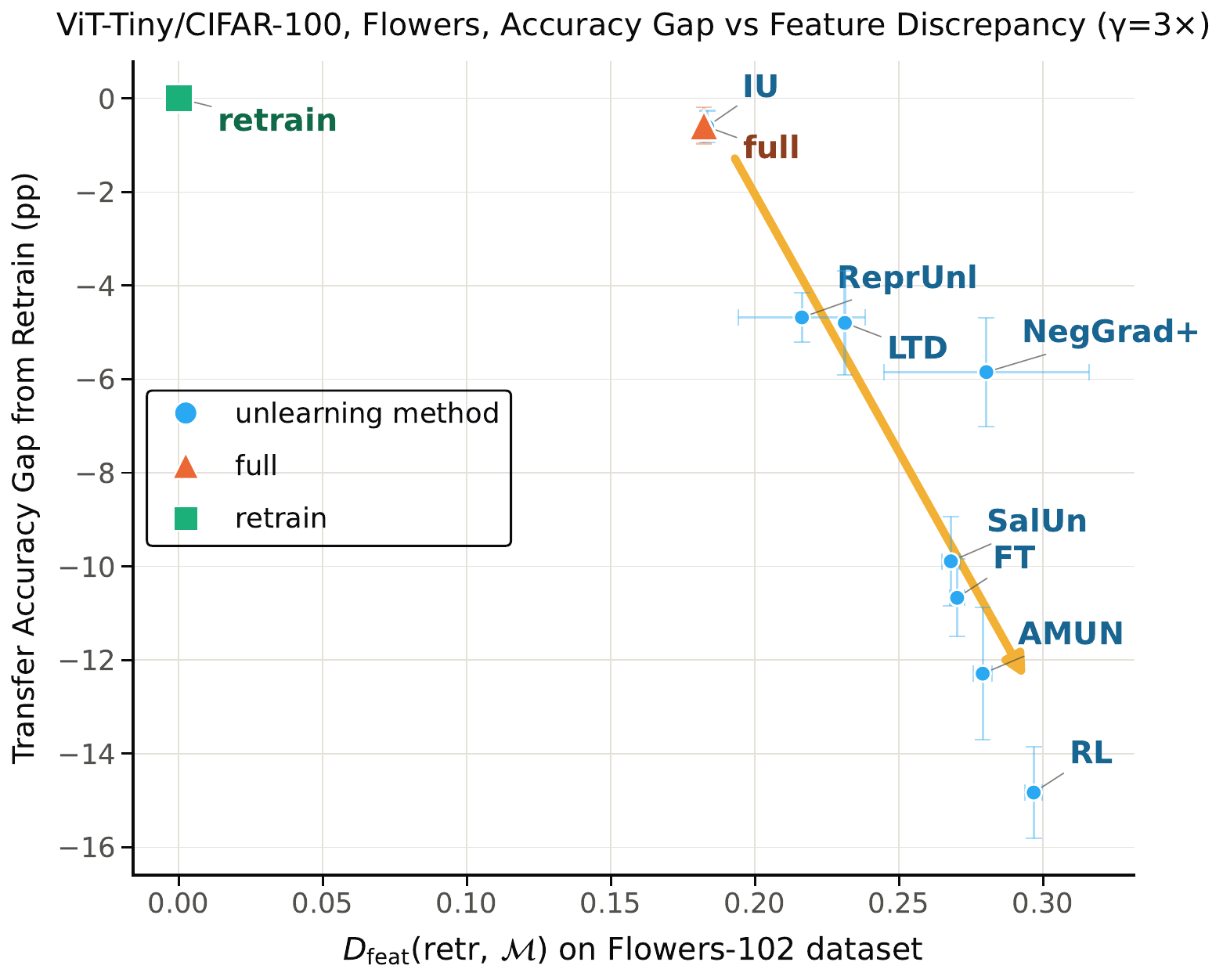}
        \caption{$\kappa=3$}
    \end{subfigure}

    \caption{Sensitivity of the relationship between feature discrepancy and downstream accuracy on Flowers-102 to the matching threshold $\gamma=\kappa\gamma_{\mathrm{corr}}$. The low-sensitivity setting $\kappa=1$ compresses the discrepancies, while the qualitative trend is preserved across larger values of $\kappa$.}
    \label{fig:gamma-flowers-ablation}
\end{figure*}

We calibrate $\gamma$ using independent retraining runs rather than unlearning results: two independently trained retraining groups should have a small discrepancy, since they represent samples from the same target solution. Figure~\ref{fig:gamma-ablation} reports this sensitivity on the evaluation set. We additionally repeat the downstream comparison from the main text on Flowers-102 across different values of \(\kappa\) (Figure~\ref{fig:gamma-flowers-ablation}). The same qualitative relationship between feature discrepancy and downstream accuracy is preserved across matching thresholds. The calibration also depends on group size because matching takes the best component across all models in the evaluated group; with more independent models, retraining groups are more likely to contain close matches and therefore become closer under the metric.

\begin{figure*}[t]
    \centering

    \begin{subfigure}{0.30\textwidth}
        \centering
        \includegraphics[width=\linewidth]{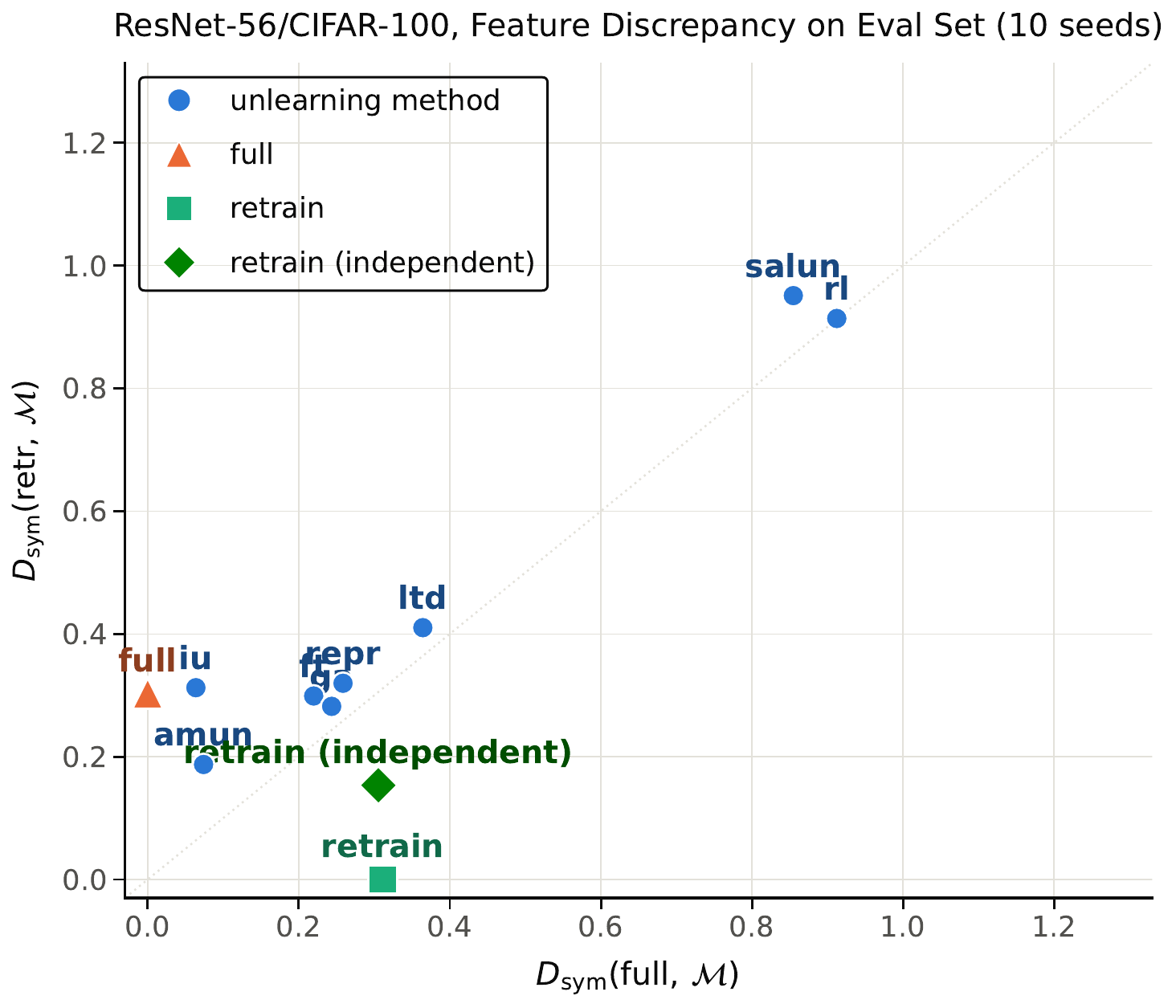}
        \caption{10 models}
    \end{subfigure}
    \hfill
    \begin{subfigure}{0.30\textwidth}
        \centering
        \includegraphics[width=\linewidth]{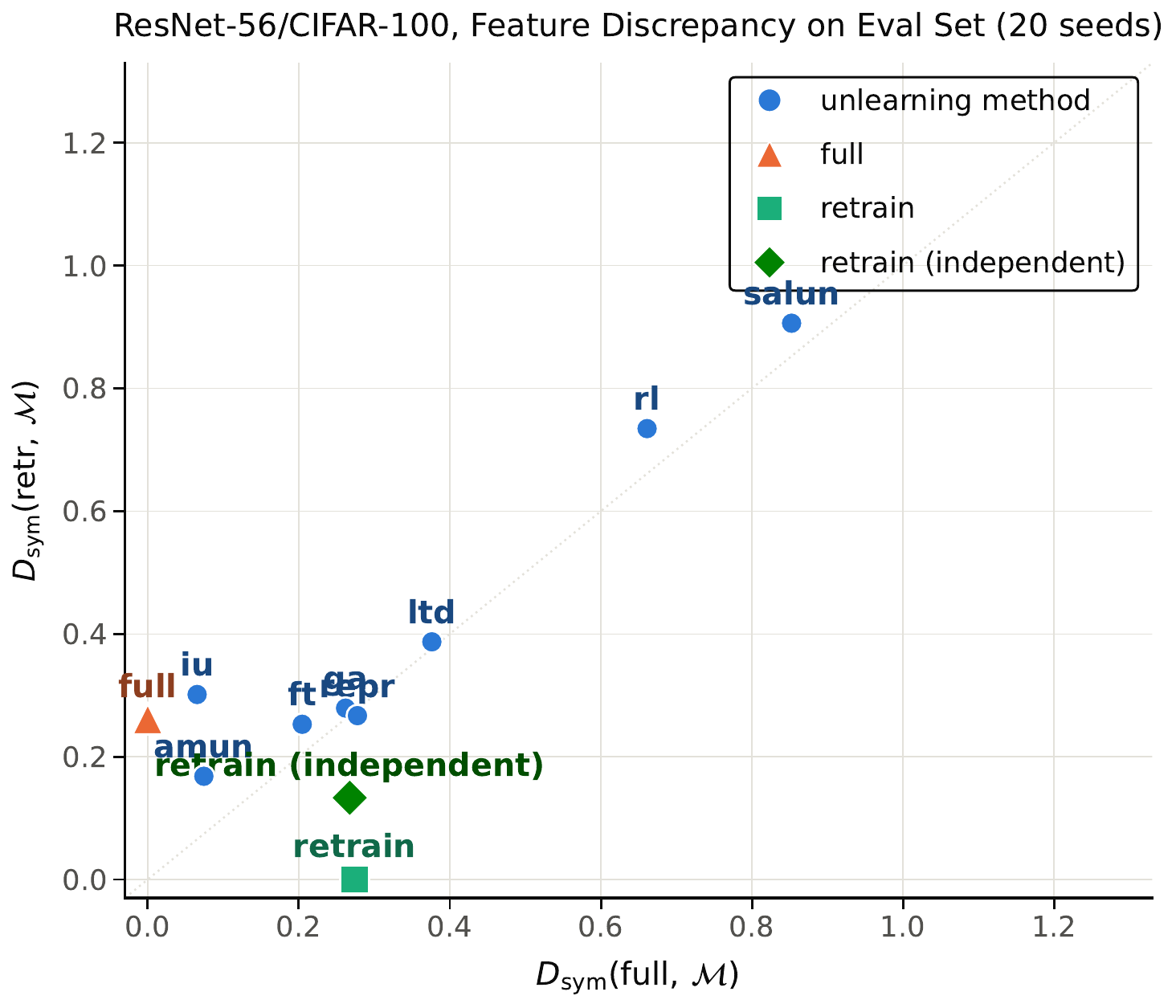}
        \caption{20 models}
    \end{subfigure}
    \hfill
    \begin{subfigure}{0.30\textwidth}
        \centering
        \includegraphics[width=\linewidth]{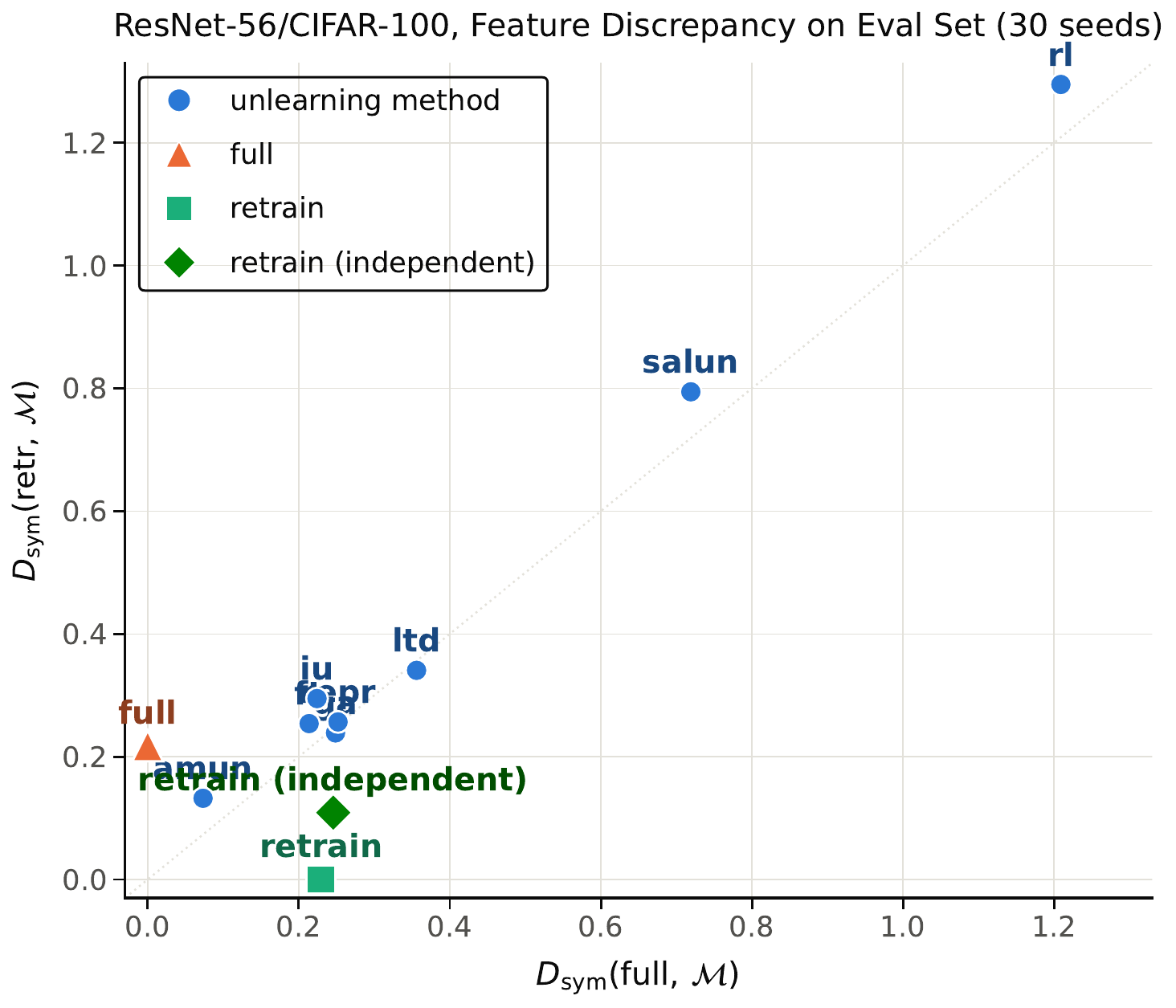}
        \caption{30 models}
    \end{subfigure}

    \caption{Sensitivity to the number of models used to construct each feature solution for ResNet-56 on CIFAR-100 under class-wise deletion. Larger groups reduce the discrepancy between independent retraining groups, while the broad relative structure of the unlearning methods is preserved despite some method-level variation.}
    \label{fig:num-models-ablation}
\end{figure*}

\paragraph{Number of models.}
Because feature solutions are constructed from recurring components across independent runs, the comparison can depend on the number of models in each group. Figure~\ref{fig:num-models-ablation} compares groups of 10, 20, and 30 models for ResNet-56 under class-wise deletion. Larger groups bring an independent retraining group closer to the reference retraining solution, consistent with improved coverage of recurring features. The broad method-level picture is preserved across group sizes, although individual discrepancies can vary, particularly for methods whose features recur less consistently across runs. We use 20 models as a compromise between feature coverage and computational cost.

\subsection{Feature-Strength Mismatch}
\label{app:feature-strength}
Our feature discrepancy primarily measures whether reference feature directions are recovered. Once a feature is well matched, differences in its strength receive little or no additional penalty. We therefore separately examine how the strengths of matched features vary across unlearning methods.

We take the ten strongest feature clusters of the full-data solution, ranked by their raw PCA eigenvalues. For each cluster and evaluated group $\mathcal M$, we identify its best-matching component using the same correlation-based matching as in our main metric and record the raw eigenvalue $\hat\lambda_{\mathcal M,i}$ of that component. We summarize the strength mismatch by
\[
\operatorname{NMSE}(\mathcal M)
=
\frac{\sum_i (\hat\lambda_{\mathcal M,i}-\lambda_{\mathrm{full},i})^2}
{\sum_i \lambda_{\mathrm{full},i}^2}.
\]

\begin{figure*}[t]
    \centering

    \begin{subfigure}{0.48\textwidth}
        \centering
        \includegraphics[
            width=\linewidth,
            trim=0 0 0 12mm,
            clip
        ]{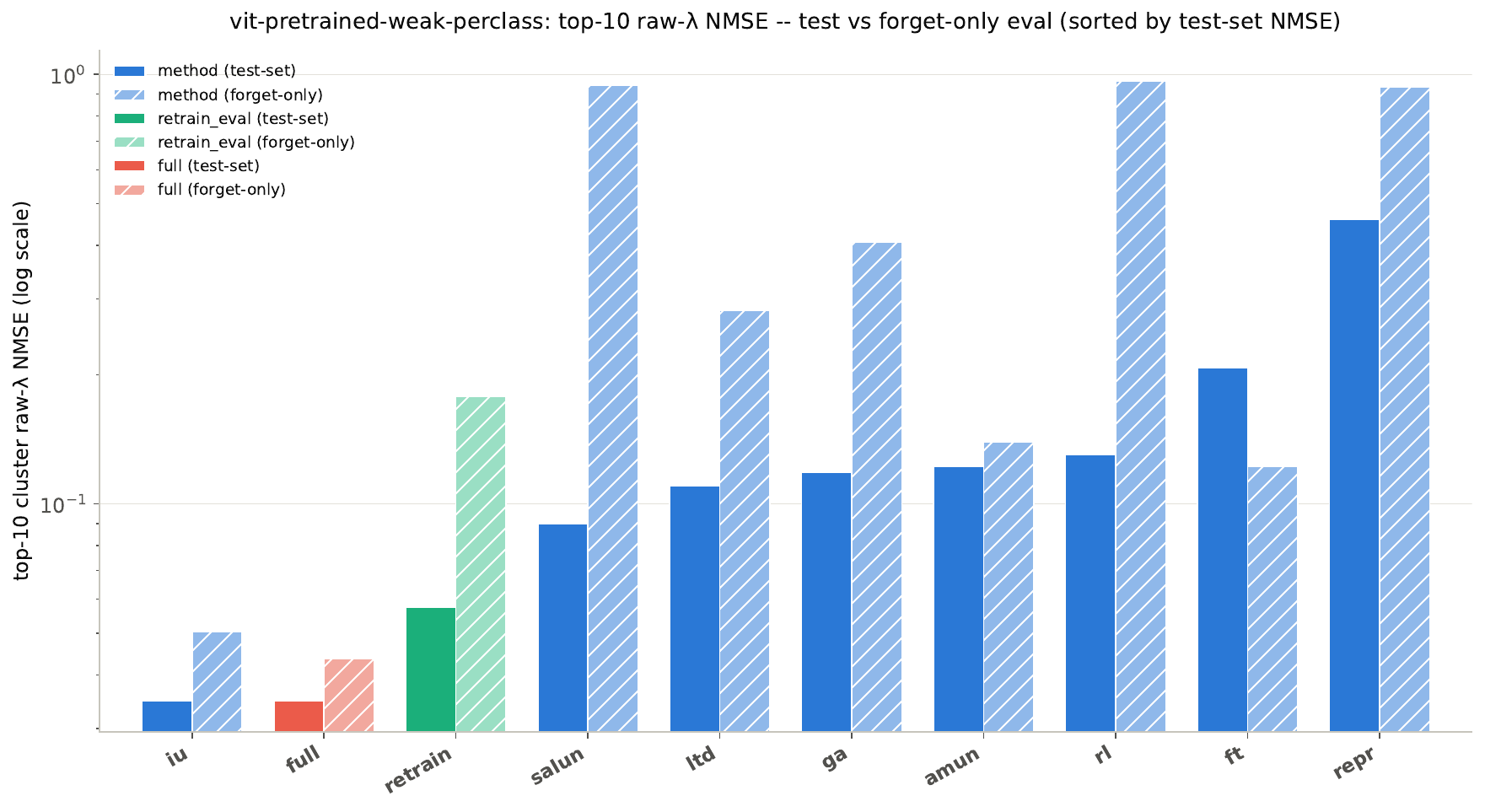}
        \caption{ViT-Tiny}
    \end{subfigure}
    \hfill
    \begin{subfigure}{0.48\textwidth}
        \centering
        \includegraphics[
            width=\linewidth,
            trim=0 0 0 12mm,
            clip
        ]{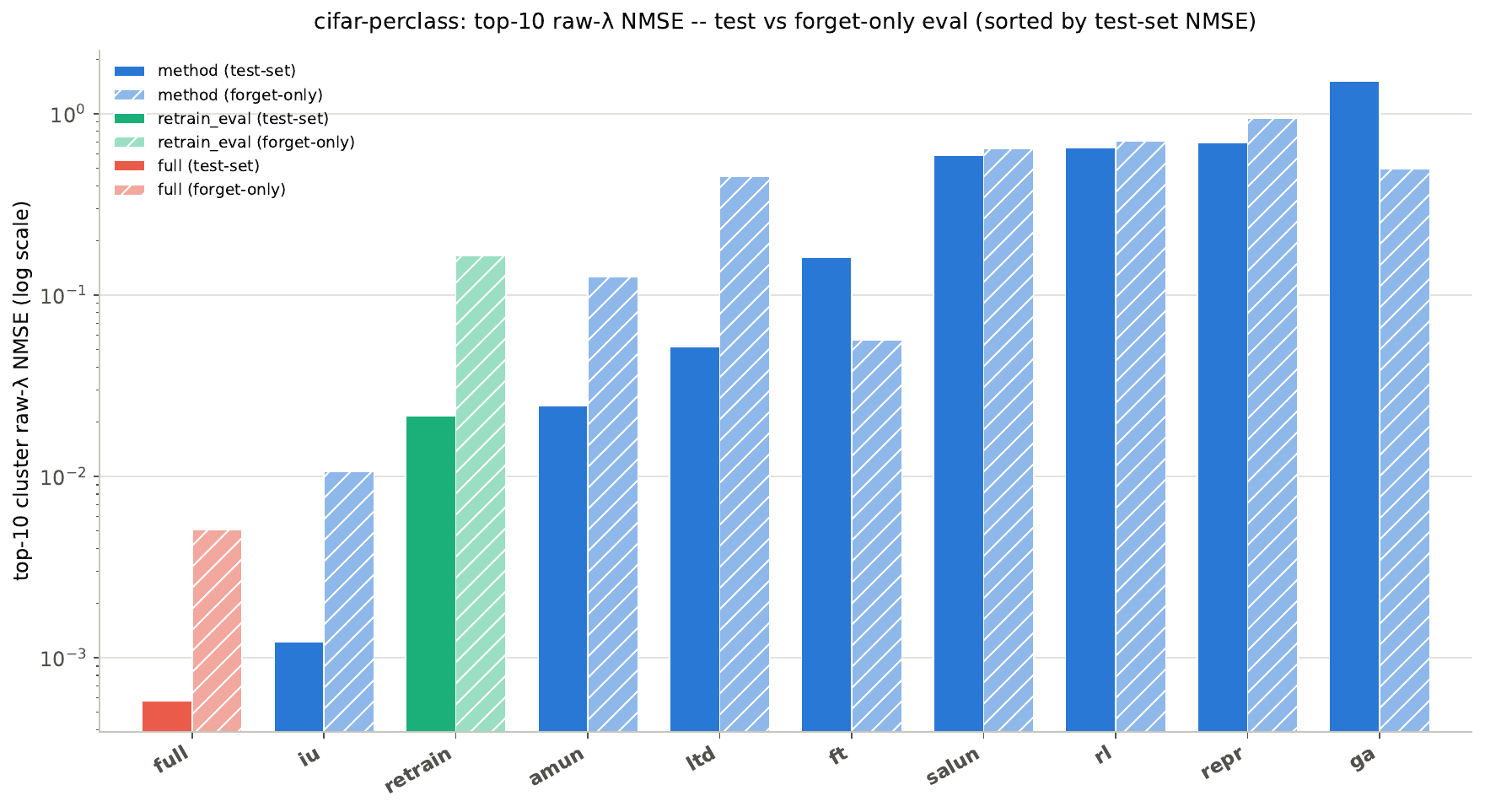}
        \caption{ResNet-56}
    \end{subfigure}

    \caption{Feature-strength mismatch for the ten strongest full-data feature clusters. For each reference cluster, we match the closest component in each evaluated group and compare its raw PCA eigenvalue with the full-data reference. Bars show normalized mean-squared error on the test and forget distributions; lower values indicate better agreement in feature strength.}
    \label{fig:feature-strength}
\end{figure*}

Figure~\ref{fig:feature-strength} shows that matching a feature direction does not imply matching its strength. Several unlearning methods exhibit substantial eigenvalue mismatch, which can become considerably larger on the forget distribution than on the standard test distribution. The effect is method- and setting-dependent, but it demonstrates that feature identity alone does not fully characterize the representation: even recovered directions can appear with substantially different strengths.

\section{Relation to Linear CKA}
\label{app:cka}
Our feature comparison is closely related to linear CKA. For centered representations with Gram decompositions
\[
K_A=\sum_i \lambda_i^A u_i^A (u_i^A)^\top,
\qquad
K_B=\sum_j \lambda_j^B u_j^B (u_j^B)^\top,
\]
linear CKA can be written as
\[
\operatorname{CKA}(A,B)
=
\frac{\sum_{i,j}\lambda_i^A\lambda_j^B\bigl((u_i^A)^\top u_j^B\bigr)^2}
{\sqrt{\sum_i(\lambda_i^A)^2}\sqrt{\sum_j(\lambda_j^B)^2}}.
\]
Because the response directions are centered and normalized, $(u_i^A)^\top u_j^B$ is their correlation. Thus, CKA aggregates the same basic ingredients underlying our comparison---feature alignment and strength---but compresses them into a single symmetric similarity score.

Our metric instead retains directional and group-level structure. $D_{\mathrm{feat}}(\mathcal R,\mathcal M)$ asks which recurring features of a reference solution $\mathcal R$ are poorly represented by $\mathcal M$. Directionality is particularly important for unlearning: comparing full data and retraining in both directions separates features that disappear after deletion from features that emerge only under retraining. Our analysis shows that this distinction matters, since successful unlearning may require not only removing features of the full-data solution but also recovering features newly selected by retraining. A symmetric aggregate such as CKA cannot distinguish these two changes.

\begin{figure*}[t]
    \centering
    \begin{subfigure}{0.48\textwidth}
        \centering
        \includegraphics[width=\linewidth]{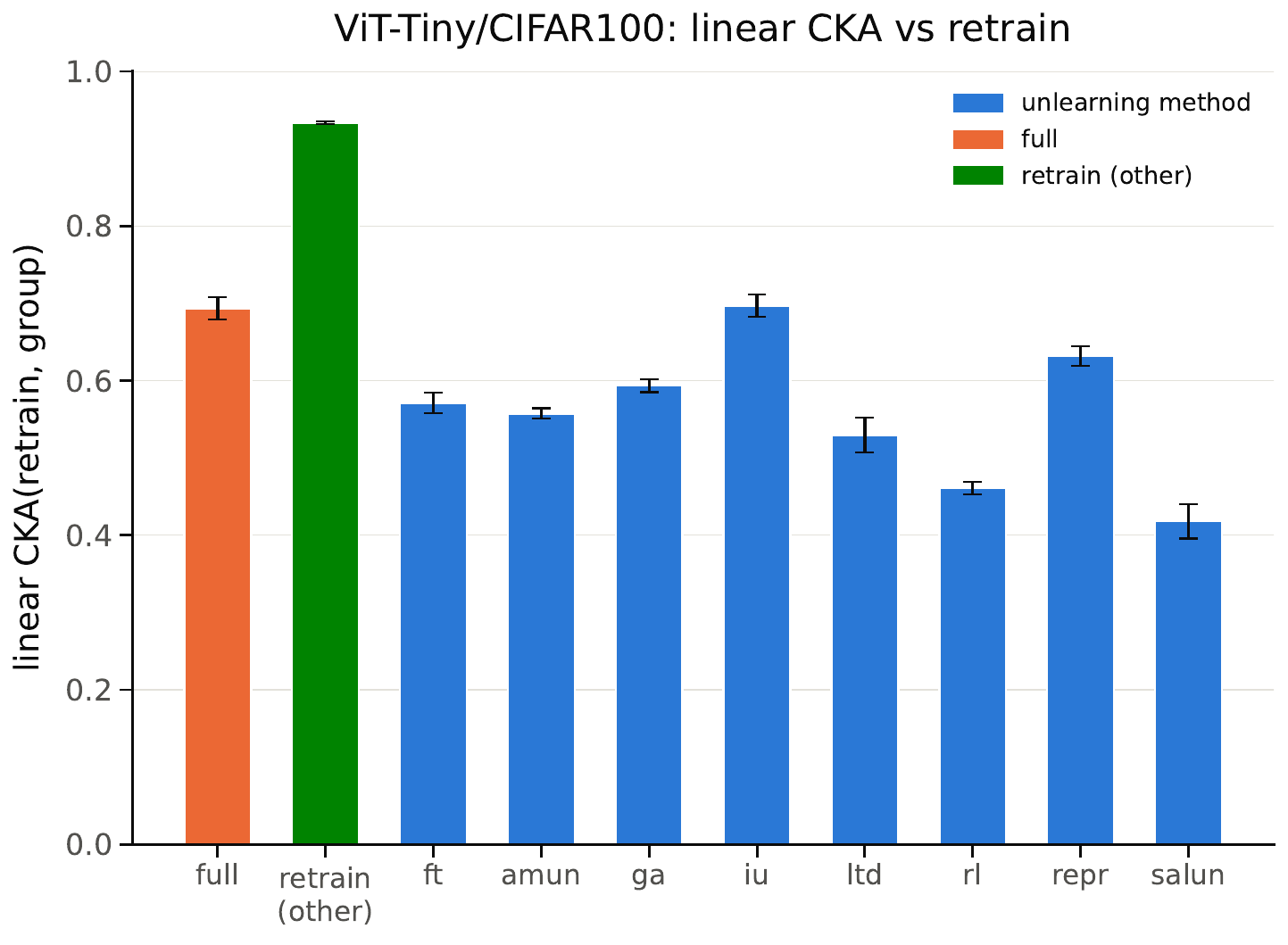}
        \caption{ViT-Tiny}
    \end{subfigure}
    \hfill
    \begin{subfigure}{0.48\textwidth}
        \centering
        \includegraphics[width=\linewidth]{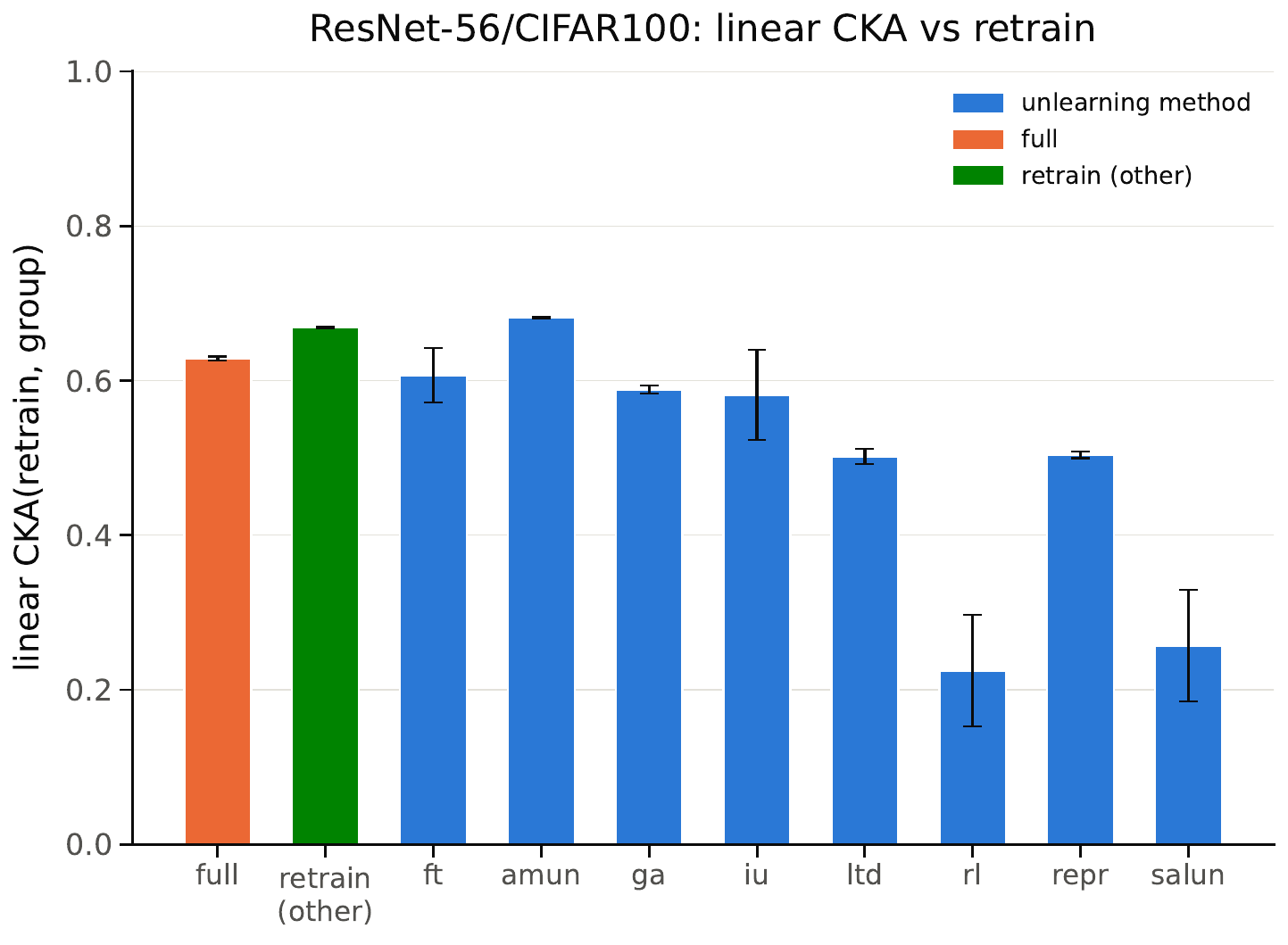}
        \caption{ResNet-56}
    \end{subfigure}
    \caption{Linear CKA with a retrained reference model. Independent retraining runs are highly similar for ViT-Tiny, whereas pairwise retraining similarity is substantially lower for ResNet-56; in the latter case, AMUN attains CKA comparable to or higher than an independent retraining run. This illustrates why pairwise representation similarity can be misleading when retraining varies across runs.}
    \label{fig:cka-comparison}
\end{figure*}

The group-level construction is also important because individual retrained models need not provide a stable reference. Figure~\ref{fig:cka-comparison} illustrates this effect. For ViT-Tiny, independently retrained models have substantially higher CKA with the retraining reference than any unlearning method. For ResNet-56, however, pairwise CKA between independent retraining runs is itself much lower, and AMUN attains comparable or slightly higher similarity to the chosen retraining reference. A pairwise similarity score can therefore give the appearance of matching retraining even when the retraining solution varies substantially across runs. Our comparison instead identifies features that recur across the retraining ensemble and separately measures which of them are recovered by another group.

\section{Feature Injection and Ablation}
\subsection{Experiment Details}\label{app:feature-injection}
We provide additional details for the feature interventions used in Section~\ref{sec:consequences}. The goal is to directly correct the retraining-relative feature mismatch identified by our comparison and test whether subsequent behavior moves toward retraining. We use two complementary operations: \emph{ablation} removes feature responses present in the unlearned solution but poorly represented by retraining, while \emph{injection} adds responses corresponding to retraining features that are poorly represented by the unlearned solution. Applying the two operations together yields \emph{substitution}. All interventions modify the full penultimate-layer representation; PCA components are used only to identify the feature responses to be removed or added.

\paragraph{From feature clusters to seed-specific responses.}
Feature clusters are defined across model groups, whereas interventions are applied to individual models. We therefore pair each unlearned model with the retrained model of the same seed and identify, in each model, the top-$K$ PCA components belonging to the selected clusters. For a component $i$, let
\[
    r_i := \widetilde\Phi v_i = \sigma_i u_i
\]
denote its centered response across the intervention inputs. Since
\[
    \widetilde\Phi = \sum_i r_i v_i^\top,
\]
$r_i v_i^\top$ is the rank-one contribution of this component to the centered representation. Thus, intervention on a selected cluster acts through these seed-specific contributions: ablation removes those of the corresponding unlearned components, while injection uses the responses of the corresponding components from the paired retrained model. The top-$K$ PCA components are used only to identify the features to modify; the interventions themselves act on the full representation rather than on a truncated PCA reconstruction.

\paragraph{Feature ablation.}
For a selected unlearned component $i$, we remove its response by modifying the representation as
\[
    \Phi^{\mathrm{abl}}
    =
    \Phi^{\mathrm{unl}}
    -
    r_i^{\mathrm{unl}}
    \bigl(v_i^{\mathrm{unl}}\bigr)^\top.
\]
Since $r_i^{\mathrm{unl}}$ is centered, the representation mean is unchanged, and therefore
\[
    \widetilde\Phi^{\mathrm{abl}}
    =
    \widetilde\Phi^{\mathrm{unl}}
    -
    r_i^{\mathrm{unl}}
    \bigl(v_i^{\mathrm{unl}}\bigr)^\top.
\]
For a set $\mathcal A$ of selected components, orthonormality of the PCA directions then gives
\[
    \widetilde\Phi^{\mathrm{abl}}
    (\widetilde\Phi^{\mathrm{abl}})^\top
    =
    \widetilde\Phi^{\mathrm{unl}}
    (\widetilde\Phi^{\mathrm{unl}})^\top
    -
    \sum_{i\in\mathcal A}
    r_i^{\mathrm{unl}}
    (r_i^{\mathrm{unl}})^\top.
\]
Thus, ablation removes exactly the selected feature contributions.

\paragraph{Feature injection.}
For a selected component $j$ of the paired retrained model, we add its response as a new coordinate of the unlearned representation:
\[
    \Phi^{\mathrm{inj}}
    =
    \bigl[\Phi^{\mathrm{unl}},\, r_j^{\mathrm{retr}}\bigr].
\]
Since $r_j^{\mathrm{retr}}$ is centered,
\[
    \widetilde\Phi^{\mathrm{inj}}
    =
    \bigl[\widetilde\Phi^{\mathrm{unl}},\, r_j^{\mathrm{retr}}\bigr].
\]
Thus, for a set $\mathcal I$ of injected components,
\[
    \widetilde\Phi^{\mathrm{inj}}
    (\widetilde\Phi^{\mathrm{inj}})^\top
    =
    \widetilde\Phi^{\mathrm{unl}}
    (\widetilde\Phi^{\mathrm{unl}})^\top
    +
    \sum_{j\in\mathcal I}
    r_j^{\mathrm{retr}}
    (r_j^{\mathrm{retr}})^\top.
\]
Hence, injection adds exactly the selected retraining feature contributions without altering the existing unlearned ones.

\paragraph{Feature substitution.}
When ablation and injection are combined, we use an ablated PCA direction to carry an injected retraining response. After ablating component $i$, the centered representation has zero response along $v_i^{\mathrm{unl}}$. We can therefore inject a retraining response $r_j^{\mathrm{retr}}$ along this direction, replacing
\[
    r_i^{\mathrm{unl}}
    \bigl(v_i^{\mathrm{unl}}\bigr)^\top
    \quad\text{with}\quad
    r_j^{\mathrm{retr}}
    \bigl(v_i^{\mathrm{unl}}\bigr)^\top.
\]
By orthogonality of the PCA directions, this replaces the corresponding Gram contribution
$r_i^{\mathrm{unl}}(r_i^{\mathrm{unl}})^\top$
with
$r_j^{\mathrm{retr}}(r_j^{\mathrm{retr}})^\top$. If there are more injected than ablated components, the remaining responses are appended as new dimensions; excess ablations are simply removed.

\paragraph{Intervention levels.}
The operations above specify how we modify a selected feature response. To test whether behavior changes progressively as more of the measured retraining-relative mismatch is corrected, we apply them to nested sets of feature clusters.

The two directional discrepancies determine which clusters are eligible for intervention: retraining clusters contributing to
$D_{\mathrm{feat}}(\mathrm{retr},\mathcal M)$
are candidates for injection, while unlearned clusters contributing to
$D_{\mathrm{feat}}(\mathcal M,\mathrm{retr})$
are candidates for ablation.

For each direction, we assign the eligible clusters a fixed random ordering and construct four nested intervention levels. Level $k\in\{1,2,3,4\}$ contains the shortest prefix whose cumulative contribution reaches at least $k/4$ of the total directional discrepancy, with level~4 containing all eligible clusters. Thus,
\[
    C_1\subseteq C_2\subseteq C_3\subseteq C_4,
    \qquad
    D_1\subseteq D_2\subseteq D_3\subseteq D_4,
\]
where $C_k$ and $D_k$ denote the retraining and unlearned clusters selected for injection and ablation, respectively.

\paragraph{Control interventions.}
To distinguish correction of the identified feature mismatch from generic changes of comparable form, we replace the targeted retraining responses with random responses. Each direction is sampled as $w\sim\mathcal N(0,I)$ and normalized to unit norm in the full activation space; its response is obtained by projecting the centered unlearned-model activations onto it, with no additional variance or norm matching. For downstream transfer, we keep the same number of injected dimensions but use random response vectors instead of the selected retraining features. For partial relearning, we keep the targeted ablations fixed and replace only the injected retraining responses by random ones. Thus, the control preserves the amount of feature removal while changing which responses are added.

\paragraph{Evaluation after intervention.}
After each intervention, we freeze the modified representation and train a fresh prediction head using the same protocol as in Section~\ref{sec:consequences}: on the downstream task for transfer experiments, and on the retain set together with half of the forget examples for partial relearning. We also recompute the feature discrepancy on the intervened representation itself by centering it, recomputing its PCA decomposition, and applying the same feature-comparison procedure.

\subsection{Additional Experiments}\label{app:feature-injection-other}
We repeat the feature intervention experiments from Section~\ref{sec:consequences} for LTD, AMUN and NegGrad+. As in the main text, we progressively substitute mismatched features with their retraining counterparts or inject missing retraining features, and evaluate the resulting representations under downstream transfer and partial relearning.

\begin{figure*}[h]
    \centering
    \begin{subfigure}{0.32\textwidth}
        \centering
        \includegraphics[width=\linewidth]{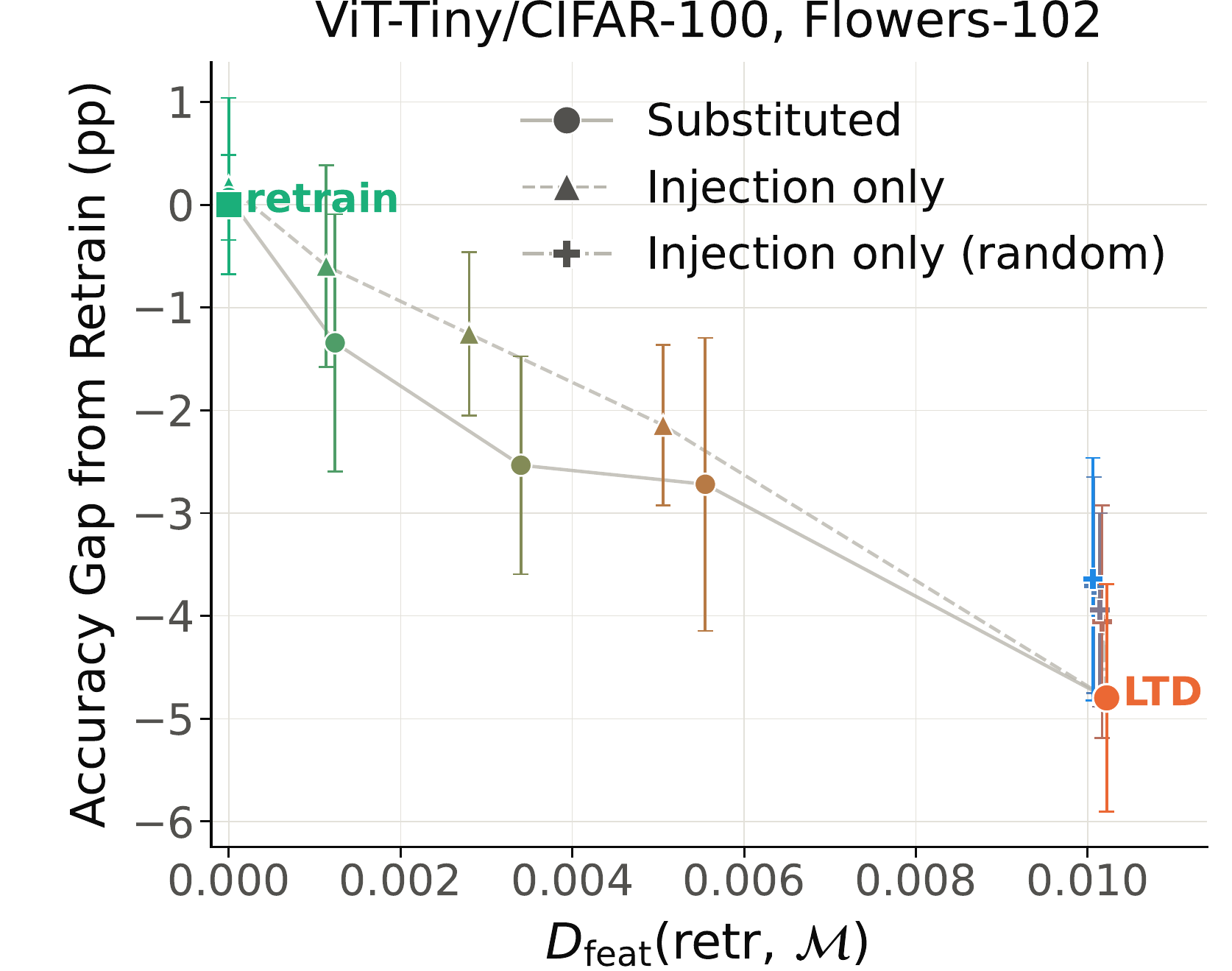}
        %\caption{Flowers-102}
        \label{fig:intervention-flowers}
    \end{subfigure}
    \hfill
    \begin{subfigure}{0.32\textwidth}
        \centering
        \includegraphics[width=\linewidth]{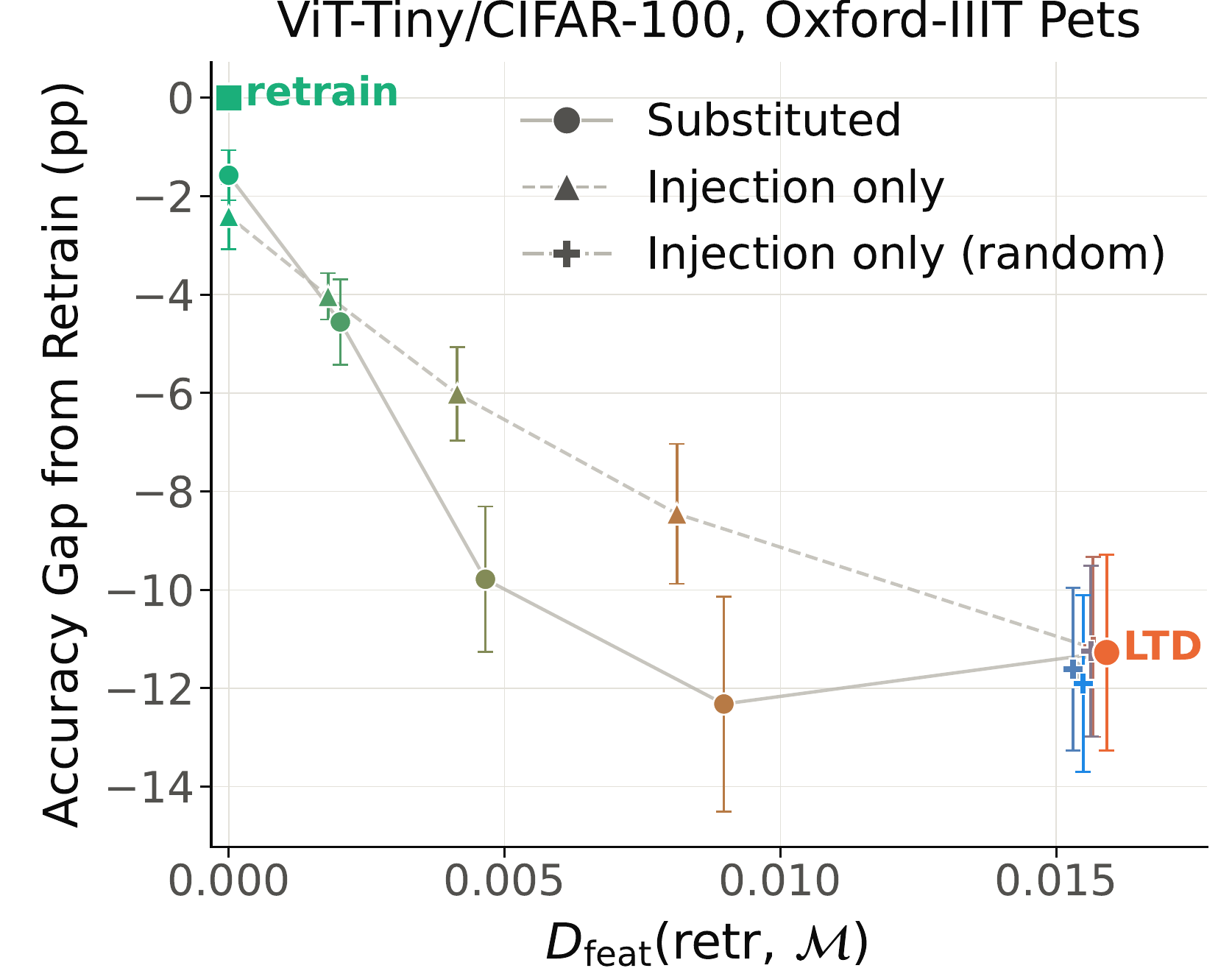}
        %\caption{Oxford-IIIT Pets}
        \label{fig:intervention-pets}
    \end{subfigure}
    \hfill
    \begin{subfigure}{0.32\textwidth}
        \centering
        \includegraphics[width=\linewidth]{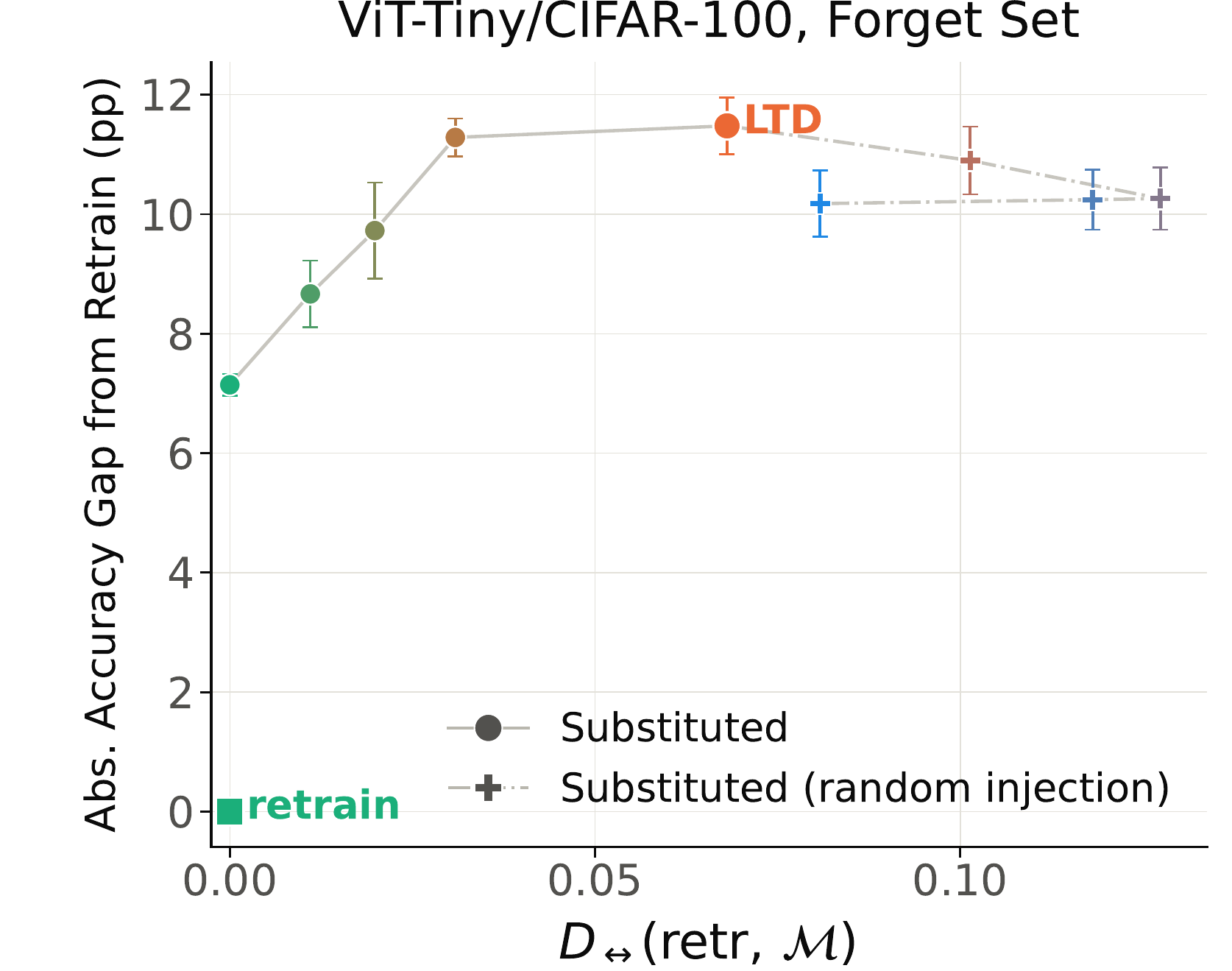}
        %\caption{Partial relearning}
        \label{fig:intervention-relearning}
    \end{subfigure}
    \caption{Behavior under direct correction of retraining-relative feature mismatch for LTD. Progressively correcting the identified mismatch moves behavior toward retraining on Flowers-102, Oxford-IIIT Pets, and partial relearning.}
    \label{fig:feature-injection-ltd}
\end{figure*}

\paragraph{LTD.}
For LTD, the intervention produces the same qualitative behavior as in the main-text experiments: progressively correcting retraining-relative feature mismatch moves subsequent performance toward the retraining baseline across the evaluated settings. Figure~\ref{fig:feature-injection-ltd} reports the corresponding intervention trajectories.

\begin{figure*}[h]
    \centering
    \begin{subfigure}{0.32\textwidth}
        \centering
        \includegraphics[width=\linewidth]{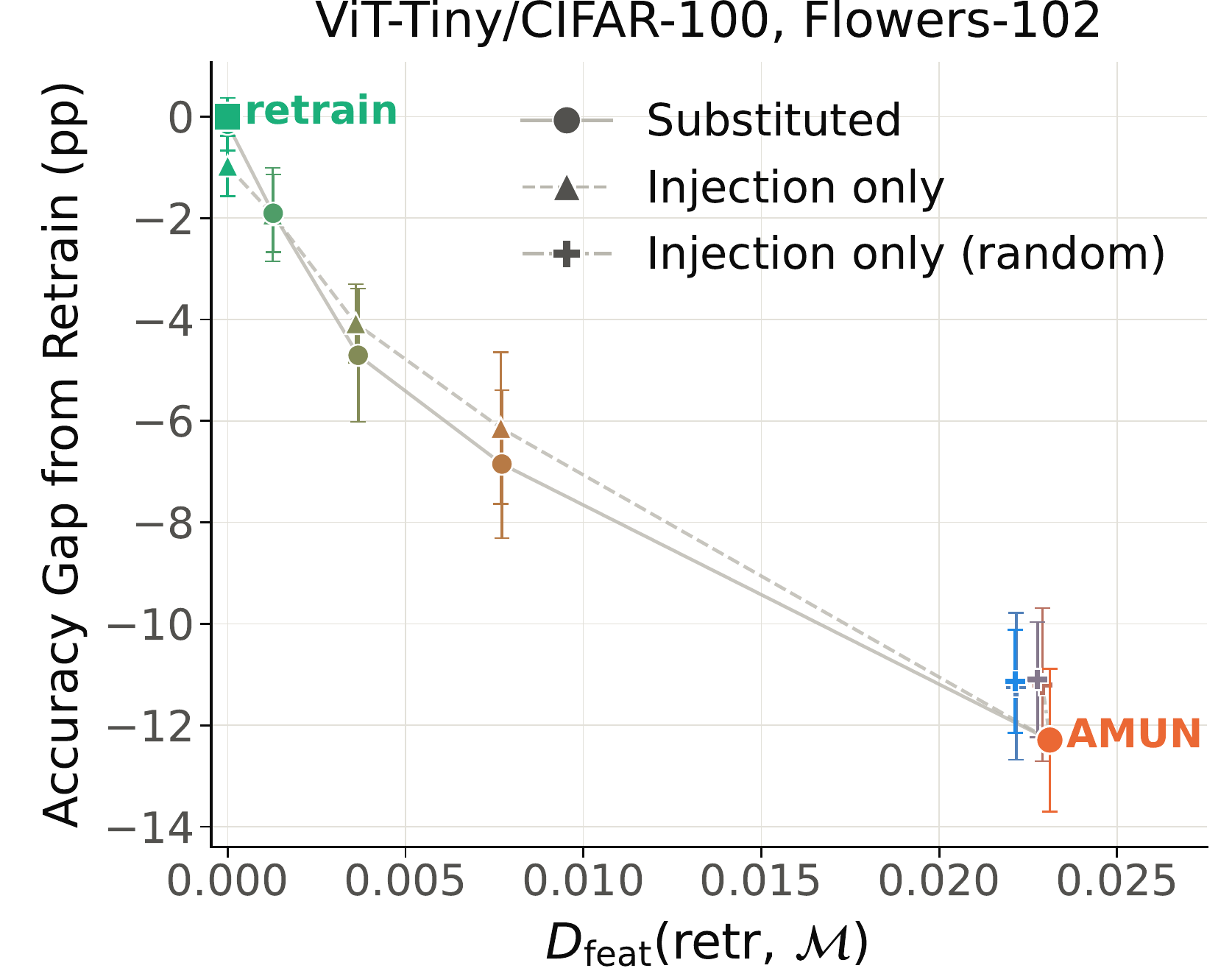}
        %\caption{Flowers-102}
        \label{fig:intervention-flowers}
    \end{subfigure}
    \hfill
    \begin{subfigure}{0.32\textwidth}
        \centering
        \includegraphics[width=\linewidth]{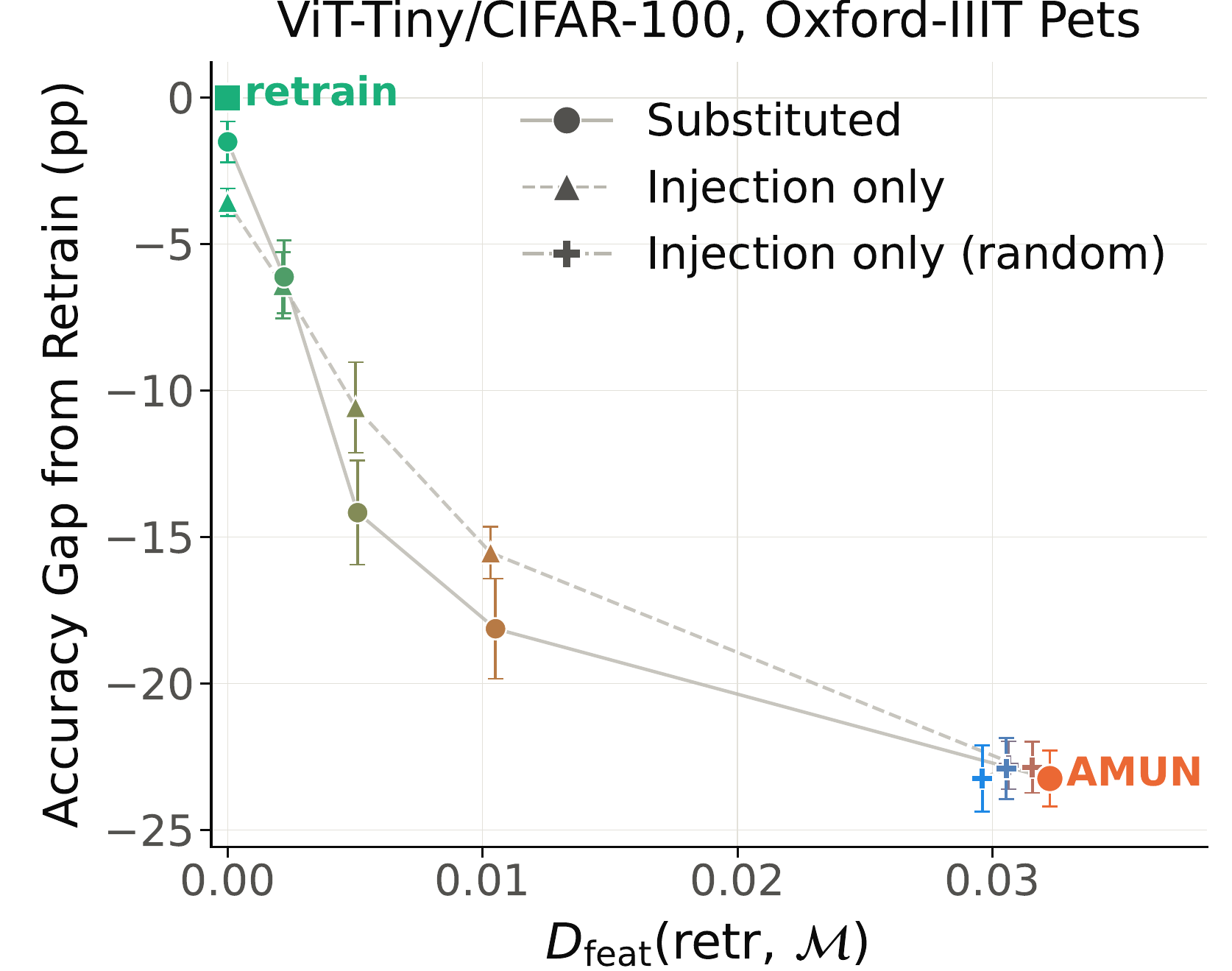}
        %\caption{Oxford-IIIT Pets}
        \label{fig:intervention-pets}
    \end{subfigure}
    \hfill
    \begin{subfigure}{0.32\textwidth}
        \centering
        \includegraphics[width=\linewidth]{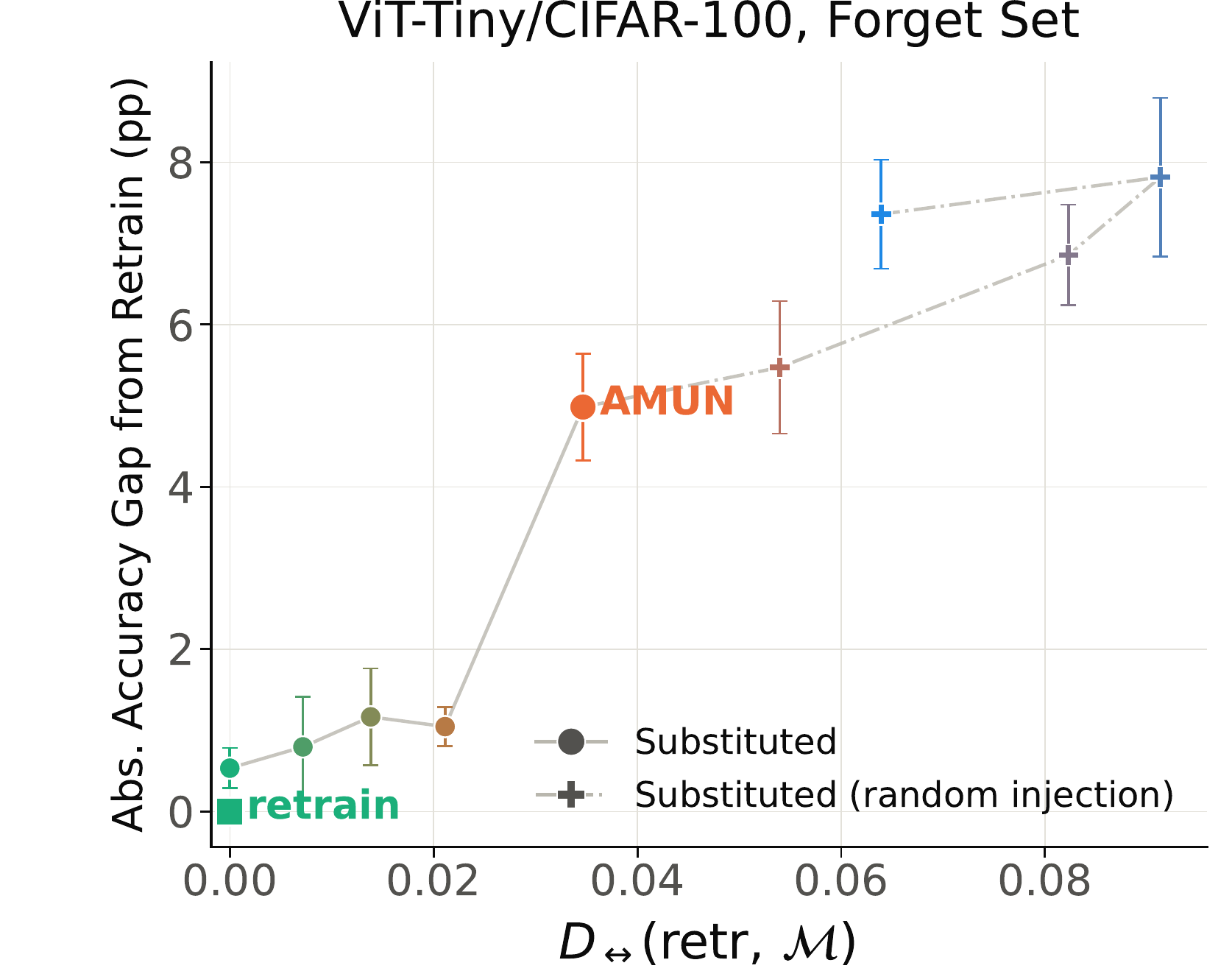}
        %\caption{Partial relearning}
        \label{fig:intervention-relearning}
    \end{subfigure}
    \caption{Behavior under direct correction of retraining-relative feature mismatch for AMUN. Correcting the identified mismatch moves downstream performance toward retraining on Flowers-102 and Oxford-IIIT Pets. Under partial relearning, targeted substitution reduces the deviation from retraining, while the corresponding random-injection control increases it.}
    \label{fig:feature-injection-amun}
\end{figure*}

\paragraph{AMUN.}
AMUN provides a complementary case in the partial-relearning experiment. Its initial deviation from retraining has the opposite sign, suggesting that missing retraining features may be more important here than removing AMUN-specific ones. Consistent with this hypothesis, targeted substitution moves relearning behavior toward retraining, whereas replacing the removed features with random directions does not. Figure~\ref{fig:feature-injection-amun} shows the corresponding trajectories.

\begin{figure*}[h]
    \centering
    \begin{subfigure}{0.32\textwidth}
        \centering
        \includegraphics[width=\linewidth]{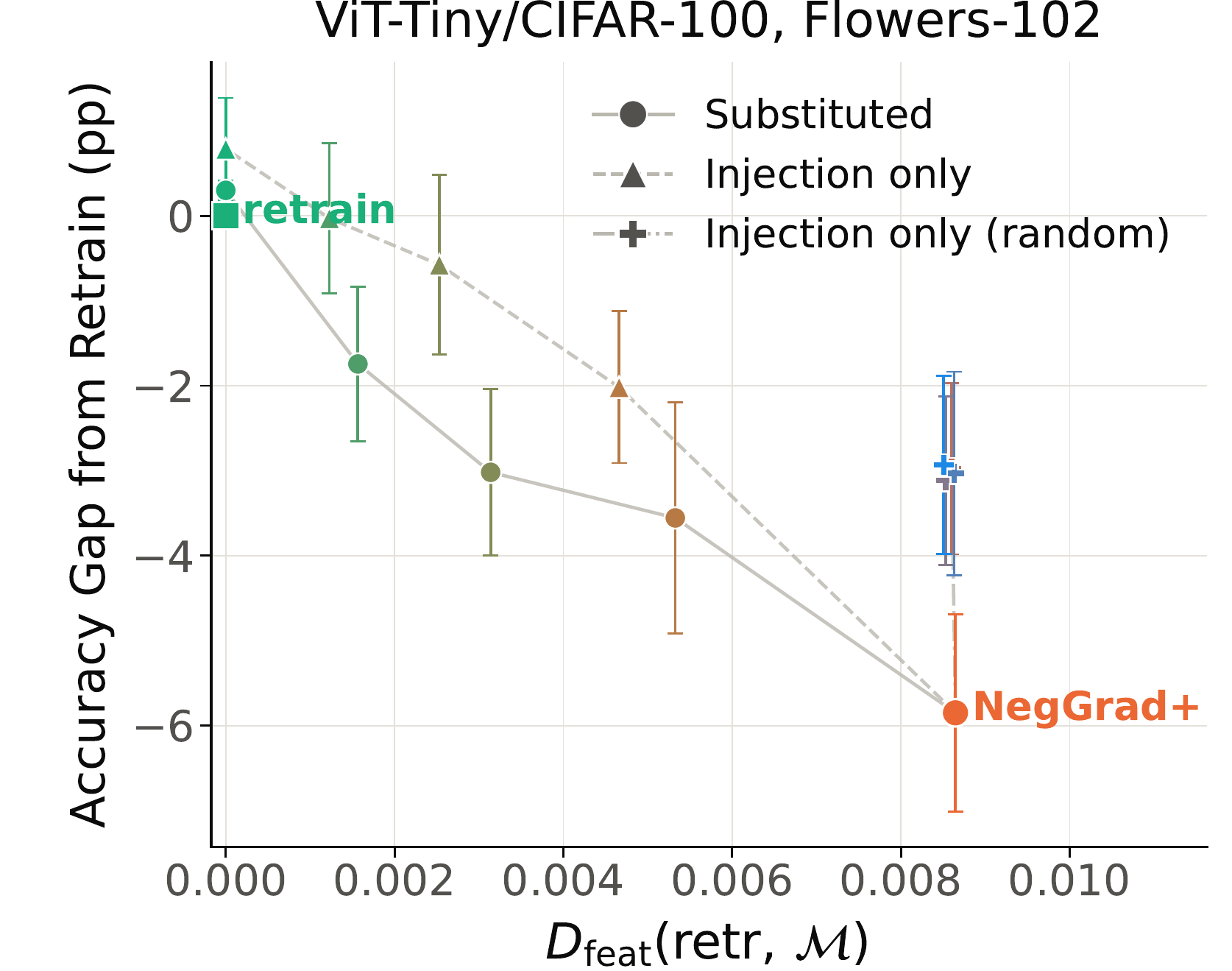}
        %\caption{Flowers-102}
        \label{fig:intervention-flowers}
    \end{subfigure}
    \hfill
    \begin{subfigure}{0.32\textwidth}
        \centering
        \includegraphics[width=\linewidth]{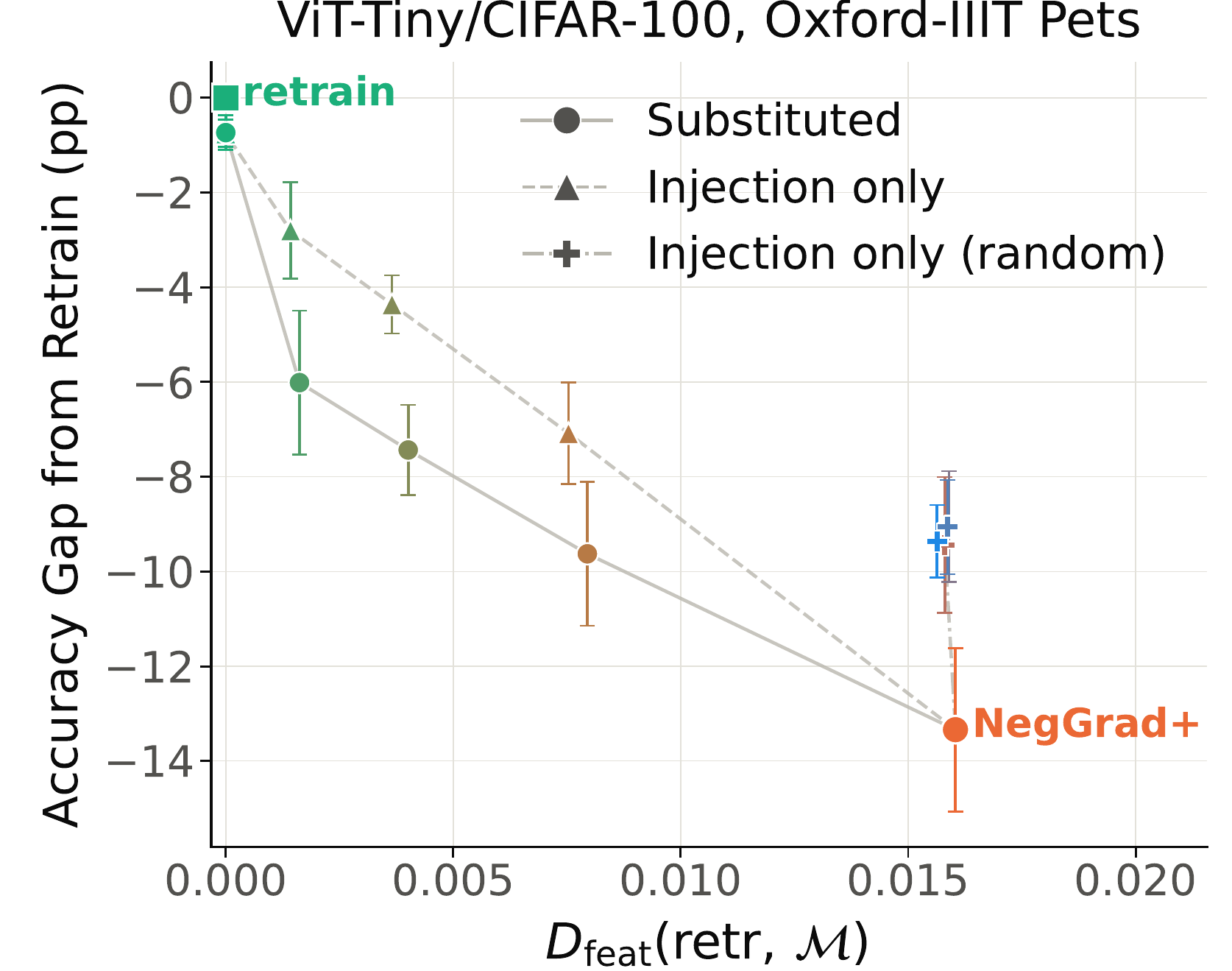}
        %\caption{Oxford-IIIT Pets}
        \label{fig:intervention-pets}
    \end{subfigure}
    \hfill
    \begin{subfigure}{0.32\textwidth}
        \centering
        \includegraphics[width=\linewidth]{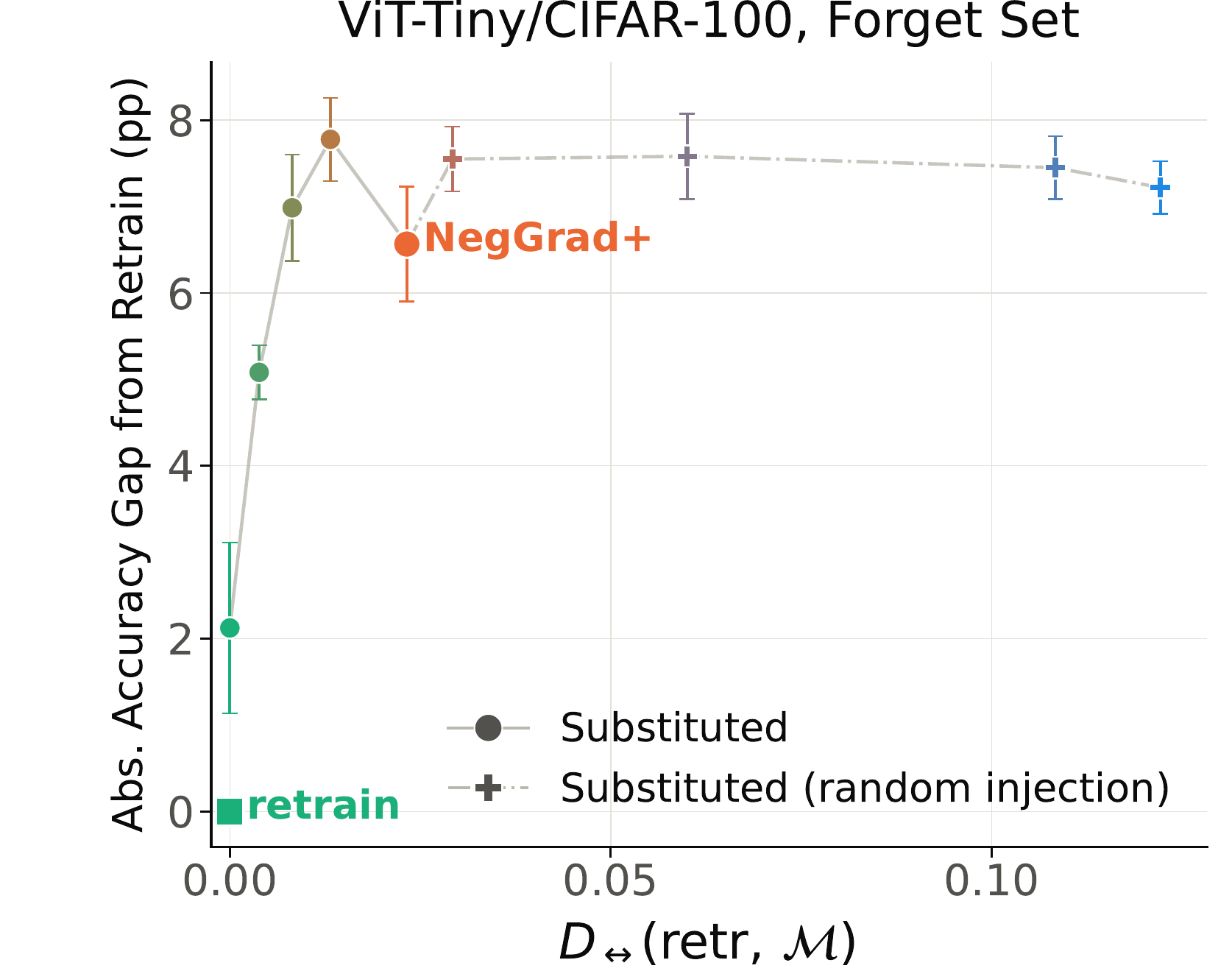}
        %\caption{Partial relearning}
        \label{fig:intervention-relearning}
    \end{subfigure}
    \caption{Behavior under direct correction of retraining-relative feature mismatch for NegGrad+. Progressively correcting the identified mismatch moves behavior toward retraining on Flowers-102, Oxford-IIIT Pets, and partial relearning.}
    \label{fig:feature-injection-ga}
\end{figure*}

\paragraph{NegGrad+.}
NegGrad+ shows the clearest intervention trajectory across the three settings. Progressively correcting the identified feature mismatch moves behavior toward retraining on Flowers-102, Oxford-IIIT Pets, and partial relearning. On the downstream tasks, injecting missing retraining features alone is already sufficient to recover much of the gap. In contrast, the corresponding random-injection control does not show the same systematic improvement, supporting the claim that the effect depends on restoring the identified retraining features. Figure~\ref{fig:feature-injection-ga} shows the corresponding trajectories.

\begin{figure*}[t]
    \centering

    \begin{subfigure}{0.48\textwidth}
        \centering
        \includegraphics[width=\linewidth]{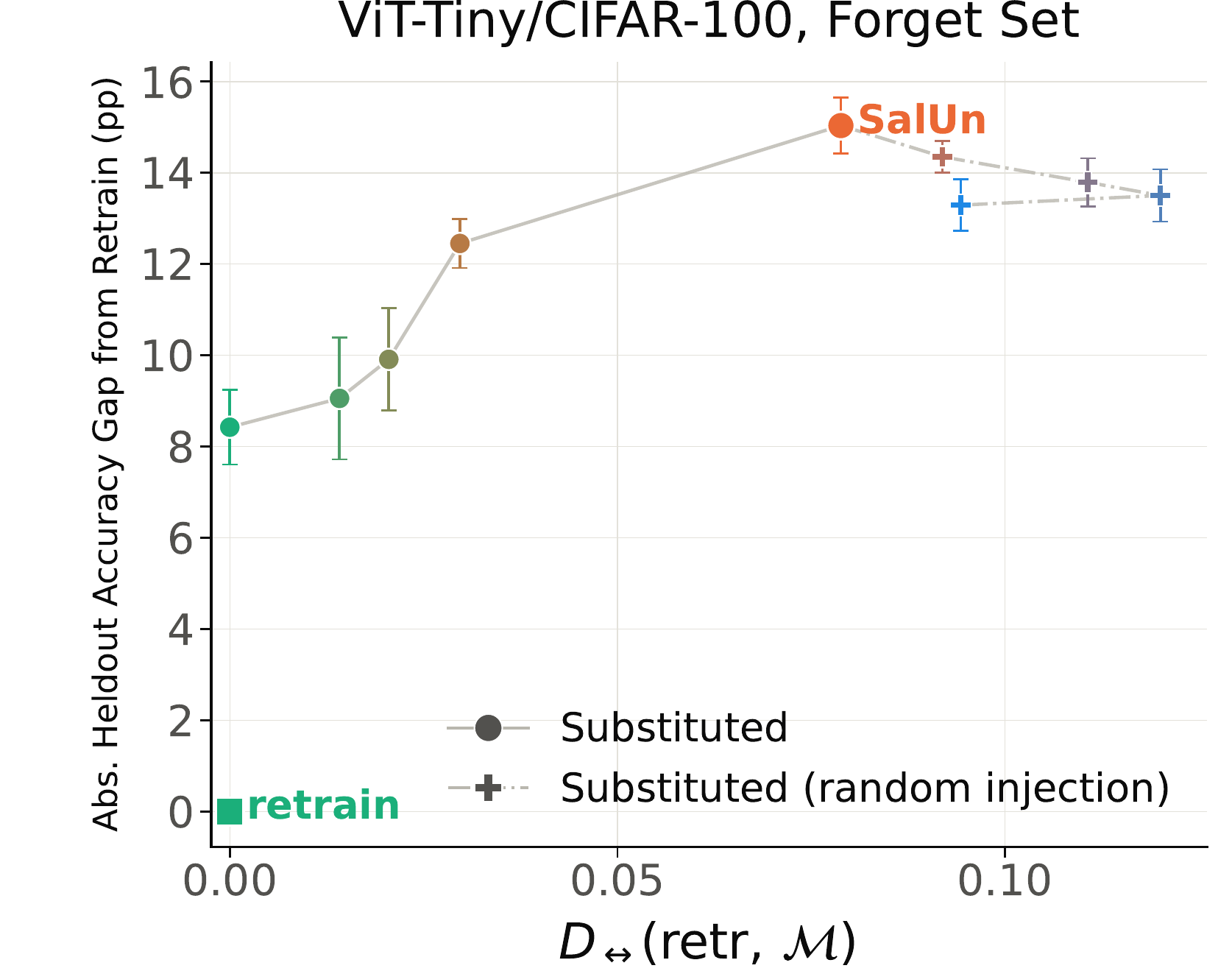}
        \caption{SalUn}
    \end{subfigure}
    \hfill
    \begin{subfigure}{0.48\textwidth}
        \centering
        \includegraphics[width=\linewidth]{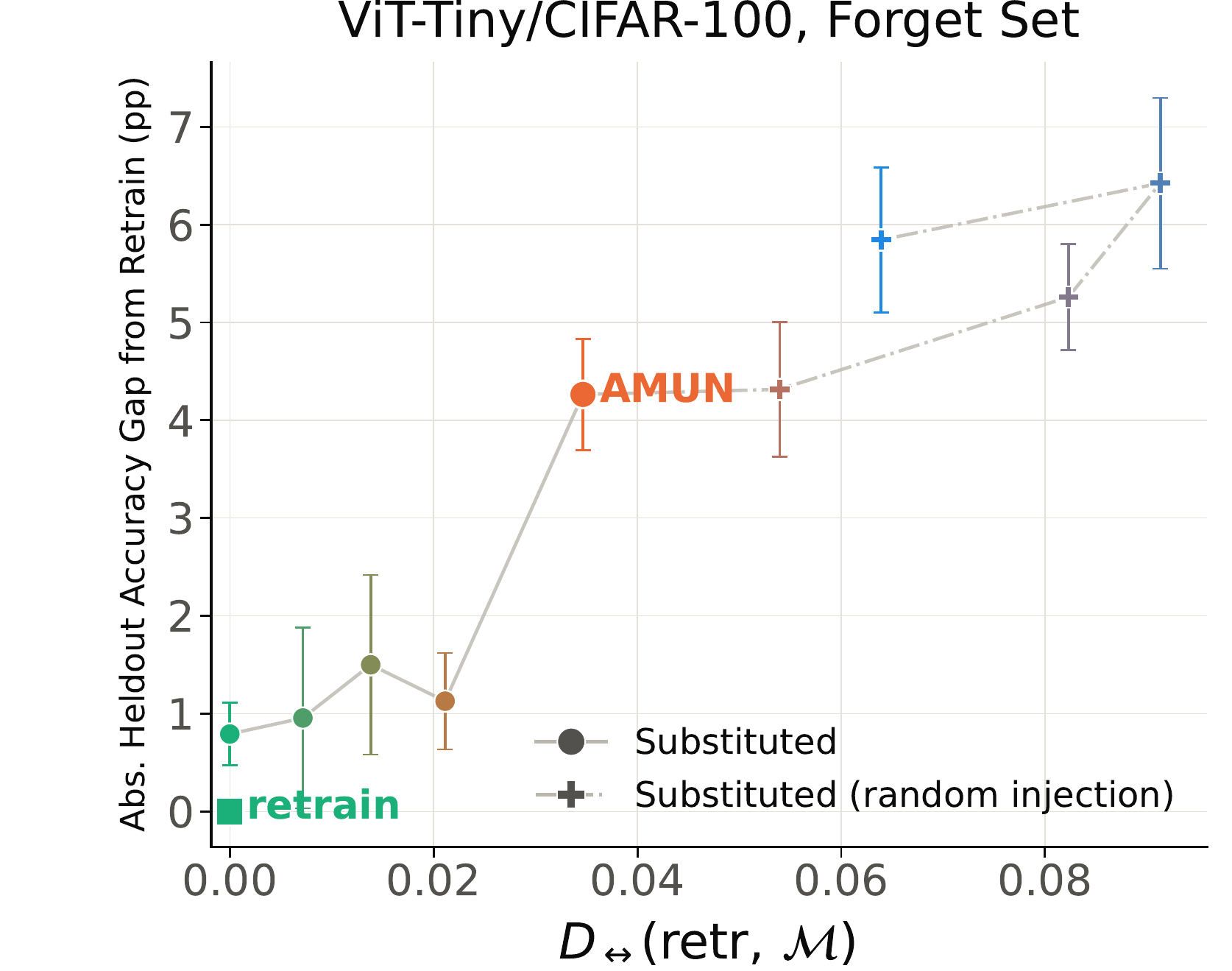}
        \caption{AMUN}
    \end{subfigure}

    \vspace{0.5em}

    \begin{subfigure}{0.48\textwidth}
        \centering
        \includegraphics[width=\linewidth]{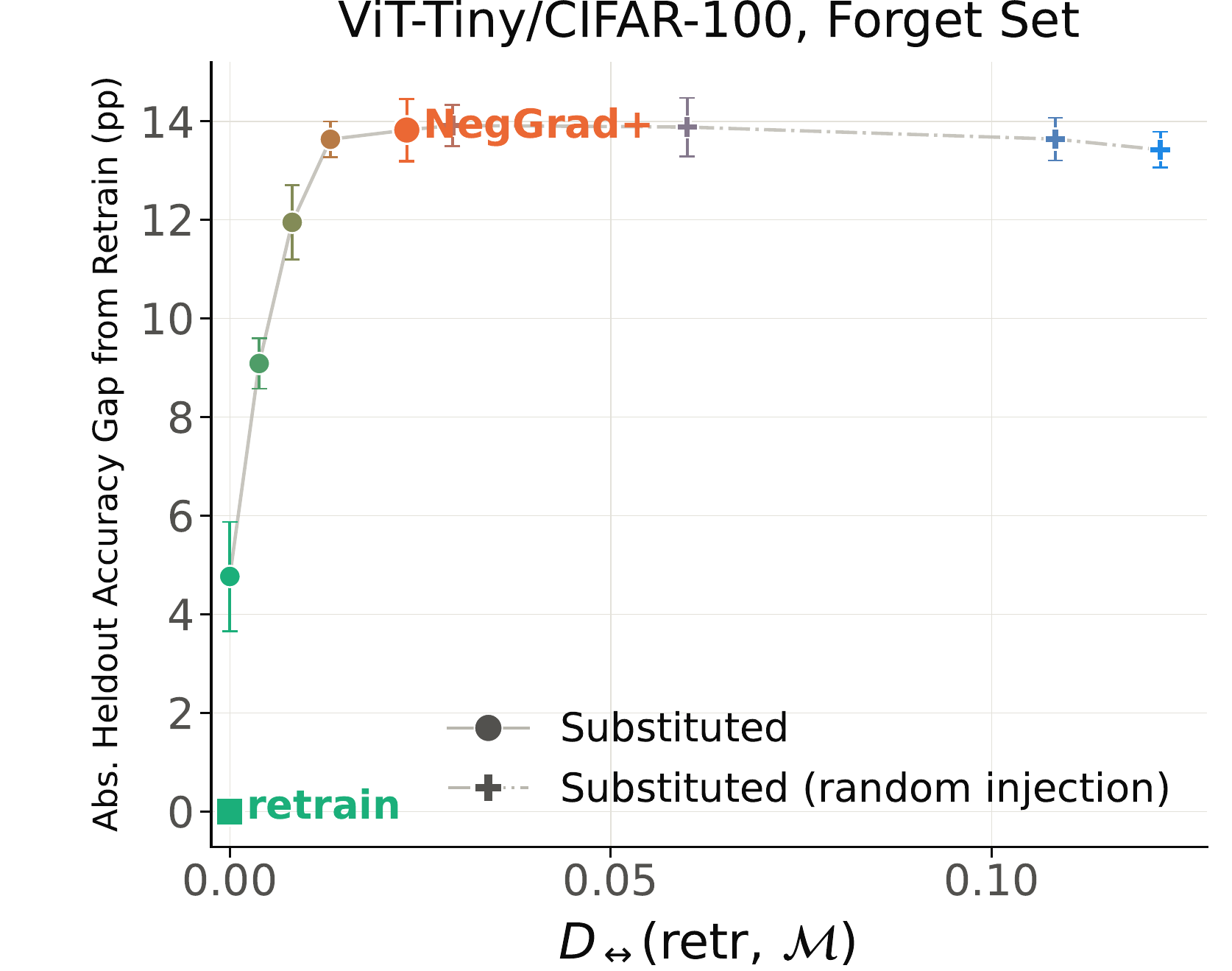}
        \caption{NegGrad+}
    \end{subfigure}
    \hfill
    \begin{subfigure}{0.48\textwidth}
        \centering
        \includegraphics[width=\linewidth]{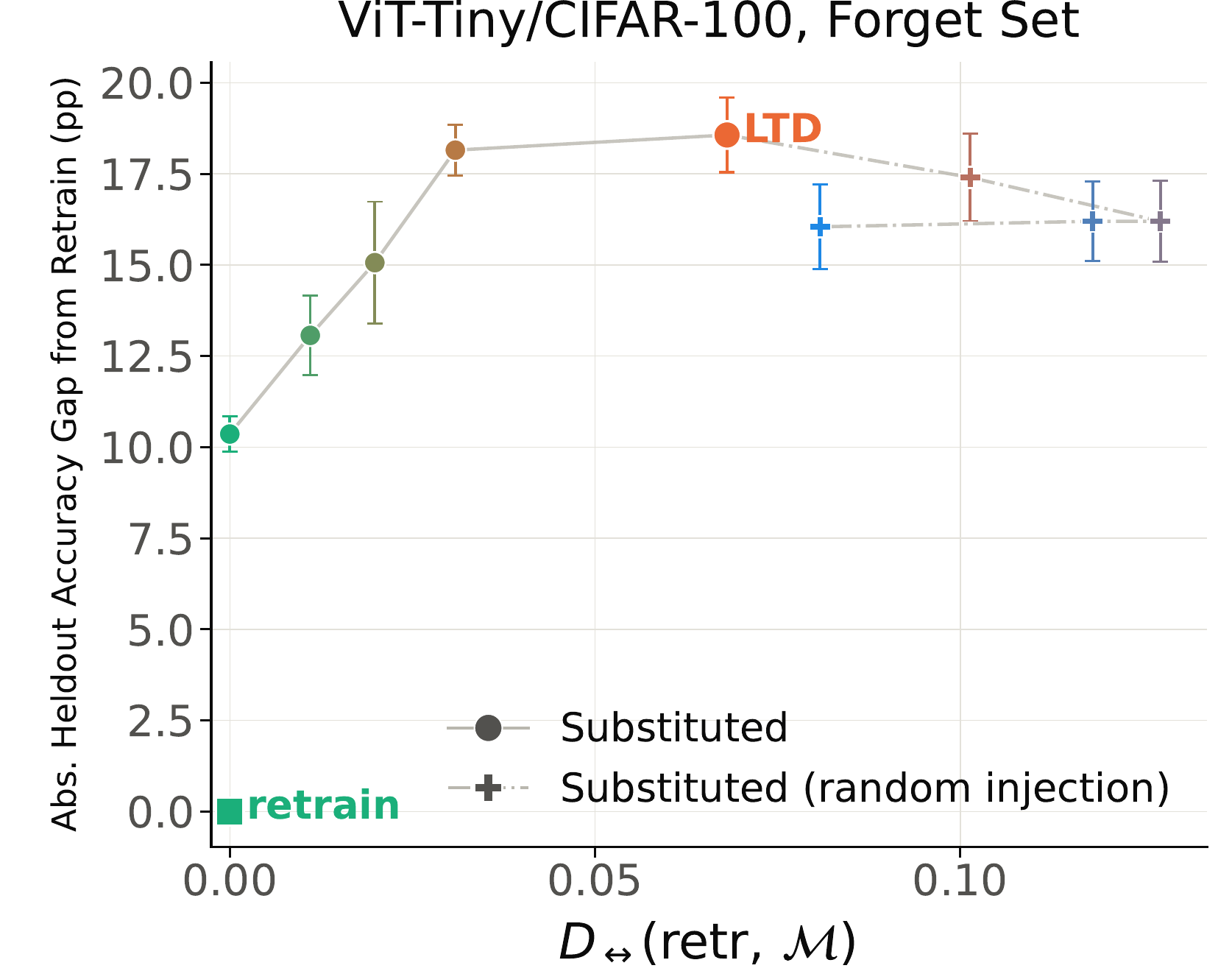}
        \caption{LTD}
    \end{subfigure}
    \caption{
    Partial-relearning intervention trajectories on a held-out forget subset for SalUn, AMUN, NegGrad+, and LTD.
    }
    \label{fig:heldout-forget-interventions}
\end{figure*}

\paragraph{Results on held-out forget set.} As an additional robustness check, we repeat the partial-relearning interventions on a held-out forget subset not used to construct the intervention. Figure~\ref{fig:heldout-forget-interventions} illustrates that SalUn, LTD, AMUN, and NegGrad+ show the same qualitative behavior as on the full forget set, with targeted correction again moving behavior toward retraining.

\begin{figure*}[t]
    \centering
    \begin{subfigure}{0.495\textwidth}
        \centering
        \includegraphics[width=\linewidth]{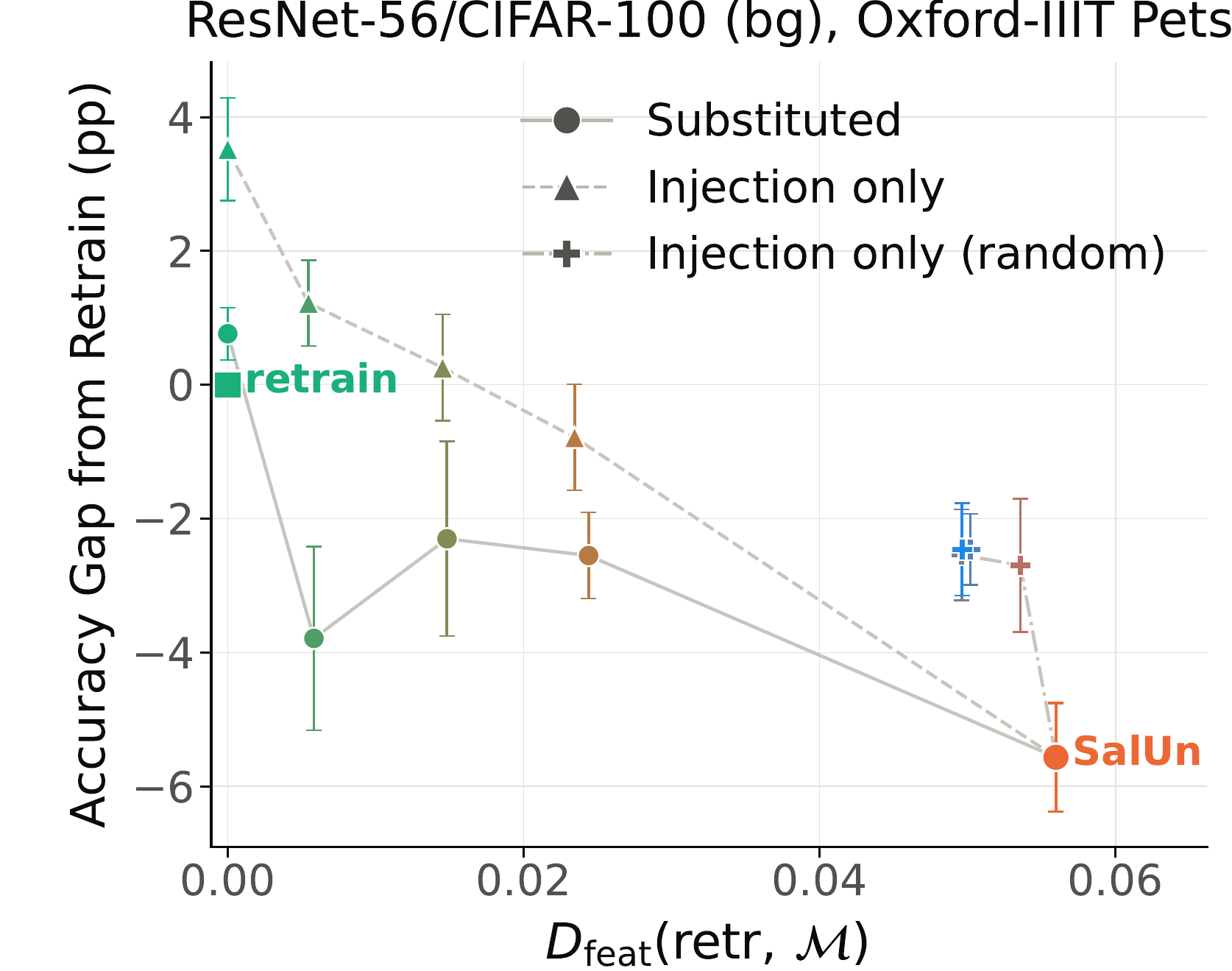}
        \caption{Oxford-IIIT Pets}
        \label{fig:bluegreen-salun-pets}
    \end{subfigure}
    \hfill
    \begin{subfigure}{0.495\textwidth}
        \centering
        \includegraphics[width=\linewidth]{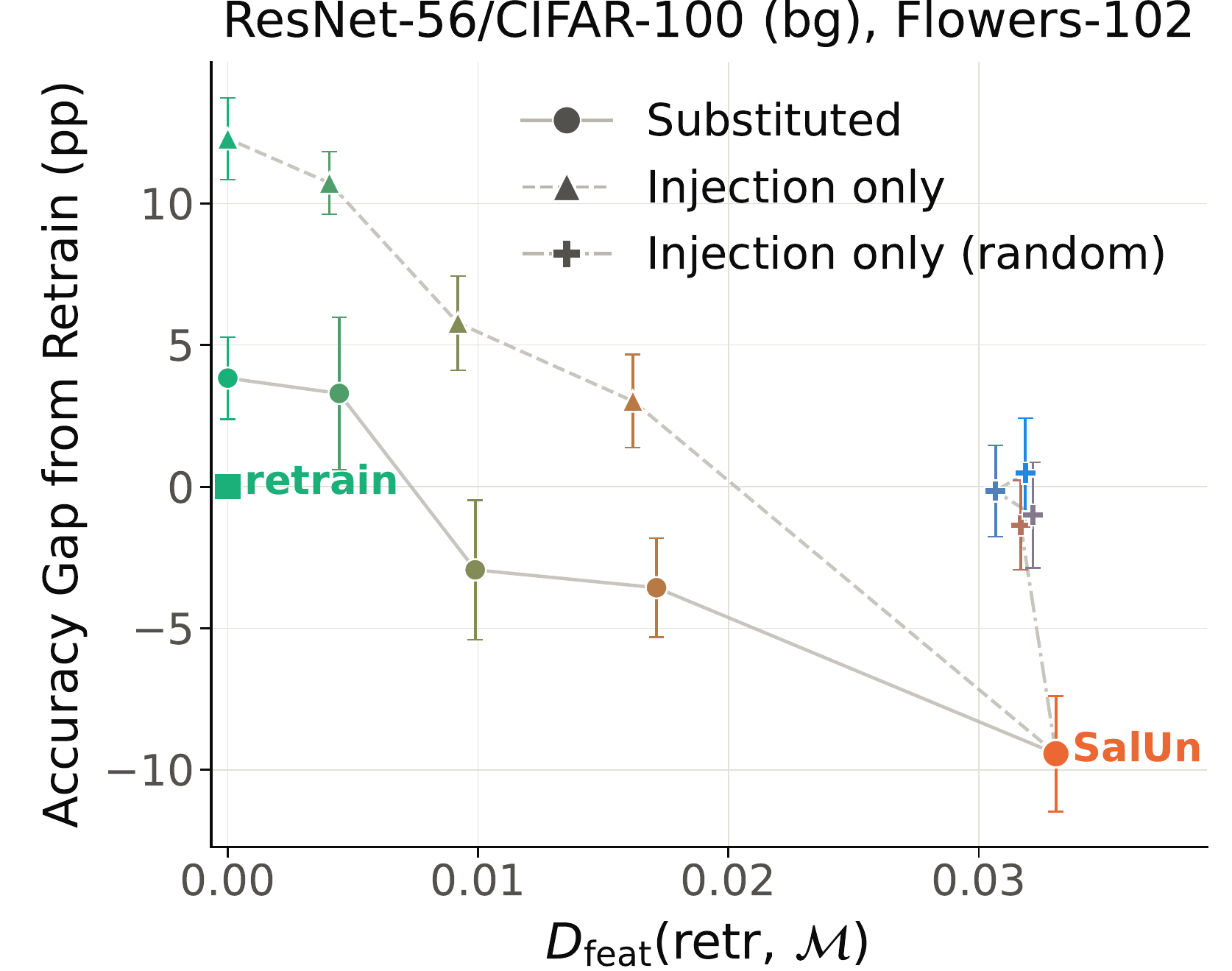}
        \caption{Flowers-102}
        \label{fig:bluegreen-salun-flowers}
    \end{subfigure}
    
        \vspace{0.8em}
        
    \begin{subfigure}{0.495\textwidth}
        \centering
        \includegraphics[width=\linewidth]{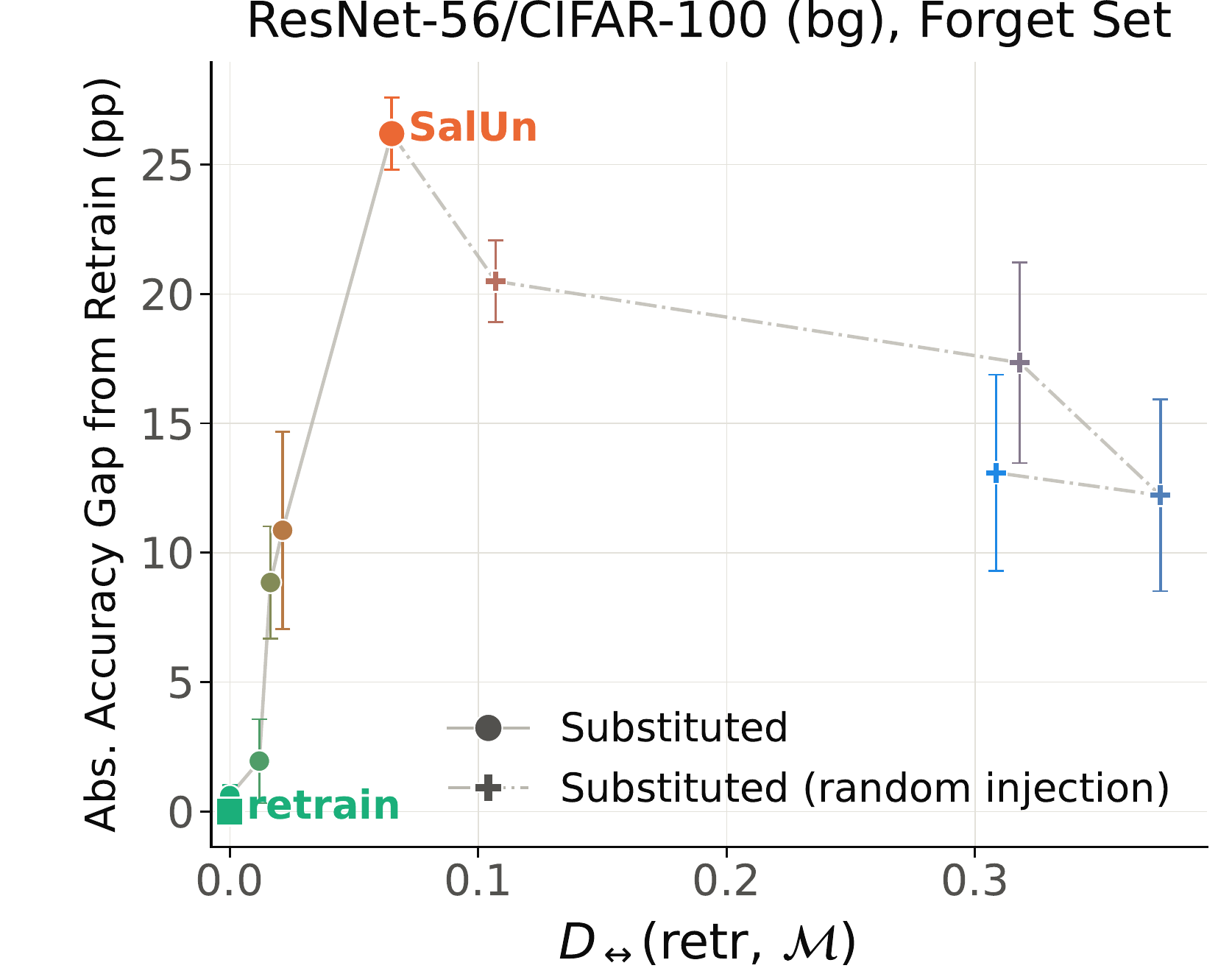}
        \caption{Forget set}
        \label{fig:bluegreen-salun-forget}
    \end{subfigure}
    \hfill
    \begin{subfigure}{0.495\textwidth}
        \centering
        \includegraphics[width=\linewidth]{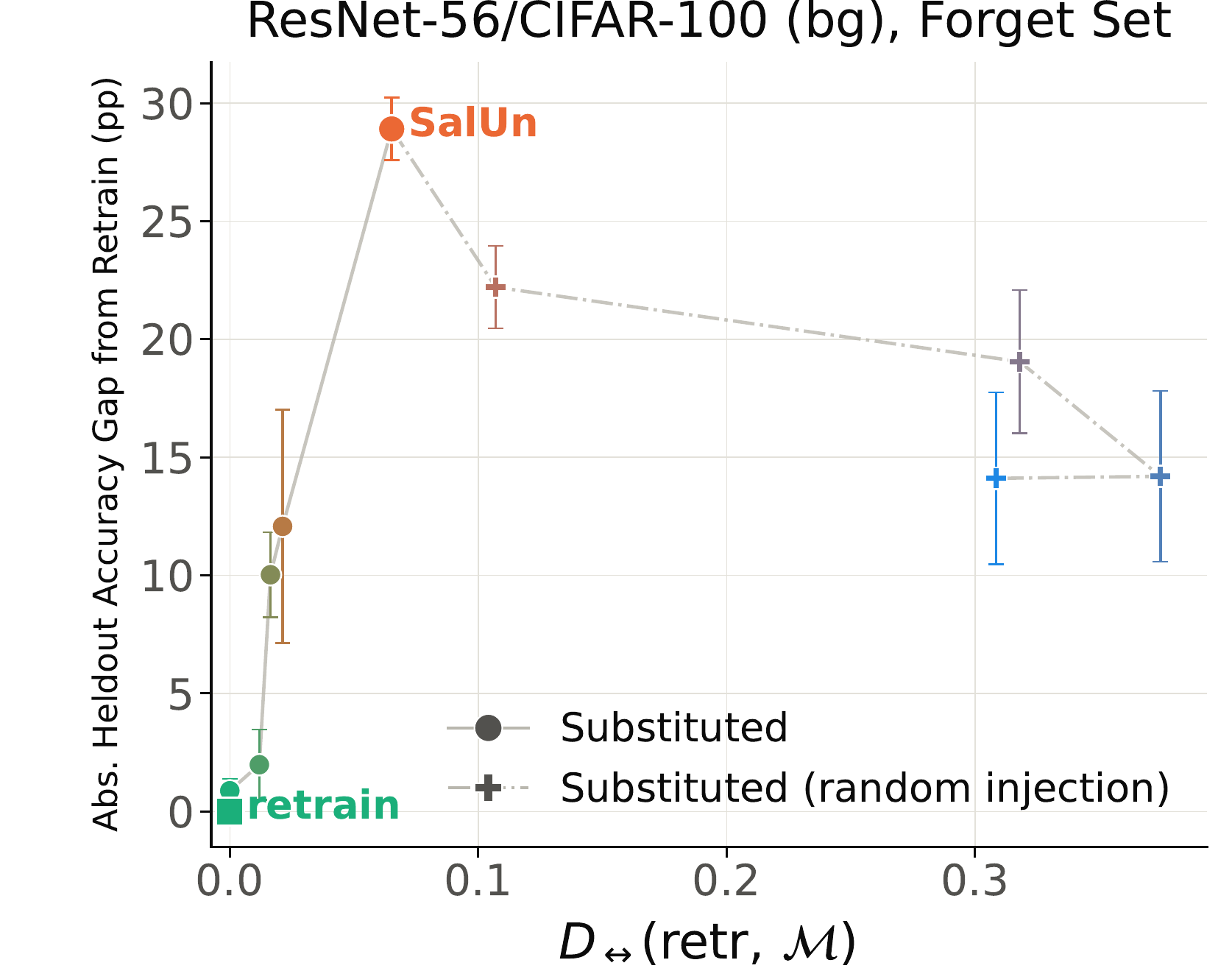}
        \caption{Held-out forget subset}
        \label{fig:bluegreen-salun-heldout}
    \end{subfigure}
    \caption{
    Intervention trajectories for SalUn under the blue-green deletion setting, evaluated on downstream transfer and partial relearning.
    }
    \label{fig:bluegreen-salun-interventions}
\end{figure*}

Finally, we repeat the SalUn intervention experiment under the blue-green deletion setting (Figure~\ref{fig:bluegreen-salun-interventions}). The same qualitative behavior is observed, with targeted correction moving the adapted model toward retraining.

\end{document}